\documentclass[twoside,11pt, preprint]{article}

\usepackage{blindtext}
\usepackage{jmlr2e}

\usepackage{balance}
\usepackage{bm}
\usepackage{amsfonts}
\usepackage{breqn}
\usepackage{amsmath}
\usepackage{pifont}
\usepackage{marginnote}
\usepackage{adjustbox}
\usepackage{lipsum}
\usepackage{graphicx}
\usepackage{fancyhdr}
\usepackage{hyperref}

\usepackage{amssymb}
\usepackage{graphicx}
\usepackage{calc}
\usepackage{amsmath}

\usepackage{ifthen}
\usepackage{xparse}
\usepackage{xcolor}
\usepackage{lmodern}
\usepackage{fix-cm}
\usepackage{subcaption}

\usepackage{tikz}
\usetikzlibrary{matrix, positioning}
\usetikzlibrary{arrows.meta, positioning, calc, fit, backgrounds}
\usepackage{color}
\usepackage{cancel}
\usepackage{soul}

\definecolor{lightred}{RGB}{255,0,0}
\definecolor{lightblue}{RGB}{0,0,255}

\begin{document}

\title{A Modular Perspective for Modeling MARL Systems} 

\author{\name Tiziano Maisonhaute 
  \email tiziano.maisonhaute@lirmm.fr\\
  \addr LIRMM, Université de Montpellier, CNRS, Montpellier, France \\
        CIRAD, UPR Recyclage et Risque, F-97743 Saint-Denis, La Réunion, France\\
        Recyclage et Risque, CIRAD, Université de Montpellier, Montpellier, France
  \AND
  \name Jean-Christophe Soulié
  \email jean-christophe.soulie@cirad.fr\\
  \addr CIRAD, UPR Recyclage et Risque, F-97743 Saint-Denis, La Réunion, France\\
        Recyclage et Risque, CIRAD, Université de Montpellier, Montpellier, France
  \AND
  \name Fabien Michel
  \email fmichel@lirmm.fr\\
  \addr LIRMM, Université de Montpellier, CNRS, Montpellier, France  \\
}

\editor{}

\maketitle

\newcommand{\eg}{for example~}
\newcommand{\ie}{that is~}

\newcommand{\class}[2][]{\textbf{\texttt{\textcolor{black}{#2}}}\ifthenelse{\equal{#1}{}}{}{ (\textit{#1})}}
\newcommand{\variable}[1]{\textbf{\texttt{#1}}}
\NewDocumentCommand{\method}{m o o}{%
  \textbf{\texttt{#1}}%
  \IfValueT{#2}{(\class{#2})}%
  \IfValueT{#3}{ $\rightarrow$~\class{#3}}%
}

\newcommand{\etc}{, etc.}

\pagestyle{fancy}
\fancyhead{}


\begin{abstract}
The growing complexity of Multi-Agent Reinforcement Learning (MARL) has fragmented the literature, obscuring underlying architectural choices behind diverse implementations. This fragmentation makes it difficult to evaluate critical modeling dimensions in isolation, identify how components affect the resulting system dynamics, and compare, reuse, or combine contributions across the literature.

This paper proposes a modular perspective for modeling, analyzing, and implementing MARL systems. We decompose them into eight functional modules: agents, environments, learning, reward modeling, training and execution modes, communication, modeling of other agents, and evaluation criteria. For each module, we clarify its role and review representative approaches.

We instantiate this decomposition in MARLKIT, a framework whose modules are explicit and interchangeable. MARLKIT demonstrates how the proposed perspective supports transparent experimental assumptions and the controlled variation and combination of MARL mechanisms. Experiments illustrate its ability to prototype, compare, reuse and combine modules while making differences between configurations explicit.

Overall, the paper provides a structured perspective on MARL modeling and implementation, offering a common language for making modeling choices explicit, supporting clearer descriptions of MARL approaches, and supporting more controlled experimentation through the systematic isolation and combination of MARL mechanisms.

\end{abstract}



\section{Introduction}
Reinforcement Learning (RL) has proven to be a flexible Machine Learning paradigm for producing autonomous agent behaviors across a wide range of, possibly very complex, tasks ranging from game theory problems \citep{moody2004stochastic} to autonomous driving \citep{elallid2022comprehensive}. The associated learning process relies on making the agent interact with its environment so that a reward is obtained with respect to the taken action and the new environment state, thus enabling the agent to evolve its behavior accordingly. More fundamentally, using RL presupposes the reward hypothesis, stating that "\textit{Goals and purposes can be well thought of as the maximization of the expected value of the cumulative sum of a received scalar signal (called reward)}" \citep{sutton2018}.

Today there exists a number of mature libraries allowing to develop RL-based experiments such as Gymnasium\footnote{Gymnasium is a fork of Gym \citep{brockman2016openai}, the latter being deprecated.} \citep{towers2025gymnasium}, RLlib \citep{liang2018rllib}, Stable-Baselines3 \citep{raffin2021stable} or TorchRL \citep{bou2024torchrl}. These frameworks are relevant for the RL community because they are backed by decades of RL research which has produced a common background and vocabulary for RL practitioners. Moreover, they are usually designed having in mind the MDP framework, which provides a useful representation of key RL modeling elements, which are themselves part of the RL community common background.

Multi-Agent Reinforcement Learning (MARL) extends RL to settings where multiple learning agents interact within a shared environment. While many RL core concepts still hold in a multi-agent setting, it turns out that crucial differences exist, new challenges arise and existing ones could be exacerbated. Most especially, the reward hypothesis requires extension in MARL because making the whole Multi-Agent System (MAS) achieve the desired task cannot be directly reduced to the maximization of a single agent's expected cumulative reward. Furthermore, having all the agents maximize individual rewards may not lead to an efficient collective outcome. In fact, one main challenge in MARL is that the outcome of an agent's action depends on the actions of others, which makes simultaneous learning quite challenging, since computing the reward of a specific action is not trivial in many cases.

So, designing a MARL experiment is thus more difficult and contains many modeling challenges and pitfalls. Therefore, a significant part of the MARL research still seeks for useful generalizable modeling principles, which could shape the field in a way that would both foster comparison among proposed approaches and ease the capitalization of relevant contributions. 

From this perspective, this paper makes the hypothesis that, while it is natural that the MARL field has been built upon RL-based foundations, certain conceptual gaps in MARL research stem from this strong inheritance. 
This observation is also consistent with recent analyses of MARL experimental practices.
We agree with the observations made by \cite{gorsane2022towards}, In MARL, high variance across implementations and inconsistent comparative baselines make it difficult to test and compare the results of different algorithms, thereby affecting the quality of reported results. In response to this problem, \cite{bettini2024benchmarl} introduce BenchMARL. The objective of BenchMARL is to improve reproducibility and comparability in MARL. BenchMARL allows users to run experiments on the same environment with different algorithms and compare the results. This is an important step towards best practices in MARL experiments.

However, we believe that this issue goes beyond the comparison of experimental results. In our view, the problem also lies in how MARL approaches are defined and described. Rather than treating an approach as a monolithic entity, we argue that it should be decomposed into a set of distinct modules. Specifically, reviewing the MARL literature reveals that it is often difficult to identify how a particular contribution addresses specific MARL aspects (\eg communication among agents, models of other agents, reward modeling, etc.). This is precisely because such aspects have not yet been made sufficiently explicit in the reference literature. In particular, papers frequently present contributions at the level of complete algorithms, rather than being explicitly associated with the specific MARL aspects they modify, extend, or address.
Consequently, authors frequently fail to clearly distinguish which MARL aspects their solution affects and how it may consequently modify others. As a result, presentation and terminology vary significantly across MARL articles.

Several popular MARL libraries exist, including PettingZoo \citep{terry2021pettingzoo}, OpenSpiel \citep{lanctot2019openspiel}, RLlib \citep{liang2018rllib}, MARLlib \citep{hu2023marllib} and EPyMARL \citep{papoudakis2021benchmarking}, each with its own scope. However, their architectures are generally not organized around MAS-specific responsibilities. Agents are rarely represented as independent components coordinating their own responsibilities, while mechanisms such as communication and reward modeling are generally embedded in other components (environments, policies\etc). This paper argues that the MARL community would benefit from adopting an explicit Agent-Oriented Software Engineering (AOSE) perspective \citep{Winikoff2013}, together with a clearer decomposition of important MAS-specific responsibilities into first-class components.

Following this view, the main contribution of this paper is proposing an explicit decomposition of the major concerns involved in modeling and implementing a MARL experiment. The underlying goal is threefold: (1) to promote a comprehensive overview of MARL that improves the understanding of each key element in the field; (2) to use this decomposition to classify the main approaches proposed with respect to each identified concern; and (3) to demonstrate that such an approach can be used to easily implement MARL frameworks, enabling the prototyping of different methods through the efficient modularization of core MARL aspects.

To this end, this research proposes to decompose MARL systems into eight core modules: (1) the agent, (2) the environment, (3) the learning module, (4) the reward model, (5) the training and execution paradigm (is the training and/or execution centralized or decentralized), (6) the communication mechanisms, (7) the modeling of other agents, and (8) the evaluation criteria. Explicitly defining the role of each dimension may facilitate a more precise classification and comparison of MARL approaches.

To support this perspective, we implemented a MARL framework based on this decomposition as a proof of concept, namely MARLKIT\footnote{\url{https://github.com/tmaisonhaute/marlkit}}.

This article is organized as follows. Section \ref{sec:rl} presents the main concepts of RL that serve as a basis for MARL concepts. Section \ref{sec:marl} provides the fundamentals of MARL. It is structured in the same way as Section \ref{sec:rl} and shows the progression from RL to MARL. Section \ref{sec:marl_core_modules} constitutes the main contribution of this article. It partitions MARL implementation into eight core modules presented above. Finally, Section \ref{sec:marlkit} presents MARLKIT, and more specifically how the eight core modules are defined in MARLKIT.

\label{sec:rl}
\section{Reinforcement Learning Backgrounds}

This section gives the fundamentals of Reinforcement Learning (RL) and more specifically highlights the key aspects that are necessary to understand the specifities and new challenges of MARL.

\subsection{Reinforcement Learning Principle}
\begin{figure}
    \centering
    \begin{tikzpicture}

\node (agent) [rectangle, draw, minimum width=2cm, minimum height=1cm, align=center] {Agent};

\node (environment) [rectangle, draw, minimum width=2cm, minimum height=1cm, align=center, below=2cm of agent] {Environment};

\draw[->, thick] (environment) -- node[midway, right] {$state$} (agent);

\draw[->, thick] (agent.east) .. controls +(right:2.0cm) and +(right:2.0cm) .. (environment.east) node[midway, right] {$action$};

\draw[->, thick] (environment.west) .. controls +(left:2.0cm) and +(left:2.0cm) .. (agent.west) node[midway, left] {$reward$};

\end{tikzpicture}
    \caption{Reinforcement Learning Scheme}
    \label{fig:rl}
\end{figure}
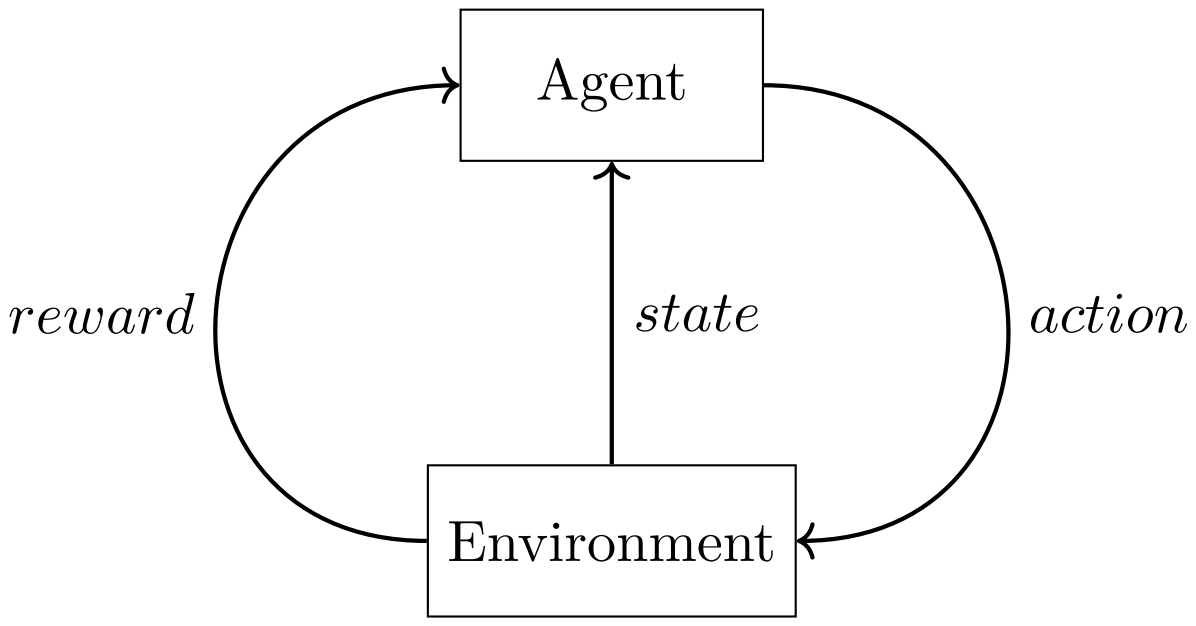

RL enables an autonomous agent (\eg robot, conversational agent, video game character, etc.), to learn behaviors adapted to a dedicated task through trial-and-error interaction with an environment, with the objective of maximizing cumulative rewards \citep{sutton2018}. 

The learning process thus consists of a loop, as illustrated in Figure \ref{fig:rl}. First, the agent observes the state of the environment, then performs an action. Once this action has been performed, the environment computes (1) its next state, according to its dynamics, and (2) a reward for the agent. Finally, the agent updates its policy based on the received reward, using a specific learning algorithm.



\subsection{Markov Decision Process}
As a mathematical model for sequential decision-making in stochastic environments~\citep{puterman2014markov}, the Markov Decision Process (MDP) is the formalism of choice in RL, as it provides a complete framework for modeling agent-environment interactions and defines many of the concepts and terms used in the field.

An MDP is defined by a tuple $(\mathcal{S}, \mathcal{A}, T, R)$ where $\mathcal{S}$ is the state space, $\mathcal{A}$ is the action space, $T(s' | s, a)$ is the transition function, which gives the probability of transitioning from a state $s$ to a state $s'$ after an action $a$, and $R(s, a, s')$ is the reward function, which computes the reward received by an agent when it performs an action $a$ leading to the transition from $s$ to $s'$.

The interaction within an MDP is described as a sequence of steps. At each step, the agent observes the current state $s \in \mathcal{S}$, takes an action $a \in \mathcal{A}$, and the environment transitions to a new state $s' \in \mathcal{S}$ based on the state-transition function $T$. The agent then receives a reward $r = R(s, a, s')$ determined by the action and the resulting state. Furthermore, the \textit{Markov property} is generally assumed, meaning that the current state $s$ contains all the information required to determine the next state; past agent-environment interactions are not needed.

\subsection{Environment Dynamics}
The environment refers to the dynamical system in which the agent operates. 
The state $s \in \mathcal{S}$ can be finite ($\mathcal{S} = \{s_1, s_2, \ldots, s_n\}$ where $n$ is the number of states in $\mathcal{S}$) or infinite (\eg when states are continuous variables such as a position). 

The modeling of the environment requires defining the two core functions defined in the previous subsection: the \textbf{Transition function} $T(s' | s, a)$ and the \textbf{Reward function} $R(s, a, s')$. 

Dynamics are \textbf{deterministic} if $T \in \{0, 1\}$, meaning the next state is fully determined by the current state and action. Conversely, dynamics are \textbf{stochastic} if $T$ takes values in $[0, 1]$, implying the next state is drawn from a probability distribution over possible outcomes.

\subsection{Reward Modeling} \label{sec:rl_reward_modeling}
A reward is the immediate feedback the environment gives after an action, indicating how good that outcome is with respect to the task's objectives. To guide the agent effectively, the reward function must accurately reflect these goals.

An RL process is typically structured into episodes, though continuous tasks may consist of a single, possibly not ending, episode. An episode is a sequence of states, actions, and rewards, starting from an initial state and terminating upon reaching a terminal state or exceeding a step limit. Through experience, the agent updates its policy to improve performance in subsequent episodes and possibly in the current episode. 

Following the reward hypothesis, the primary goal of an RL agent is to maximize cumulative rewards over time within an episode, rather than just the immediate reward. 
Thus, a key concept is $G_t$, the discounted return which is defined as the sum of discounted rewards received from time $t$ to the end of the episode. A parameter called the discount rate $\gamma$ is used to balance the importance of immediate versus future rewards. 

Defining an effective reward function is challenging, as it fundamentally shapes the agent's behavior. Poorly defined reward functions can lead to unintended behaviors or significantly slow down the learning process.

\subsection{Agent} \label{sec:rl_agent}

In the scope of RL, the term agent refers to the learning entity that actually operates in the environment by taking actions. Assuming that an \textbf{action space $\mathcal{A}$} is already defined, specifying an RL agent requires defining at least the following: 
\begin{itemize}
    \item its \textbf{policy}: a function that maps a state $s \in \mathcal{S}$ to an action $a \in \mathcal{A}$,
    \item its \textbf{learning algorithm}: a set of rules used to update its policy. The updates are based on experiences (obtained rewards).
\end{itemize}

At the start of the learning process, the agent's policy is initialized, often by randomly mapping states to actions. The learning algorithm then iteratively updates this policy to improve performance on the task. Specifically, the algorithm uses the obtained rewards as feedback signals to guide these updates.

    \subsection{Learning} \label{sec:rl_learning_module}

        \subsubsection{Policy} \label{sec:rl_policy}
A policy\footnote{Policy is sometimes referred as strategy \citep{claus98}} is a mapping from states to probabilities of taking actions: 

\begin{equation}
    \begin{aligned}
        \pi : \mathcal{S} \times \mathcal{A} \rightarrow [0, 1] \\
        \pi(s, a) \mapsto P(a|s)
    \end{aligned}
    \label{eq:policy}
\end{equation}

Where $\mathcal{S}$ is the state space and $\mathcal{A}$ is the action space. $\pi(s, a)$ is the probability of taking action $a$ in state $s$.

A policy is either deterministic or stochastic. A policy is deterministic if the probability of selecting action $a$ in state $s$ is strictly 0 or 1. In this case, the policy is thus a direct mapping: 

\begin{equation}
    \begin{aligned}
        \pi : \mathcal{S} \rightarrow \mathcal{A} \\
        \pi(s) \mapsto a 
    \end{aligned}
        \label{eq:deterministic_policy}
\end{equation}


In RL, policies are implemented either as tabular methods or via function approximation. Tabular methods store values in a table indexed by states, actions, or both, typically maintaining a value function. The table is updated by the learning algorithm during the process.

However, tabular approaches become infeasible when the state space is large or continuous. Furthermore, tabular methods treat similar states as distinct, preventing generalization. Function approximation addresses these limitations by learning a parameterized function to approximate the policy or value function. This approximation uses a large number of parameters $\theta$, which are updated during the learning process. Most of the time, a neural network is used to approximate the function and the parameters $\theta$ are the network's weights. Both methods are detailed by \cite{sutton2018}.

With respect to their underlying mechanisms, there are three main types of policies an agent can apply to select actions.

\paragraph{Value-based} \label{sec:value_based}
Value-based algorithms use a value function to estimate the expected return. If this value function accurately estimates expected returns, the agent can identify states or actions associated with higher returns. There are two primary types of value functions: the state-value function, which evaluates states, and the action-value function, which evaluates state-action pairs.

\paragraph{Policy-based}
Policy-based algorithms directly optimize policies without explicitly using a value function. A policy is typically represented by the function $\pi_\theta(s, a)$, where $\theta$ is the vector of policy parameters. During training, $\theta$ is updated to maximize the expected return.

A common approach is to model the policy as a neural network, where the network's weights correspond to the parameters $\theta$. The network takes a state representation as input and outputs either the probability distribution over actions (for a stochastic policy) or the specific action to take (for a deterministic policy). 

\paragraph{Actor-Critic}\label{sec:actor-critic}
Actor-Critic algorithms combine value-based and policy-based approaches by executing two processes in parallel: the actor and the critic \citep{konda99}.

The actor represents the policy-based component; it selects actions based on the current policy and the environment's state.

The critic represents the value-based component; it estimates a value function to evaluate the states and actions selected by the actor. By providing feedback on the quality of these decisions, the critic guides the actor's policy updates. The value function estimated by the critic may be the state-value function $V$, the action-value function $Q$, or the advantage function $A$.

        \subsubsection{Learning Algorithm}
To support a modular description of reinforcement learning systems, we consider that a learning algorithm specifically denotes the process responsible for updating the agent's policy from available data.
And therefore, it is not responsible for data generation or data collection, which are handled by other components of the system. 

In reinforcement learning, the data used by the learning algorithm typically comes from the agent's interaction with the environment. It is commonly represented as experiences, where each experience is a tuple containing the state (or observation), the action taken, the reward received, and possibly the next state (or observation). Using these experiences, the learning algorithm updates the agent's policy, value function, or other learned elements involved in decision-making.

Many learning algorithms have been proposed in the literature. For a broader presentation of learning algorithms, we refer the reader to the review by \cite{shakya2023reinforcement}.

Each algorithm has its own learning parameters. These learning parameters depends on the learning algorithm, however, most use a learning rate $\alpha$ to update the policy. The learning rate determines the magnitude of the policy update based on new experiences. The greater $\alpha$ is, the larger the update.

A widespread criterion\footnote{This criterion does not originate from RL; it was proposed by \cite{robbins1951stochastic} for \textit{stochastic approximation} methods.} for the learning rate is to have it decrease over time, satisfying the following two conditions~\citep{sutton2018}:
\begin{equation}
\sum_{t=1}^{\infty} \alpha_t = \infty
\; ; \;
\sum_{t=1}^{\infty} \alpha_t^2 < \infty
\end{equation}
The first condition ensures that the learning rate does not decrease too rapidly, guaranteeing that the cumulative updates are sufficient to overcome a poor initial policy or initial conditions. The second condition ensures that the learning rate decreases fast enough to allow convergence.

    \subsection{Challenges}
Implementing an RL solution presents significant challenges. The difficulty depends on the use case, but there are main challenges that occur in many RL implementations. This section identifies these main challenges, analyzing their causes and the problems they pose in the learning process. These challenges persist in MARL, and are often exacerbated by the multi-agent paradigm.

The primary difficulties identified in this work are: (1) non-stationarity, (2) partial observability, (3) temporal credit assignment, and (4) the curse of dimensionality.

        \subsubsection{Non-stationarity} 
\label{sec:rl_nonstationarity}
RL algorithms typically assume a stationary environment, meaning the state-transition and reward functions are time-invariant and constant over time.

Stationarity simplifies learning. Agents learn through trial and error by associating actions in specific states with received rewards. If the environment is stationary, the reward function satisfies: 

\begin{equation}
    s_{\textcolor{lightblue}{t1}}, a_{\textcolor{lightblue}{t1}} = s_{\textcolor{lightred}{t2}}, a_{\textcolor{lightred}{t2}} \Rightarrow R(s_{\textcolor{lightblue}{t1}}, a_{\textcolor{lightblue}{t1}}) = R(s_{\textcolor{lightred}{t2}}, a_{\textcolor{lightred}{t2}})
\end{equation}
Where $t1$ and $t2$ are two different time step. 
The same holds for the state-transition function.

Consequently, the expected reward for a specific action in a specific state remains constant, facilitating stable learning. Conversely, in a completely chaotic or highly non-stationary environment, past experiences become irrelevant for evaluating current actions, making learning impossible. However, an RL environment can be non-stationary while still allowing an agent to learn \citep{padakandla2020reinforcement}, though this presents a significant challenge that must be explicitly addressed.

\subsubsection{Partial Observability} \label{sec:rl_partialobs} Agents often lack access to the full state of the environment, receiving only a partial observation instead. In such cases, the agent must make decisions based on incomplete information, which may not fully capture the environment's dynamics. Partially observable environments are common in real-world applications, such as robotics or autonomous driving, where sensors may not capture all relevant information.

Learning is significantly more challenging in these settings. Consider an agent operating in an environment where two distinct states are indistinguishable from its perspective. Since the agent selects actions based on observations, it may behave identically in both states, even if the optimal actions differ. Consequently, even with perfect knowledge of the optimal policy, the agent cannot act optimally because it cannot determine its current state.

To address this issue, the Partially Observable Markov Decision Process (POMDP) framework is widely used. In this framework, the agent maintains a belief state—a probability distribution over possible states—or leverages memory mechanisms to track the history of observations and actions \citep{igl2018deep}.

\subsubsection{Temporal Credit Assignment}
\label{sec:rl_credit_assignment}
The temporal credit assignment problem refers to the difficulty of determining which specific actions are responsible for a received reward \citep{sutton1984temporal, ausin2021tackling}. Effectively, an agent's actions affect not only the immediate reward but also future rewards.

As discussed in Section \ref{sec:rl_reward_modeling}, an action should not be evaluated solely based on the immediate reward it yields, as it can have long-term consequences by altering the environment's state and influencing future outcomes. Consequently, when an agent receives a reward, it is often unclear whether that reward is attributable to the most recent action or a sequence of previous actions.

Consider a game of chess as a simple example. The agent receives a reward only at the end of the game: $+1$ for a win, $-1$ for a loss, and $0$ for a draw. If the agent wins, identifying the specific cause of the reward is challenging; it is certainly not attributable solely to the final move, but rather to the cumulative effect of the entire sequence of moves.

\subsubsection{Curse of Dimensionality} 
\label{sec:rl_curse_dimensionality}
The size of the state-action space is given by $|\mathcal{S}| \times |\mathcal{A}|$. The difficulty of exploring this space grows exponentially with the number of states and actions, a phenomenon known as the \textit{curse of dimensionality} \citep{wang2021sample}.

This poses a significant challenge for tabular RL solutions \ref{sec:rl_policy}, which rely on maintaining a value table for every state-action pair. Furthermore, high dimensionality can severely slow down the learning process due to the vast exploration requirements. In some high-dimensional spaces, it may be impossible to explore all state-action pairs \citep{korkmaz2024survey}.

\section{Multi-Agent Reinforcement Learning}\label{sec:marl}
The transition from single-agent to multi-agent learning in a shared environment fundamentally alters the concepts discussed previously. This section outlines the new setting, organized as follows: (1) an introduction to Multi-Agent Systems (MAS) and the revised learning loop; (2) the definition of Stochastic Games as a generalization of MDPs for MAS; (3) an analysis of dynamic shifts; (4) a discussion on why simple reward maximization is no longer a sufficient objective; (5) an examination of agent and policy adaptations in MARL; and (6) the identification of key challenges, which represent exacerbated versions of those found in single-agent RL.

    \subsection{From RL to MARL}
    
        \subsubsection{Multi-Agent Systems}

Multi-Agent Systems (MAS) comprise several autonomous agents that interact with their environment and with one another. These interactions can vary in intensity (from loose to tight coupling) and nature (cooperative, competitive, or mixed), and may involve explicit communication \citep{ferber1999multi}.

MAS offer several practical advantages for solving complex problems \citep{balaji2010introduction}: (1) Parallelism: Multiple agents allow for the parallelization of the problem-solving process; (2) Robustness: The system is more resilient to failures, as the malfunction of a single agent does not necessarily compromise the entire system; (3) Flexibility: Agents can be dynamically added, removed, or modified to adapt to changing requirements; (4) Decentralization: MAS do not require a central entity to coordinate all tasks, enabling distributed control.

Beyond problem-solving, MAS are also widely used to model and simulate complex systems \citep{michel09mas_and_ms}, particularly those involving multiple interacting entities. Common applications include economics \citep{guessoum2004using}, crowd simulation \citep{ulicny2002towards}, or stakeholder behavior analysis \citep{arnejo2025, zhang2026multi}.

        \subsubsection{MARL Learning Loop}

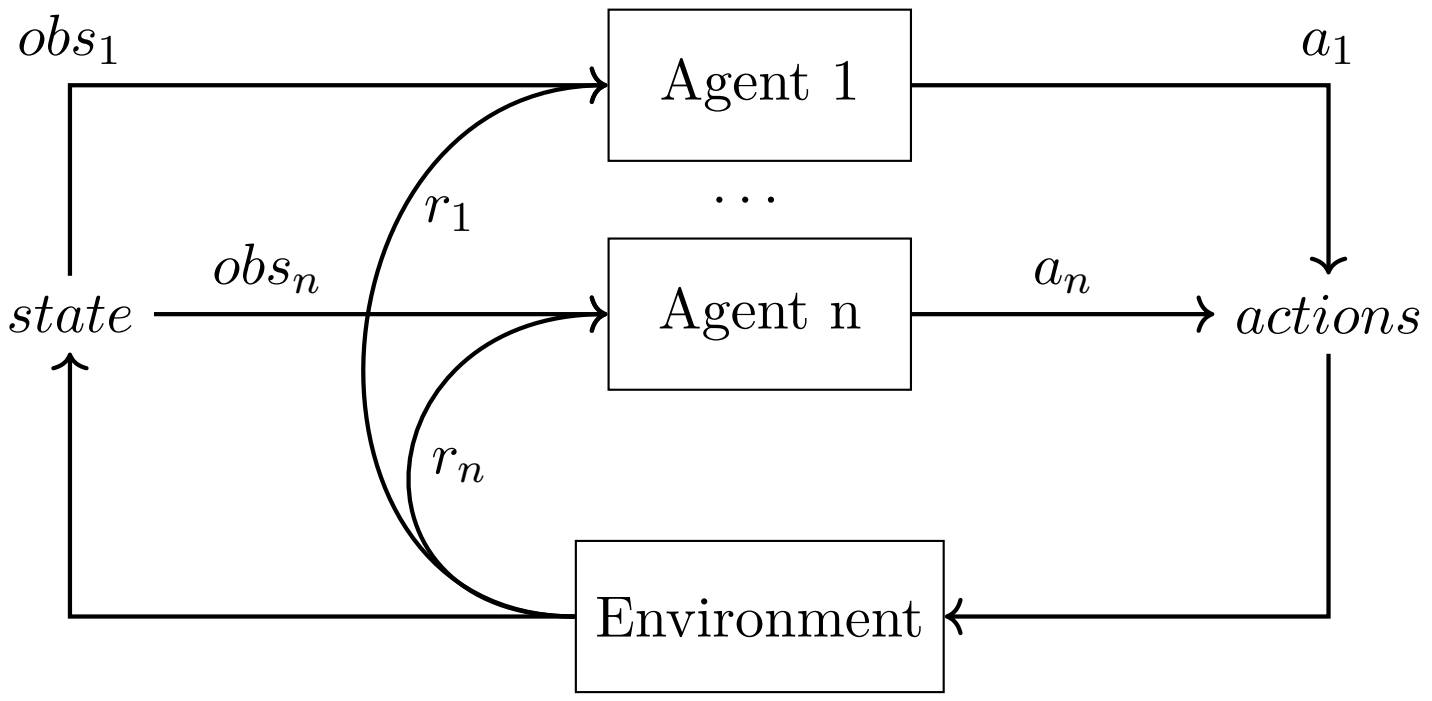
\begin{figure}[htbp] 
    \centering 
    \begin{tikzpicture}
\node (agent_1) [rectangle, draw, minimum width=2cm, minimum height=1cm, align=center] {Agent 1};
\node (agent_n) [rectangle, draw, minimum width=2cm, minimum height=1cm, below=0.5cm of agent_1] {Agent n};

\node[align=center] at ($(agent_1)!0.5!(agent_n)$) { $\dots$ };

\node (environment) [rectangle, draw, minimum width=2cm, minimum height=1cm, align=center, below=2.5cm of agent_1] {Environment};

\node (state) [left=3cm of agent_n] {$state$};
\node (actions) [right=2cm of agent_n] {$actions$};

\draw[->, thick] (environment.west) -| (state.south) ;
\draw[->, thick] (state.north) |- (agent_1.west) node[midway, above] {$obs_1$} ;
\draw[->, thick] (state.east) -- (agent_n.west) node[pos=0.25, above] {$obs_n$};

\draw[->, thick] (agent_1.east) -| (actions.north) node[midway, above] {$a_1$} ;
\draw[->, thick] (agent_n.east) -- (actions.west)  node[midway, above] {$a_n$};
\draw[->, thick] (actions.south) |- (environment.east) ;

\draw[->, thick] (environment.west) .. controls +(left:2.0cm) and +(left:2.0cm) .. (agent_1.west) node[pos=0.68, right] {$r_1$};
\draw[->, thick] (environment.west) .. controls +(left:1.6cm) and +(left:1.6cm) .. (agent_n.west) node[midway, right] {$r_n$};

\end{tikzpicture} 
    \caption{Multi-Agent Reinforcement Learning Scheme} 
    \label{fig:marl} 
\end{figure}

In contrast to RL, Multi-Agent Reinforcement Learning (MARL) enables multiple agents to learn and interact within a shared environment. Consequently, the MARL learning loop is illustrated in Figure \ref{fig:marl}. In this setting, state transitions and rewards depend on the \textbf{joint action} of all agents. In most MARL formulations, agents select their actions simultaneously; therefore, they cannot condition their decisions on the concurrent actions of others.

Since MARL operates within a Multi-Agent System (MAS), agents can interact directly with one another. This interaction enables the emergence of new mechanisms, such as coordination and communication. However, these interactions also increase system complexity, often making the learning process significantly more difficult than in single-agent settings.

    \subsection{Stochastic Games}\label{sec:stochasticgame}
    

Stochastic games (SG), introduced by \cite{shapley1953stochastic}, generalize multi-agent MDPs.
They are also known as Markov games.

Just as some RL systems do not strictly adhere to the MDP formalism, MARL problems do not always strictly adhere to the SG formalism. However, SG terminology and concepts are widely used in MARL, particularly in MARL systems where agents act simultaneously.

        
An SG is defined by five elements $(N, \mathcal{S}, \{\mathcal{A}_i \}, T, \{R_i\})$ where $N$ is the number of agents, $\mathcal{S}$ is the state space, $\mathcal{A}_i$ is the action space for an agent $i$, $T(s' | s, a_1, \ldots, a_N)$ is the transition function, and $R_i(s, a_1, \ldots, a_N, s')$ is the reward function for an agent $i$. Transitions and rewards depend on the joint action of all agents.
The interaction within an SG is similar to that of an MDP.



    \subsection{Environment and Dynamics} \label{sec:marl_dynamics}

Except in extensive-form games\footnote{This article focuses on the stochastic game framework, so we will not discuss the extensive-form games modelisation furthermore.} modelisation~\citep{kuhn1953extensive}, where agents take actions in turns, most MARL settings follow the stochastic game framework, where agents select actions simultaneously.  Therefore, the main difference between the environment in RL and MARL is that in MARL, the environment is influenced by the joint actions of all agents, not each action individually. That will change the definition domain of the dynamics functions.

For describing the environment dynamics in MARL, the state transition function and the reward function are still used. However, since agents select actions simultaneously, these functions do not have the same form, now they take as input the joint actions of all agents. 

The new definition domains of the two functions are as follows:
\begin{equation}
    T(s'|s, \textbf{a*}) : \mathcal{S} \times \mathcal{A^*} \times \mathcal{S} \rightarrow [0, 1]
    \label{eq:marl_transition_function}
\end{equation}
\begin{equation}
    R(s, \textbf{a*}, s') : \mathcal{S} \times \mathcal{A^*} \times \mathcal{S} \rightarrow \mathbb{R}
    \label{eq:marl_reward_function}
\end{equation}
Where $\mathcal{A^*} = \mathcal{A}_1 \times \mathcal{A}_2 \times ... \times \mathcal{A}_n$ is the joint action space of all agents.

When moving from single-agent RL to MARL, the dynamics function must be adapted. 
In MARL, transitions depend on joint actions. Because agents act concurrently, the outcomes cannot be computed based on individual actions without considering others'.

    \subsection{From Rewards to System Goal}\label{sec:from_rewards_to_system_goal}

As mentioned in Section \ref{sec:rl}, the objective in a single-agent system is to maximize its cumulative reward. 
Designing the reward function is crucial in RL, as it fully determines the agent's objective and how it will learn from its interactions with the environment.
The evaluation criterion in reinforcement learning is well defined and is based on reward maximization. In a multi-agent system, the question is more nuanced because the goal has to be defined at the system level \citep{zhang2021multi}. 
Increasing one agent's reward may decrease another's. The measurement of learning quality must be done at the system level, and cannot be related to the performance of only one entity. What has to be maximize to achieve the system's goal?

The modeling of the reward functions will determine the dynamics that emerge in the system. To evaluate if the resulting dynamics is good, one must choose evaluation criteria. \cite{busoniu2010} identify two dimensions for the system objectives: the search for stability and the search for adaptability. Stability is a convergence criterion that aims to achieve a stationary policy. Adaptability is a performance criterion that aims at having agents adapt their strategy to that of others. The choice of an appropriate solution concept depends on the use case and the desired outcome, some of them are presented in Section \ref{sec:evaluation_criteria}.

    \subsection{Agent}

In MARL, agents may be able to interact with each other, depending on the setting. This enable new mechanisms such as communication, model predictive of others, centralisation etc. 

This new mechanisms can be used to help the agents to be more efficient. Communication allows agents to share information and increase the potential of coordination \citep{zhu2024survey}. Predictive models of others can be used to predict the actions and therefore adapt their own action accordingly \citep{albrecht2018autonomous}. Centralisation can be used to share information, build a global model to better evaluate actions, etc \citep{sharma2021survey}.

However, these interactions can also increase system complexity and introduce or worsen new challenges, they are presented in Section \ref{sec:marl_challenges}. Agents needs to face these new challenges using the new tools at their disposal.

        \subsubsection{Self-Play \textit{vs} Mixed-Play} \label{sec:marl_selfplay_mixedplay}
        
In MARL there are two distinct approaches: self-play and mixed-play \citep{albrecht2024multi}. In self-play, agents learn by playing against others that use the same learning algorithm \citep{bowling2002multiagent, prasad2015}. This is the most common approach in MARL papers. Self-play simplifies the learning process, there is no need to design multiple algorithms. Self-play is also sometimes used in a more restricted definition, where agents learn by playing with copies of themselves \citep{tesauro1995temporal, silver2016mastering}. In this case agents share the same policy. \cite{albrecht2024multi} call them algorithm self-play and policy self-play respectively. 

In mixed-play agents learn by playing against others that use different learning algorithms. This approach is less common in MARL papers, but it may be a necessity when agents are heterogeneous. In zero-sum games, mixed-play allows one to compare the performance of different algorithms \citep{lee2022investigation}. 

Mixed-Play tends to exacerbate more the non-stationarity problem \citep{albrecht2024multi}.

    \subsection{Challenges} \label{sec:marl_challenges}

The major difficulties identified here are based on two recent articles by \cite{canese2021multi, zhang2021multi} which provide a synthesis of MARL.

        \subsubsection{Non-stationarity} \label{sec:marl_nonstationarity}

As mentioned in Section \ref{sec:rl_nonstationarity}, learning in a stationary environment makes learning easier since the dynamics do not change over time. Learning in a non-stationary environment, where the dynamics can change, is more challenging \citep{gronauer2022multi, hernandez2017survey, albrecht2024multi}. 

In MARL, the environment may be stationary in principle. However, because other agents are also learning and updating their policies, the dynamics perceived by any individual agent become non-stationary \citep{albrecht2024multi, zhang2021multi, gronauer2022multi}. Since agents do not have access to the policies of others, two states may appear identical to one agent while in reality the underlying dynamics can differ significantly due to changing strategies of others. In other words, for the same observed state of the environment, taking the same action may yield different distributions of rewards and state transitions. Therefore the optimal action to take may vary. This is why the non-stationarity is also called the \textit{moving target problem} \citep{albrecht2024multi,hernandez2017survey, tuyls2012multiagent,gronauer2022multi}.
The less agents know about each other's policies, the more pronounced this non-stationarity effect becomes.
This problem causes the loss of the Markov property \citep{gronauer2022multi}.

For example, consider a two-agent Rock-Paper-Scissors game. If agent A plays Rock with probability 1, then the best response for agent B is to play Paper with probability 1. However, as agent A updates its policy during learning, the best response for agent B will change. Agent B must also adapt its strategy accordingly. From agent B's perspective, the environment is therefore non-stationary because for the same observed state, the optimal action will vary over time.

The non-stationarity problem in MARL is not exactly the same as the non-stationarity problem in single-agent RL, where it is caused directly by the environment itself \citep{albrecht2024multi}.

        \subsubsection{Partial Knowledge of the Environment}
        
As explained in Section \ref{sec:rl_partialobs}, in RL, partial observability can already pose difficulties \ref{sec:rl_partialobs}. An agent may not have access to the full state of the environment when making observations, which deprives it of information that can sometimes be crucial for decision-making. Different strategies can be employed to mitigate this limitation, such as maintaining a history of past observations and actions \citep{igl2018deep}.

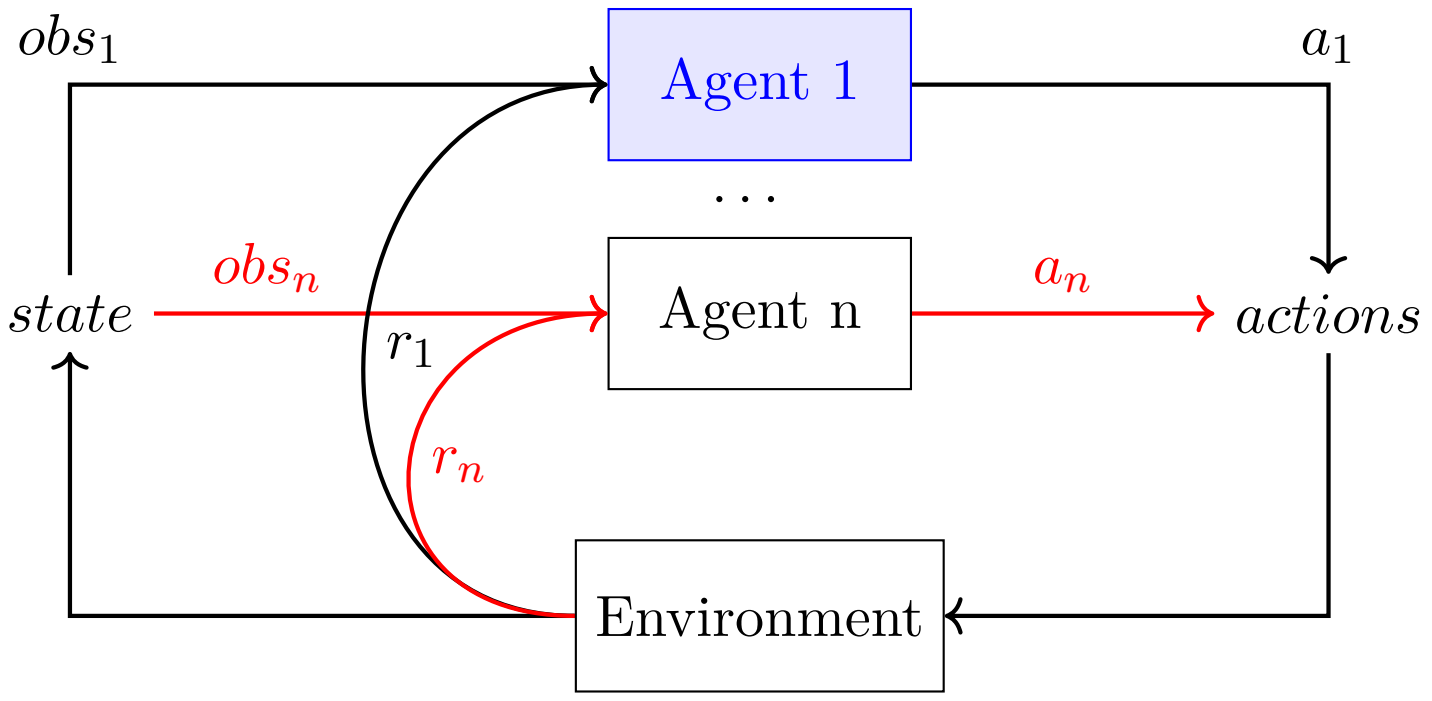
\begin{figure}
    \centering
    \begin{tikzpicture}
\node (agent_1) [rectangle, draw, blue, fill=blue!10, minimum width=2cm, minimum height=1cm, align=center] {Agent 1};
\node (agent_n) [rectangle, draw, minimum width=2cm, minimum height=1cm, below=0.5cm of agent_1] {Agent n};

\node[align=center] at ($(agent_1)!0.5!(agent_n)$) { $\dots$ };

\node (environment) [rectangle, draw, minimum width=2cm, minimum height=1cm, align=center, below=2.5cm of agent_1] {Environment};

\node (state) [left=3cm of agent_n] {$state$};
\node (actions) [right=2cm of agent_n] {$actions$};

\draw[->, thick] (environment.west) -| (state.south) ;
\draw[->, thick] (state.north) |- (agent_1.west) node[midway, above] {$obs_1$};
\draw[->, thick, red] (state.east) -- (agent_n.west) node[pos=0.25, above] {$obs_n$};

\draw[->, thick] (agent_1.east) -| (actions.north) node[midway, above] {$a_1$} ;
\draw[->, thick, red] (agent_n.east) -- (actions.west)  node[midway, above] {$a_n$};
\draw[->, thick] (actions.south) |- (environment.east) ;

\draw[->, thick] (environment.west) .. controls +(left:2.0cm) and +(left:2.0cm) .. (agent_1.west) node[midway, right] {$r_1$};
\draw[->, thick, red] (environment.west) .. controls +(left:1.6cm) and +(left:1.6cm) .. (agent_n.west) node[midway, right] {$r_n$};

\end{tikzpicture}
    \caption{Partial knowledge representation. Red arrows represent missing information for Agent 1.}
    \label{fig:partial_knowledge}
\end{figure}

Thus, partial observability is not exclusive to MARL. However, in the multi-agent setting this issue becomes more pronounced. Agents do not have access to each other's observations, rewards or policies and must choose their actions without access to the joint action at decision time \ref{fig:partial_knowledge}.

        \subsubsection{Credit Assignment}  \label{sec:marl_credit_assignment}

As mentioned in Section \ref{sec:rl_credit_assignment}, in RL, the credit assignment issue refers to the fact that an action may result in a high reward while pushing the environment into an undesirable state, or conversely, an action with little immediate reward may alter the environment in a way that leads to higher future rewards. Identifying which actions truly contributed to a received reward is therefore a non-trivial task.

In MARL, this problem becomes even more complex \citep{albrecht2024multi, gronauer2022multi}. The question is no longer only which action led to a given reward, but also which agent was responsible for it. Because the reward function depends on the joint actions of all agents, an individual agent cannot easily determine whether an improvement in its reward comes from its own policy update or from the behavior changes of others. 

For instance, consider a fully cooperative MARL scenario where all agents share the same global reward. Two agents work together to collect resources in a partially observable 2D grid. Collecting a resource gives +1 to the global reward. Suppose that at time $t$, agent A observes a resource and decides to move away from it, this action should be penalized. However, if at the same time agent B collects one resource, the global reward will be positive. Therefore, agent A may reinforce its action of moving away from the resource, even though it was detrimental to the team's objective. If agents could estimate their individual contributions to the global reward, it would prevent this issue.

        \subsubsection{Scalability} \label{sec:marl_scalability}

As explained in Section \ref{sec:rl_curse_dimensionality}, scalability can be a great challenge in RL. The size of the state-action space given by $|\mathcal{S}| \times |\mathcal{A}|$ grows exponentially with the number of states and actions.

With the multiplication of agents, scalability becomes all the more a major challenge in MARL. As the number of agents increases, the joint action space grows exponentially. For example, in a system where $N$ agents can choose among 10 possible actions, the joint action space consists of $10^N$ possible combinations of actions. Agents whose policies depend on the joint action space will therefore face significant difficulties as the number of agents increases. Consequently, some algorithms become impractical for large MARL systems.

This scalability challenge also amplifies other challenges. As the number of agents increases, the dynamics become more non-stationary, each agent's knowledge becomes more partial, and credit assignment becomes more complex.

\section{Identifying MARL Core Modules} \label{sec:marl_core_modules}

    \subsection{MARL Modular Decomposition}
    \subsubsection{Motivating a MARL Modular Decomposition}
    
When designing a MARL system, many choices have to be made depending on the use case. This article argues that these design choices should be presented in a more standardized way, by organizing MARL implementations into explicit modules.

This perspective is motivated by observations from the MARL literature. The term algorithm is often used to encompass a wide range of components and processes in MARL. When presenting their approach, articles do not always clearly specify all components of their approach. 

When it comes to explaining how it works, the organization and terms vary from one article to another. \cite{lowe2017multi} use a CTDE approach where agents are said to infer policies of others, which is referred to as agent modeling by \cite{albrecht2024multi}, modeling other agents (MOA) by \cite{jaques2019social} or opponent modeling by \cite{chernyavskiy2025applying}.

\cite{ocana2019cooperative} claim comparing IL and JAL agents. However, in this context the term JAL is the equivalent of central learning by \cite{albrecht2024multi}, while in the latter, JAL refers to agents considering joint action to select their action but while keeping their autonomy and not merging into a super agent. \cite{zhang2022neighborhood} use JAL with the same definition as \cite{albrecht2024multi}. Central learning is also named centralized RL by \cite{busoniu2006decentralized}. 

Not only do the terms vary from one article to another, but the way the approaches are presented also differs. For instance, the training and execution paradigm is clearly specified in some works \citep{shou2020reward,lowe2017multi}, but not in others \citep{ocana2019cooperative,yang2023reinforcement}.

\cite{shou2020reward} clearly specify the reward model, whereas \cite{ocana2019cooperative} refer to a cooperative reward without clarifying whether it is fully cooperative or follows another structure, and \cite{lowe2017multi} present multiple experiments but do not always specify the reward model.

Approaches that rely on communication clearly specify the communication mechanism \citep{kar2013cal, zhang2018fully, jaques2019social}. Conversely, when communication is not included, this absence is usually not explicitly stated or justified \citep{ocana2019cooperative,shou2020reward}, although there are some exceptions \citep{chernyavskiy2025applying}.

Taken together, these examples highlight the lack of a standardized way to describe MARL design choices, both in terms of terminology and in terms of the explicitness of the components involved. This makes it harder to identify which components are involved, compare approaches, and systematically explore possible solutions.

Identifying core modules and explicitly stating the choice made for each of these modules would make it easier to understand and compare approaches, even when the chosen option is the absence of an additional mechanism, such as no communication or fully decentralized execution. 
It could also help identify which choices contribute to improvements and facilitate their transfer across MARL approaches and experimental settings by reusing the corresponding core module implementations.

        \subsubsection{Identified Modules}
The core modules identified in this paper correspond to the main concerns involved in modeling, describing, and implementing MARL experiments. In the following, we present and justify the criteria used to identify them.

A MARL experiment can be described through three main dimensions:
\begin{itemize}
    \item the definition of the problem faced by the agents;
    \item the behavior of the agents, including how they learn, make decisions, and interact;
    \item the evaluation of the resulting behaviors and outcomes.
\end{itemize}
A MARL experiment must specify the problem faced by the agents and how the agents behave within it. It must also specify how the resulting behaviors and outcomes are evaluated, since evaluation tasks, protocols, and metrics are critical to the validity and comparability of MARL results \citep{papoudakis2021benchmarking,gorsane2022towards}.

Accordingly, the first criterion is that, to be considered a core module, a module must be central to the description of one of these three dimensions.

One objective of the decomposition into core modules is to provide a clearer view of the modeling choices made in MARL experiments and facilitate their comparison. Therefore, a core module should correspond to a genuine modeling choice for which multiple implementations can be defined, substituted, and studied as independent experimental factors.

Finally, to avoid defining multiple overlapping modules, core modules should be non-redundant with one another. Each module should capture a distinct aspect of the MARL experiment and have a clearly identified responsibility. This does not imply that modules are independent: interactions between them may make some combinations of module implementations incompatible.

Table~\ref{table:core_module_criteria} summarizes how each identified core module satisfies these criteria.

The implementation chosen for each core module can significantly affect the experimental outcomes and the dynamics of the underlying MARL system. Making these modules explicit therefore helps isolate and study the impact of specific design choices.

\begin{table*}[t]
    \centering
    \begin{adjustbox}{width=\textwidth,center}

\begin{tabular}{|p{4.0cm}|p{4.2cm}|p{4.2cm}|p{4.2cm}|}
\hline
\textbf{Core module} &
\textbf{Necessary dimension} &
\textbf{Modeling choices} &
\textbf{Specific responsibility} \\
\hline

Environment Dynamic \ref{sec:env_core} &
Definition of the problem faced by the agents: interaction dynamics and state of the environment &
State representations and environment dynamics &
Scheduling the agents' action and handling the Reaction phase by computing state transitions from agents' actions \\
\hline

Reward modeling \ref{sec:marl_reward_modeling} &
Definition of the problem faced by the agents: objectives and feedback signals provided through rewards &
Different reward structures &
Computing the rewards returned to agents \\
\hline

Learning \ref{sec:learning_core} &
Behavior of the agents: action selection and learning &
Many policies and learning algorithms are available &
Selecting actions and updating policies \\
\hline

Training and execution modes \ref{sec:tem_core}&
Behavior of the agents: degree of centralization of information and processes during training and execution &
Different degrees of centralization during training and execution can be selected and compared &
Specifying the information available in each phase and defines centralized process when required \\
\hline

Communication \ref{sec:communication} &
Behavior of the agents: whether and how agents communicate  &
Message content, communication structure, and message-handling mechanisms can each vary &
Managing the exchange of information between agents when communication is enabled \\
\hline

Model of Other Agents \ref{sec:moa} &
Behavior of the agents: whether other agents are modeled, and how such models are represented, learned, and used &
MOA can vary in its representation and learning method &
Determining whether and how agents model others, and for what purposes, such as action prediction \\
\hline

Agent \ref{sec:core_agent} &
Behavior of the agents: agent decision-making and interaction with the environment &
Alternative ways of orchestrating internal modules and carrying agent-level responsibilities &
Orchestrating the different modules used by an agent \\
\hline

Evaluation criteria \ref{sec:evaluation_criteria} &
Evaluation of the resulting behaviors and outcomes &
Alternative measures of performance, fairness, convergence, and other properties &
Measuring specific aspects of system behavior and outcomes according to the objectives \\
\hline
\end{tabular}

    \end{adjustbox}
    \caption{Core modules and their correspondence with the identification criteria.}
    \label{table:core_module_criteria}
\end{table*}

This Section identifies eight core modules that should be specified when implementing a MARL system: (1) agent, (2) environment dynamics, (3) learning, (4) reward modeling, (5) training and execution modes, (6) communication, (7) model of other agents (MOA), and (8) evaluation criteria. 

For each module implementation presented, we provide a definition, discuss the motivation behind its use, identify the challenges it addresses or is most affected by, and, when relevant, give illustrative examples.

        \subsubsection{Potential Modules Not Retained}

Several potential modules were considered but ultimately rejected because they did not satisfy all the identification criteria. The following paragraphs discuss some examples to illustrate how these criteria were applied.

Several potential modules were considered but ultimately rejected because they did not satisfy all the identification criteria. The following paragraphs discuss two examples to illustrate how these criteria were applied. For each potential module, we provide its definition, the rationale for considering it, and the reasons for its exclusion.

Not being retained as a core module in the conceptual decomposition of MARL does not imply that a concept is unimportant or that it should not be represented as a dedicated component in a software implementation. The core module criteria are intended to provide a conceptual decomposition of MARL solutions and experiments, rather than prescribe a software architecture.

\paragraph{Scheduler}
The scheduler is the component that orchestrates all the other components. It has a specific responsibility. In the literature, we mainly identified two types of scheduling: agents either select their actions simultaneously or make decisions sequentially. However, the choice of scheduling mechanism is completely determined by the environment dynamics. More specifically, the scheduler must call agents simultaneously when state transitions depend on their joint actions, and sequentially when each transition depends on the action of a single agent. Therefore, we did not include the scheduler as a separate component in our core module decomposition, since its behavior is entirely determined by the environment dynamics\footnote{Although it is not retained as a core conceptual module, the scheduler is represented as a dedicated class in our software implementation.}.

\paragraph{Exploration Strategy}
The exploration \textit{vs.} exploitation dilemma is an important question in both RL and MARL, and different exploration strategies exist. However, exploration strategies are too tightly coupled with the learning module to constitute an independent core module and are most commonly integrated into the policy, which maps inputs to actions.

\paragraph{Learning Parameter Adaptation} 
Policies and learning algorithms have parameters. It is possible to use approach that adapt learning parameter based on the dynamics.  
Several approaches, such as WoLF and hysteretic learning, exist and can improve learning. 
However, these techniques apply only to specific learning mechanisms. They are therefore too specialized and tightly coupled with the learning module to constitute an independent core module.

\paragraph{Neural Network Architecture} 
Neural networks architectures can take many different forms. 
Some components rely on neural networks, whose architecture can vary and affect the results. However, providing mechanisms for selecting or switching between neural network architectures is primarily a software design concern. The neural network model is therefore considered an internal aspect of the learning module rather than an independent core module.

    \subsection{Environment Dynamic}\label{sec:env_core}
    
The environment is a central module of MARL systems. Agents interact and evolve within the environment, whose dynamics determine how the system changes over time. Among the components of the environment, the dynamics are the most implementation-dependent, since they define how the environment reacts to agents' actions.

The environment also defines how agents are scheduled, namely whether they select their actions simultaneously or in a sequential order.

As explained in Section~\ref{sec:marl_dynamics}, the transition function must be computed based on joint actions of all agents (not each action individually). While such a perspective is widely admitted, its conceptual consequences are rarely translated into explicit modeling guidelines, and are often handled in an ad-hoc manner.

The need to manage the transition depending on multiple agents is a problem that already exist in the MAS literature. To capture this, the model of the environment we use is based on IRM4S (Influence Reaction Model for Simulation)~\citep{ferber1996influences}. Since the dynamics depend on joint actions, the environment reaction should be computed separately from the phase in which agents choose and perform their actions. In this paper, we make this separation explicit by relying on the Influence/Reaction principle, a two-phase modeling paradigm introduced in the MAS literature. This separation helps preserve the independence of the environment module by keeping agents' intended actions distinct from the computation of their effects.

The Influence/Reaction principle offers a paradigm for modeling MAS in which agents never alter the environment directly. Instead, they generate influences that express their intended actions, while the environment subsequently reacts to the complete set of influences produced simultaneously. This approach clearly separates the process into two complementary phases: an Influence phase, where agents and the environment emit their influences, and a Reaction phase, during which the system combines these influences and computes the resulting state~\citep{michel07irm4sShort}. The dynamics of the environment then change accordingly to the joint actions, namely the influences, of all agents. Allowing the agents to directly modify the environment may lead to conflicts and will be in opposition to the stochastic game framework where the dynamics depend on the joint action.

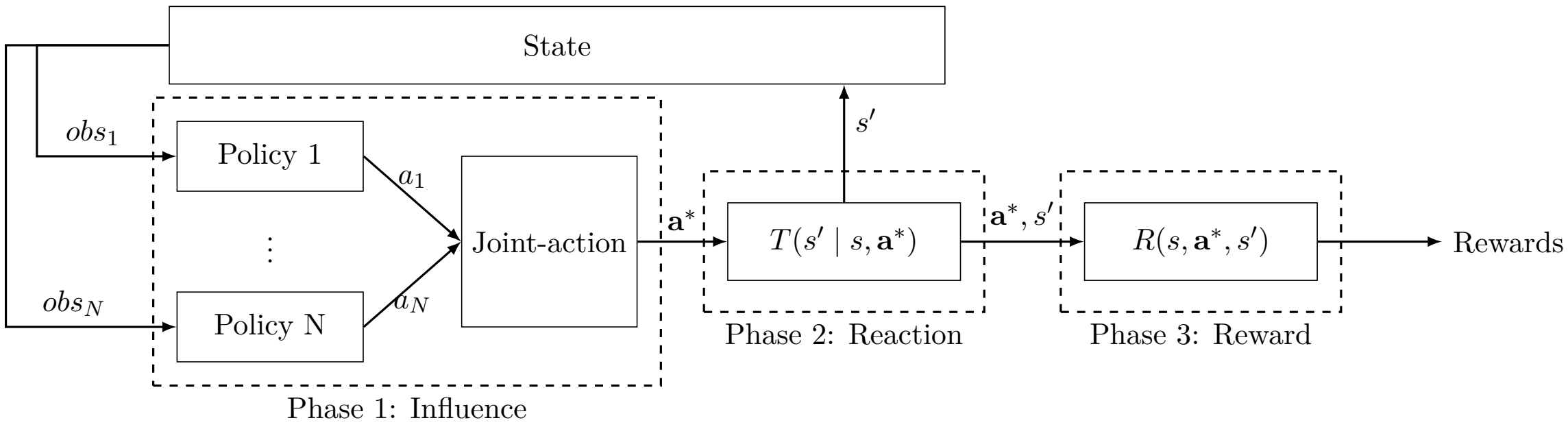
\begin{figure}[ht]
    \centering
    \resizebox{\linewidth}{!}{\begin{tikzpicture}[>=latex]

\node (policy1) [rectangle, draw, minimum width=2.4cm, minimum height=0.9cm, align=center] at (-4.0,-1.8) {Policy 1};
\node (policydots) [align=center] at (-4.0,-2.9) {$\vdots$};
\node (policyN) [rectangle, draw, minimum width=2.4cm, minimum height=0.9cm, align=center] at (-4.0,-4.0) {Policy N};

\node (joint) [rectangle, draw, minimum width=1.8cm, minimum height=2.2cm, align=center] at (-0.4,-2.9) {Joint-action};

\node (T) [rectangle, draw, minimum width=3.0cm, minimum height=1.0cm, align=center] at (3.4,-2.9) {$T(s'\mid s, \textbf{a}^*)$};
\node (R) [rectangle, draw, minimum width=3.0cm, minimum height=1.0cm, align=center] at (8,-2.9) {$R(s, \textbf{a}^*, s')$};

\node (state) [rectangle, draw, minimum width=10cm, minimum height=1cm, align=center]
	at ($ (policy1.north)!0.5!(T.north) + (0,1.5cm) $) {State};

\draw[->, thick] (state.west) -- ++(-1.7cm,0) |- (policy1.west) node[pos=0.7, above] {$obs_1$};
\draw[->, thick] (state.west) -- ++(-2.1cm,0) |- (policyN.west) node[pos=0.7, above] {$obs_N$};

\draw[->, thick] (policy1.east) -- (joint.west) node[midway, above] {$a_1$};
\draw[->, thick] (policyN.east) -- (joint.west) node[midway, below] {$a_N$};

\draw[->, thick] (joint.east) -- (T.west) node[midway, above] {$\textbf{a}^*$};

\draw[->, thick] (T.north) -- ($(state.south)+(3.7cm,0)$) node[pos=0.7, right] {$s'$};
\draw[->, thick] (T.east)  -- (R.west) node[midway, above] {$\textbf{a}^*, s'$};

\draw[->, thick] (R.east) -- ++(1.6cm,0) node[right] {Rewards};

\node (ibox) [draw, dashed, thick, fit=(policy1)(policyN)(joint), inner xsep=0.3cm, inner ysep=0.3cm] {};
\node[anchor=north, align=center] at (ibox.south) {Phase 1: Influence};

\node (rbox) [draw, dashed, thick, fit=(T), inner xsep=0.3cm, inner ysep=0.4cm] {};
\node[anchor=north, align=center] at (rbox.south) {Phase 2: Reaction};

\node (rewbox) [draw, dashed, thick, fit=(R), inner xsep=0.3cm, inner ysep=0.4cm] {};
\node[anchor=north, align=center] at (rewbox.south) {Phase 3: Reward};

\end{tikzpicture}}
    \caption{Influence/Reaction principle: agents produce influences that are combined into a joint action, which is used to compute the environment reaction (state transition). Then the rewards are compute in the reward phase.}
    \label{fig:influence_reaction}
\end{figure}

Figure~\ref{fig:influence_reaction} illustrates the Influence/Reaction principle and its separation between the Influence and Reaction phases. In the MARL setting considered here, this process is followed by an additional phase dedicated to computing agents' rewards. This third phase, called the Reward phase, is executed after the Reaction phase.

Unlike other core modules, whose implementations may introduce mechanisms to support or modify agents' behavior, the environment primarily defines the situation faced by the agents, including the constraints and difficulties they must handle.

    \subsection{Reward Modeling} \label{sec:marl_reward_modeling}
    
The reward modeling is more difficult to design in MARL than in RL because one has to compute a reward value for each agent, in a way that can drive the whole system toward its goal.
The modeling of the reward function in a MAS is a critical aspect that determines the behavior of the agents in the environment. 

In the same environment, variations in reward functions can cause agents to engage in either intense competition or fully cooperative behavior. For instance, \cite{tampuu2017multiagent} experiments with different reward models for its system and shows that with the same environment, they affect the dynamics of the system and how agents learn to act. 

Although there are infinite numbers of ways to define the reward function, they can be grouped into three main categories: (1) fully cooperative, (2) fully competitive, and (3) mixed \citep{liang2025review, canese2021multi}.

        \subsubsection{Fully Cooperative} \label{sec:fully_cooperative}
        
\paragraph{Definition}
A MARL system is said to be fully cooperative\footnote{Can also be called team game or common-reward game in game theory \citep{yang2020overview, zhang2021multi}.} when all agents share a unique global reward: $\forall (i,j) \in [1, N], R_i = R_j$. In this case, the agents should work together to maximize the global reward \citep{liang2025review, canese2021multi}. 

\paragraph{Motivation}
Whenever agents optimize non-identical reward functions, conflicts of interest may emerge, even in settings intended to be cooperative. A fully cooperative formulation, where all agents share the same global reward, removes this misalignment of incentives. In that case, an agent cannot improve its own return without also improving the others', which prevents individually beneficial yet globally suboptimal behaviors.

\paragraph{Example}
Consider a foraging task in which two agents are tasked with collecting resources. The episode ends after all resources are collected or a maximum number of timesteps is reached. Let's consider this experience with two different reward function designs:
\begin{itemize}
    \item \textbf{Fully Cooperative Reward:} All agents receive the same global reward: a positive reward is given whenever any agent collects a resource, and a negative reward, representing time penalty, is given at each timestep.
    \item \textbf{Individual Reward:} Each agent receives positive rewards for collecting a resource and negative rewards every timestep. 
\end{itemize}
In the first case, the optimal policy for the agents is to divide the resources among themselves to minimize overall travel time and reduce the time to collect all the resources. However, in the second case, there is a balance to strike between collecting resources and minimizing the global time to collect all the resources. In specific situations, it may be optimal for an agent to collect all the resources, prioritizing the one close to the other agent, even if it increases the global time to collect all the resources.

In MARLKIT, there is a foraging experiment that illustrates this situation (see Section \ref{sec:marlkit_reward_modeling})

\paragraph{Challenges}
In this setting, the objective is straightforward: optimize the global reward. This objective is close to single-agent RL, which is one-dimensional. 

The main drawback of this approach is related to the credit assignment problem \citep{albrecht2024multi, ning2024survey, gronauer2022multi}. It is harder for an agent to improve its policy because it cannot identify which actions contributed to the global reward. As the number of agents increases, the credit assigned to an agent's individual actions becomes diluted.

        \subsubsection{Fully Competitive} \label{sec:fully_competitive}
        
\paragraph{Definition}
A MARL system is said to be fully competitive\footnote{Can also be called constant-sum or zero-sum game in game theory \citep{yang2020overview, zhang2021multi}.} when the sum of all agents' rewards is constant: $\sum_{i=1}^{N}R_i = C$, where $C$ is a constant and $N$ is the number of agents \citep{liang2025review, canese2021multi}. When the constant is zero ($C=0$), the system is said to be a zero-sum game.

\paragraph{Motivation}
Consequently, if one agent increases their reward, the others' rewards decrease by the same amount. In this setting, agents have completely opposed interests. This is a common setting in 1v1 competitive games, such as chess or Rock-Paper-Scissors.

\paragraph{Example}
In this setting, agents are incentivized to compete against each other. The dynamics that emerge from this setting are different from the ones that emerge in other settings. \cite{tampuu2017multiagent} show that in a simple 1v1 Pong game, the reward modeling affects the dynamics of the game and how agents learn to play. In a fully competitive setting, agents learn to make angled shots that bounce off the walls. The number of paddle-bounces per point is lower than in other settings.

\paragraph{Challenges}
In this setting, agents' rewards are strictly opposed and thus completely dependent on other agents' actions. This exacerbates the non-stationarity problem \citep{lowe2017multi}, as each agent's optimal action is highly sensitive to the actions of other agents.

        \subsubsection{Mixed} \label{sec:reward_modeling_mixed}
        
\paragraph{Definition}
The reward modeling is considered mixed when it is neither fully cooperative nor fully competitive\footnote{Can also be called general-sum game in game theory \citep{yang2020overview, zhang2021multi}} \citep{liang2025review, canese2021multi}.

\paragraph{Motivation}
Mixed systems can exhibit varying degrees of cooperative or competitive behavior \citep{liang2025review}. 
In fully cooperative and fully competitive settings, agents' rewards exhibit a strong correlation, reflecting the aligned or strictly opposed nature of their objectives. In contrast, in mixed settings, the correlation between agents' rewards can be weaker or reflect more complex interdependencies among their objectives.

\paragraph{Example}
A simple example of a mixed game oriented toward competition is a resource collection game, in which agents should collect resources to maximize their own reward. Collecting a resource increases an agent's reward without directly decreasing the others' rewards. However, it decreases the amount of resources available to them. An example of a cooperatively oriented game is the robovac scenario presented above, with reward functions that are not shared. 

\paragraph{Challenges}
The credit assignment problem is due to agent reward being affected by other agents' actions \citep{gronauer2022multi}. Therefore, the less an agent's reward is affected by the actions of other agents, the easier it is for the agent to identify which actions contributed to its reward. This is the reason why the credit assignment problem is important in fully cooperative settings \citep{albrecht2024multi}. 
In fully competitive settings, agents share the same reward function, in fully competitive settings, one agent's reward function is fully determined by the others' reward functions. In mixed settings, the correlation between agents' rewards function can be weaker. Less correlation between agents' rewards and others' actions can mitigate the credit assignment problem.

The non-stationarity problem is due to the fact that an agent's optimal action is affected by other agents' actions (see Section~\ref{sec:rl_nonstationarity}). Because the policy of other agents changes over time, the optimal action for an agent can change over time, then the agents face a moving target problem \citep{gronauer2022multi, hernandez2017survey}. Therefore, the fewer an agent's reward is affected by the actions of other agents, the less non-stationarity issues the agent will face. Depending on the reward model chosen, the non-stationarity problem can be less problematic than in fully cooperative or fully competitive settings. However, even if the reward function of an agent is only affected by its actions and the current state, the transition function of the environment is still affected by the actions of other agents, then the non-stationarity problem is still present even in a completely independent reward setting.

    \subsection{Learning} \label{sec:learning_core}
    
The learning module is responsible for managing agents' decision-making and learning processes. It consists of two sub-modules that are too closely tied to be considered as independent modules: learning algorithm and policy.

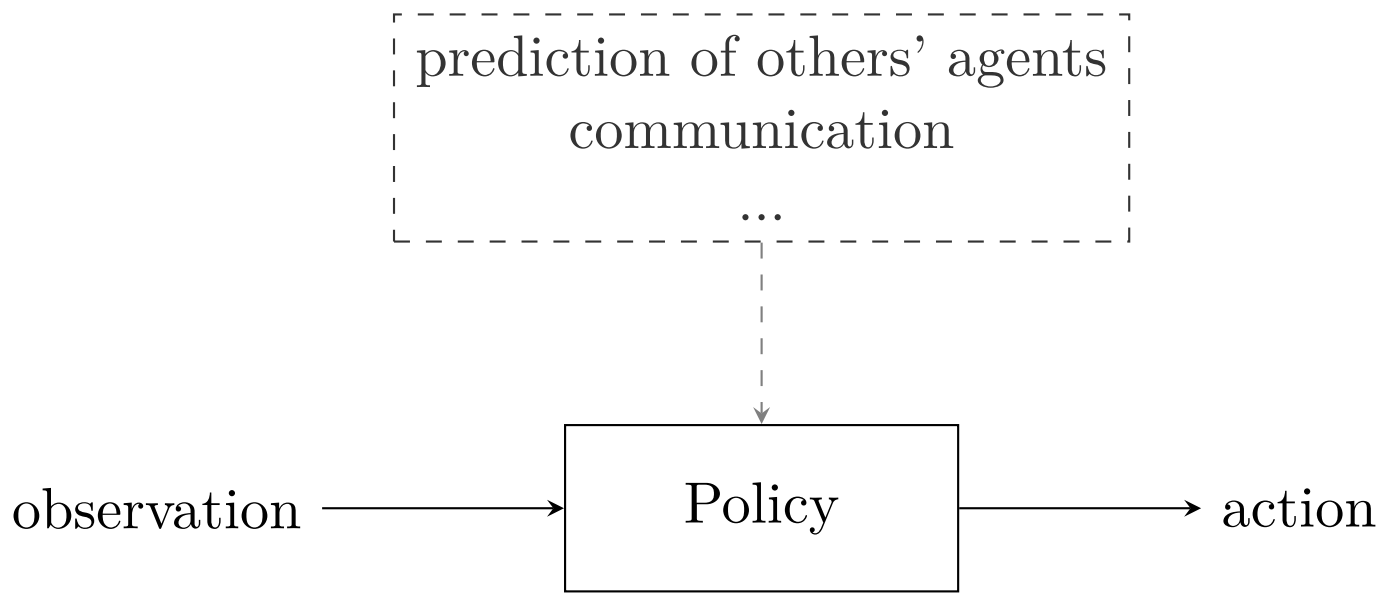
\begin{figure}[h]
	\centering
	\begin{tikzpicture}[>=stealth]
	\node[draw, rectangle, minimum width=2.6cm, minimum height=1.1cm] (policy) {Policy};
	\node[left=1.6cm of policy] (obs) {observation};
	\node[right=1.6cm of policy] (act) {action};
	\node[above=1.2cm of policy, align=center, text=black!80, draw=black!80, dashed, rectangle, inner sep=4pt] (pred) {prediction of others' agents\\communication\\...};

	\draw[->] (obs) -- (policy);
	\draw[->] (policy) -- (act);
	\draw[->, dashed, gray] (pred) -- (policy);
\end{tikzpicture}
	\caption{policy can now take other input such as prediction of other agent action, or information that came from communication}
	\label{fig:marl-policy-extended-input}
\end{figure}

The policy is still a function that takes as input the observation of the agent and outputs an action. However, now the policy may also take as input other information. Different approach in MARL consist in considering other information before selecting an action, such as communication from other agents, or a prediction of the action of other agents. Then the policy function will have to be adapted to take this new information as input. Figure~\ref{fig:marl-policy-extended-input} illustrates this idea, where the policy can now take other input.

Even if the input may differ, the different types of policies presented in Section~\ref{sec:rl_policy} remains relevant in MARL. 

MARL implementations can use single-agent RL learning algorithms. However, MARL learning algorithms are often more complex: although many of them build upon single-agent RL algorithms, they include additional mechanisms to account for interactions between agents.

The core RL algorithmic categories introduced in Section~\ref{sec:rl_learning_module} remain relevant in MARL. However, some algorithmic aspects require particular attention in multi-agent settings. This is especially the case for learning parameters, whose adaptation can help address the non-stationarity induced by simultaneously learning agents. The following subsections present several MARL techniques based on adaptive learning parameters.

In stationary single-agent RL settings, the transition and reward functions are fixed. Consequently, the optimal value function and the set of optimal policies are also fixed, providing the agent with a stable target toward which it can converge. In this context, learning parameters, such as the learning rate or exploration rate, are often decreased over time to progressively stabilize learning and support convergence toward an optimal behavior.

In a multi-agent system, agents learn simultaneously. Moreover, the optimal action for a given agent depends on the actions of the other agents, and therefore on their policies. As a result, agents must update their policies with respect to those of others, which are themselves evolving. This gives rise to the non-stationarity problem discussed in Section \ref{sec:marl_nonstationarity}. To address this issue and support policy convergence, one algorithmic approach is to use adaptive learning parameters.

Adaptive learning parameters approaches are more commonly used in MARL than in single-agent RL. Learning parameters can also vary in single-agent RL, but these variations are usually predefined rather than driven by interactions with the environment.

The following subsections present different adaptive learning-parameter techniques used in MARL.

        \subsubsection{Win or Learn Fast (WoLF)} 

\paragraph{Definition}
In the Win or Learn Fast (WoLF) approach \citep{bowling2002multiagent}, the learning rate is adapted according to whether the agent is winning or losing: when winning, the agent learns slowly; when losing, it learns faster in order to improve its performance.

\paragraph{Motivation} 
The main motivation behind WoLF is to promote rational learning behavior while improving the stability and convergence properties of the overall system.
 
From the perspective of a fast-learning agent, slower-learning agents have more stable and predictable policies, which makes learning easier for fast-learning agents.

\paragraph{Examples}
Several implementations of WoLF have been proposed. For example, WoLF-PHC (WoLF-Policy Hill Climbing) \citep{bowling2002multiagent}, WoLF-IGA (WoLF-Infinitesimal Gradient Ascent) \citep{bowling2002multiagent}, and GIGA-WoLF (Generalized Infinitesimal Gradient Ascent-WoLF) \citep{bowling2004convergence}.

WoLF-PHC has been applied in Rock-Paper-Scissors game, where two agents learn to play against each other \citep{bowling2002multiagent}. WoLF-PHC is compared to PHC. The results show that WoLF-PHC converges to the Nash equilibrium, while PHC does not. 

\paragraph{Challenge}
Because slower-learning agents have more stable and predictable policies, faster-learning agents face a less severe form of non-stationarity. Therefore, WoLF helps reduce the non-stationarity problem \citep{bowling2002multiagent, bowling2004convergence}.

        \subsubsection{Two-Timescale (2TS)}
\paragraph{Definition}
The 2TS approach, originally introduced by \cite{borkar97Stochastic}, consists of defining two different learning rates for two processes\footnote{We use the term "process" on purpose as a broad label: the two timescales may apply either to different agents, or to different learning components within one agent (for example, actor and critic).}, so that one process evolves much faster than the other\footnote{This threshold is formally defined by \cite{leslie2003}.}. 

\paragraph{Motivation} 
From the perspective of one agent, non-stationarity is caused by other agents updating their policies. Consequently, the optimal policy for this agent keeps changing as its learning progresses. 

The key idea of 2TS is that when one process evolves much faster than the other, the faster process sees the slower one as nearly stationary, which helps it converge toward an appropriate response. 
Conversely, from the slower timescale, the faster process can be viewed as being close to its response to the current state of the slower process, since it adapts much more rapidly.

In this sense, the motivation behind 2TS is close to that of WoLF: both approaches rely on asymmetric learning speeds to improve stability in multi-agent learning.

\paragraph{Examples}
Several implementations of 2TS have been proposed. \cite{leslie2003} generalize 2TS to $n$ agents by assigning a different timescale to each agent. In this work, agent $1$ learns slower than agent $2$ which learns slower than agent $3$ and so on. As a result, faster-learning agents can adapt their policies to slower-learning ones.

The 2TS approach can also be applied to different learning components within an agent. \cite{perkins2015mixed} use 2TS in an Actor Critic algorithm to separate Critic (fast) and Actor (slow). In this second example, all agents learn at the same overall pace but the critic learns faster than the actor, leading to a more stable evaluation of the current policy.

The 2TS approach was applied in a Stick Together Game (STG), where two agents have to learn to group and stay together \citep{prasad2015}. 2TS is compared to two other algorithms: Nash Q-learning and Friend-or-Foe Q-learning, two MARL variants of Q-learning. The results show that 2TS outperforms these two algorithms in terms of convergence and performance.

\paragraph{Challenge}
By separating learning dynamics across different timescales, 2TS makes some processes appear nearly stationary from the perspective of others. In this way, it helps mitigate the non-stationarity problem \citep{borkar2002reinforcement,leslie2003, perkins2015mixed}.

        \subsubsection{Hysteretic Learning Rates}
\paragraph{Definition}
Hysteretic learning rates consist of using two different learning rates: $\alpha$ for $Q$-value increases and $\beta$ for $Q$-value decreases, with $\alpha > \beta > 0$.

\paragraph{Motivation}
The same action can lead to a high or low reward, depending on the joint actions of all agents. Therefore, an agent can receive a low reward even when it takes the optimal action, because others are exploring and taking suboptimal actions. 

Hysteretic learning rates are designed to address this issue by using two different learning rates, so agents remain optimistic about beneficial experiences and are less misled by outcomes induced by others' exploration.

\paragraph{Examples}

This approach was first introduced by \cite{matignon2007hysteretic}, with the Hysteretic Q-learning (HQL) algorithm. It has since been combined with other algorithms, such as DQN (HDQN) \citep{zhang2022neighborhood}.

HQL was applied in a prey-predator game with partial observability \citep{matignon2007hysteretic}. In this setting, four predators have to coordinate to catch a prey. HQL is compared to Distributed Q-learning (DQL, defined by \cite{lauer2000algorithm}). The results show that HQL outperforms DQL.

\paragraph{Challenge}
By reducing the impact of negative updates caused by other agents' exploration, hysteretic learning rates prevent agents from prematurely discarding actions that may be beneficial under different joint actions. Therefore, hysteretic learning rates make learning more robust to the variability induced by simultaneously learning agents and help mitigate one aspect of the non-stationarity problem.

    \subsection{Training and Execution Modes} \label{sec:tem_core}
    
In MARL, a distinction can be made between the training phase, during which agents update their policies, and the execution phase, during which agents use these policies to act in the environment. In some settings, these two phases are merged: agents are evaluated while they are still learning. In other settings, they are separated: agents are first trained, and their performance is then evaluated during a distinct execution phase.

This distinction is important because agents may not have access to the same information during training and execution. For example, centralized information may be available during training to facilitate learning, while agents may have to act only from local information during execution. Training and execution modes therefore define how centralized or decentralized the agents are in each phase.

In many use cases, the training and execution phases are distinct, allowing for different approaches in each phase. There are three main modes for training and execution in MARL: Centralized Training and Execution (CTE), Decentralized Training and Execution (DTE), and Centralized Training Decentralized Execution (CTDE).

        \subsubsection{CTE} \label{sec:cte}
        
\paragraph{Definition}
CTE implies that agents are trained and executed using centralized information \citep{zhang2021multi, albrecht2024multi}. This means that during both training and execution, agents should have access to a central process that can aggregate information from all agents and communicate it back to them. These information can include observations, environment dynamics, other agent models... They can keep their autonomy, but they interact with a central entity.

But CTE is sometimes used to refer to a more restrictive definition. In this case, all agents are interacting with a central process that pilot the joint policy of all agents \citep{gronauer2022multi, wong2023deep}. This is sometimes referred as Central Learning \citep{albrecht2024multi}. Using this is transforming a multi-agent problem into a single-agent one. In this case, agents are not autonomous.

\paragraph{Motivation}
CTE allows agents to access more information during training and execution, which can facilitate learning and coordination \citep{albrecht2024multi,sharma2021survey}. 

\paragraph{Challenges}
Increasing knowledge of other agents tends to reduce non-stationarity issues. CTE then helps to mitigate this difficulty \citep{sharma2021survey, amato2024introduction}. For the same reason, it also helps to reduce the partial knowledge difficulty and could reduce credit assignment issues depending on the shared information \citep{albrecht2024multi}.
However, CTE can be impractical because it worsens the scalability issue \citep{gronauer2022multi, albrecht2024multi}.

        \subsubsection{DTE}

\paragraph{Definition}
In DTE, agents are trained and executed using only their own local information. There is no central entity to aggregate information or coordinate actions. 

\paragraph{Motivation}
This approach requires fewer agents-coordination mechanisms. This is a natural choice when the agent cannot be trained or executed in a centralized setting \citep{albrecht2024multi}. It can be more realistic for many applications where agents operate independently because a central entity cannot coordinate them \citep{zhang2021multi, albrecht2024multi}.

\paragraph{Challenges}
In this setting, the non-stationarity problem is at its peak \citep{amato2024introduction}, and becomes more severe if agents' rewards are highly correlated with others' actions. However, DTE can be more scalable because it does not rely on joint observation, or joint action spaces which grow exponentially with the number of agents \citep{albrecht2024multi}.

        \subsubsection{CTDE} \label{sec:ctde}
        
\paragraph{Definition}
The execution phase can involve numerous constraints and hidden information (about other agents' policies, about the environment, etc). This hinders learning, especially in cooperative settings where agents would benefit from freely sharing all this information. The CTDE idea is to provide agents with additional information during training so that they can learn better policies, and then execute these policies in a decentralized manner \citep{amato2024introduction}. The first instance of CTDE was proposed by \cite{kraemer2016multi}. 

\paragraph{Motivation}
CTDE allows agents to exchange and access substantially more information than they will have during execution. This additional information facilitates learning by making it easier for agents to assess the consequences of their actions. CTDE is designed to combine the benefits of both CTE and DTE \citep{albrecht2024multi}.

\paragraph{Examples}
There are different versions of CTDE. This section presents two. The most common being the combination of CTDE with Actor Critic algorithms, in which a centralized critic is used during training while the actors are decentralized \citep{lowe2017multi, amato2024introduction}. The motivation is to obtain a better-trained critic that can mitigate the non-stationarity problem.

Parameter sharing is another technique used by \cite{gupta2017cooperative} that can be used for centralized training. It consists of training a shared policy using the experiences collected by all agents. \cite{gupta2017cooperative} argue that this technique increases training efficiency.

\paragraph{Challenges}
CTDE requires a training phase, and if training and execution are not separated, it cannot be applied. This condition makes it sometimes less realistic than DTE where agents are trained independently \citep{jaques2019social}. 

Some CTDE techniques, such as parameter sharing, requieres agents to be homogeneous \citep{gupta2017cooperative}.

For the same reasons as CTE, by increasing knowledge of other agents, CTDE tends to reduce non-stationarity issues \citep{gronauer2022multi, amato2024introduction, ning2024survey}, helps to reduce the partial knowledge difficulty and could reduce credit assignment issues depending on the shared information.%
However, centralized training itself raises scalability issues: as the number of agents increases, the training process becomes progressively more expensive and complex \citep{amato2024introduction}.

    \subsection{Communication} \label{sec:communication}
    
As with training and execution modes, selecting a communication method is a design choice that shapes the ways in which agents interact. 

The messages exchanged between agents can take various forms. Agents can share their observations, communicate their intentions, share value functions\etc \citep{albrecht2024multi, wong2023deep}. The content of the messages exchanged can also be learned \citep{zhu2024survey}.

By sharing their observations, agents can have a better understanding of the environment state, which helps mitigate the partial observability challenge. By sharing their intentions or estimation of the value of actions, agents can better predict the behavior of others, helping to mitigate the challenge of non-stationarity \citep{wong2023deep, gronauer2022multi}.

The main drawback of communication is the potential increase in computational complexity and scalability concerns. The impact depends heavily on the chosen communication mode.

Several communication modes exist. This section presents the main categories. For a more detailed survey of communication methods in multi-agent systems, readers are referred to \cite{zhu2024survey}.

\begin{figure}
    \centering
    \begin{subfigure}{0.3\textwidth}
        \centering
        \resizebox{\linewidth}{!}{\begin{tikzpicture}[node distance=1.8cm]

\node (center) [align=center] {};

\node (b) [circle, draw, below of=center, align=center] {Agent};
\node (l) [circle, draw, gray, left of=center, align=center] {Agent};
\node (r) [circle, draw, gray, right of=center, align=center] {Agent};
\node (a) [circle, draw, gray, above of=center, align=center] {Agent};

\draw[->, thick, lightgray] (l) -- (b);
\draw[<->, thick, lightgray] (l) -- (r);
\draw[<->, thick, lightgray] (l) -- (a);

\draw[->, thick, lightgray] (r) -- (b);
\draw[<->, thick, lightgray] (r) -- (a);

\draw[->, thick, lightgray] (a) -- (b);

\draw[->, thick] (b) -- (l);
\draw[->, thick] (b) -- (r);
\draw[->, thick] (b) -- (a);

\end{tikzpicture}}
        \caption{Broadcasting communication.}
        \label{fig:broadcast_comm}
    \end{subfigure}
    \hfill
    \begin{subfigure}{0.3\textwidth}
        \centering
        \resizebox{\linewidth}{!}{\begin{tikzpicture}[node distance=2.5cm]

\node (center) [circle, draw, align=center] {Agent};

\node (ul) [circle, draw, above left of=center, align=center] {Agent};
\node (ur) [circle, draw, above right of=center, align=center] {Agent};
\node (bl) [circle, draw, below left of=center, align=center] {Agent};
\node (br) [circle, draw, below right of=center, align=center] {Agent};

\draw[<->, thick] (center) -- (ul);
\draw[<->, thick] (center) -- (ur);
\draw[<->, thick] (center) -- (bl);
\draw[<->, thick] (ur) -- (br);
\draw[<->, thick] (ul) -- (bl);
\end{tikzpicture}}
        \caption{Networked communication.}
        \label{fig:networked_comm}
    \end{subfigure}
    \hfill
    \begin{subfigure}{0.3\textwidth}
        \centering
        \resizebox{\linewidth}{!}{\begin{tikzpicture}[node distance=2.5cm]

\node (center) [circle, draw, align=center] {Agent};

\node (ul) [circle, lightgray, draw, above left of=center, align=center] {Agent};
\node (ur) [circle, lightgray, draw, above right of=center, align=center] {Agent};
\node (bl) [circle, lightgray, draw, below left of=center, align=center] {Agent};
\node (br) [circle, lightgray, draw, below right of=center, align=center] {Agent};

\draw[->, thick] (center) -- (ul);
\draw[->, thick] (center) -- (bl);
\draw[->, lightgray, thick] (br) -- (ur);
\draw[->, lightgray, thick] (ur) -- (center);
\draw[->, lightgray, thick] (ur) -- (ul);

\end{tikzpicture}}
        \caption{Targeted communication.}
        \label{fig:targeted_comm}
    \end{subfigure}
    \caption{Comparison of communication modes: broadcasting, networked, neighborhood and targeted. Broadcasting communication (a) involves all agents sending messages to all others. Networked communication (b) involves agents communicating only with their direct neighbors in a predefined network structure. 
    Targeted communication (c) allows agents to send messages to specific other agents. Figures adapted from \cite{gronauer2022multi}.}
    \label{fig:communication_modes}
\end{figure}
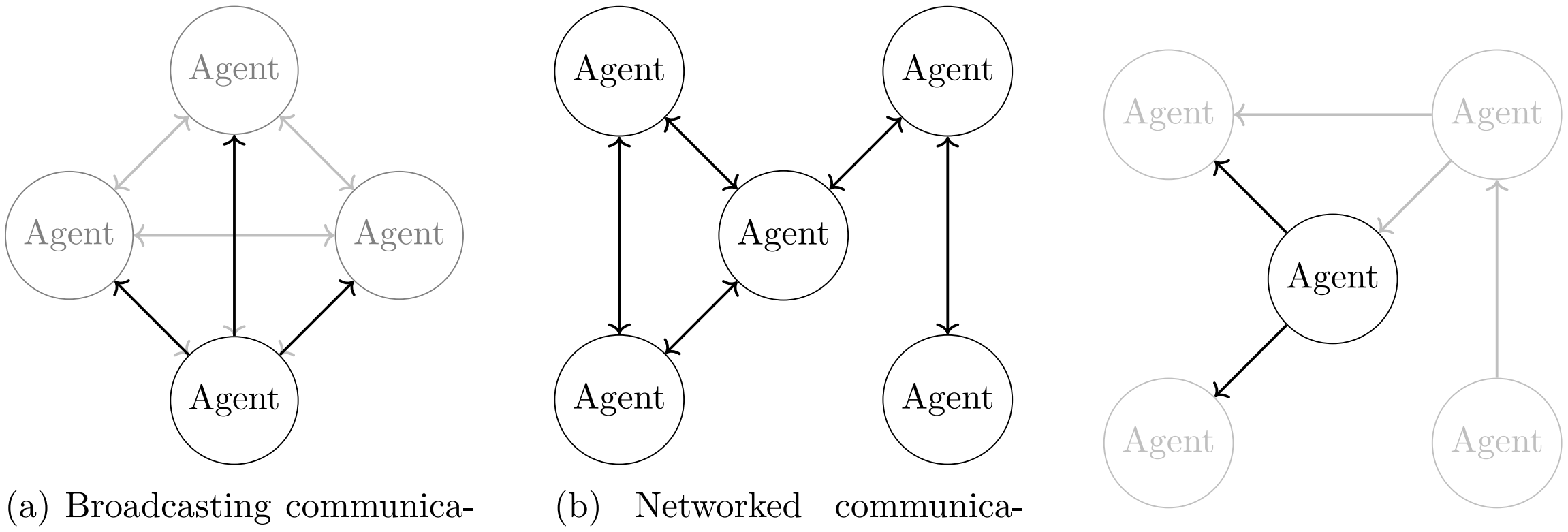

        \subsubsection{No Communication}
        
\paragraph{Definition}
The first trivial communication mode is: no communication. In this case, agents do not share any information with each other proactively. It limits the information available to each agent to its local observation.

\paragraph{Motivation}
This prevents the need for communication protocols, which can be complex to design. 

\paragraph{Challenges}
The absence of communication does not offer a solution to partial observability and non-stationarity, since agents lack explicit access to information about other agents' states, actions, or policies. However, it does not introduce additional computational complexity nor scalability concerns.

        \subsubsection{Broadcast Communication}
        
\paragraph{Definition}
Broadcast communication, illustrated in Figure~\ref{fig:broadcast_comm}, involves each agent sending messages to all others in the system. 

\paragraph{Motivation}
In this setup, agents receive information from all others. Then, agents can take actions based on more information: their local observation and more global information from other agents \citep{chen2020}.

\paragraph{Challenges}
This method can be very computationally expensive, since each agent has to process information from all others. 

Broadcast communication faces strong scalability issues. The more agents there are in the system, the more messages each agent has to process. This is not just computationally expensive, but also can lead to information overload, where agents have to process large amounts of information, including irrelevant messages \citep{jiang2018learning}.

        \subsubsection{Networked Communication}
        
\paragraph{Definition}
Networked communication, illustrated in Figure~\ref{fig:networked_comm}, involves agents communicating only with predefined agents. Such a communication graph must be defined.

How to set up the communication graph is a design choice.  For instance, it can be based on a distance metric, such as spatial proximity.

The network can be static or dynamic. When the network is dynamic, the connections between agents can change over time. If the neighbors are defined based on spatial proximity, then the network is dynamic because it can change when agents move. 

\paragraph{Motivation}

In this setup, agents can send information to their network's neighborhood but do not have to be able to communicate with all others, nor do they have to process information from all agents in the system. When the dynamics are mostly local, it is often sufficient to condition decisions on information only from nearby agents.

\paragraph{Example}
\cite{kar2013cal} present a networked communication method where agents communicate with neighboring agents over a sparse, pre-specified, and possibly time-varying communication graph. The communication is motivated by a setting where the global evaluation is not directly observable by individual agents, while agents have to minimize a global cost. A centralized architecture would require each agent to continuously send its local cost to a remote central unit, which is impractical, especially due to limited energy resources. Therefore, the authors adopt a networked communication approach. 

\paragraph{Challenges}
This technique is less computationally expensive than broadcast communication, since each agent only has to process information from a subset of agents.

        \subsubsection{Targeted Communication}
        
\paragraph{Definition}
Targeted communication, illustrated in Figure~\ref{fig:targeted_comm}, involves agents able to choose which other agents to communicate with. 

\paragraph{Motivation}
While networked communication tackles the scalability issue of broadcast communication by restricting communication to predefined neighbors, targeted communication methods enable agents to dynamically select what, with whom, or when to communicate, depending on the specific architecture.

\paragraph{Examples}
Agents in papers by \cite{jiang2018learning,das2019tarmac} can choose whether to share information with other agents either by selecting the content and recipient of a message \citep{das2019tarmac} or by deciding whether to initiate communication \citep{jiang2018learning}. 

\cite{jiang2018learning} present an application of this communication mode. In this work, agents can initiate a group of communication with other agents named collaborators. For a given number of steps, agents in the group can exchange messages. Agents use an attention mechanism to select when to initiate communication. However, the agents cannot select with whom to communicate, because agents initiate communication with other agents in their field.

\cite{das2019tarmac} present TarMAC, a targeted communication method where agents can select which other agents to communicate with and also what message to send. 

\paragraph{Challenges}
Unlike fixed network structures, targeted communication allows agents to potentially address any other agent in the system, while avoiding unnecessary message exchanges. This design aims to improve scalability and flexibility compared to broadcast communication, while preserving relevant information \citep{jiang2018learning,das2019tarmac}.

    \subsection{Model of Other Agents} \label{sec:moa}
    
In MARL, agents interact with the environment by receiving observations and rewards, and by taking actions. The non-stationarity problem arises because from the point of view of an agent, the dynamics of the environment are not only determined by the state and the actions of the agent, but also time-dependent. The time dependency is due to the policies of the other agents that are updating during the learning process. However, learning in a non-stationary environment is more challenging than learning in a stationary environment (see Section~\ref{sec:marl_nonstationarity}). 

Being able to consider the behavior of other agents is crucial in many MARL systems \citep{huh2023multi}. By modeling the other agents, separately from the environment, an agent can better understand the dynamics of the environment and predict other agents' behavior, which can help to take better actions and to reduce the impact of non-stationarity \citep{albrecht2024multi}. \cite{jaques2019social} argue that MOA is a good alternative to CTDE. The information or predictions produced by MOA can then be used by the learning module for action selection or policy updates.

This section focuses on how agent can predicts others' actions. However, MOA also encompasses other tasks, including classification\footnote{Classification of the behavior of agents. For instance, in games, it could be "aggressive" or "defensive".} and plan recognition\footnote{Plan recognition is the task of identifying the objectives of an agent.}. These two topics will not be discussed further here, a more detailed treatment is provided by \cite{albrecht2018autonomous}.

This section identifies two main components of MOA: (1) the representation of MOA, and (2) the learning of MOA (how such models are learned). 

By representation of MOA, we mean how other agents are modeled. Are they modeled individually as learning agents, are they modeled as a group, do they assume a default strategy for other agents? 
After one has decided on the representation of MOA, the next step is to learn the MOA. Some modeling do not require learning, but for those that do, the learning of MOA is about how an agent learns its models of others. There are many ways to learn the MOA, Section \ref{sec:learningmoa} identifies some of them.

        \subsubsection{Representation of MOA}
        
\paragraph{Independent Learner (IL)}
In Independent Learning, each agent ignores the others and treats them as part of the environment. Agents do not attempt to predict each other's behavior; they base their actions solely on their own observations, as in standard RL. 

This allows the use of single-agent RL algorithms, which are often simpler than multi-agent methods. 

IL was used by \cite{tan93} and subsequently identified as Independent Learning by \cite{claus98}. \cite{busoniu2008comprehensive} argue that Independent Learning is a valid approach with advantages (scalability, robustness, parallel computation). More recently, Independent Learning has been implemented in the StarCraft Multi-Agent Challenge (SMAC) \citep{samvelyan2019smac} and has demonstrated good performance \citep{witt2020}.

\cite{matignon2012independent} identify five major difficulties that arise with IL in a cooperative framework. These difficulties are: \textit{Pareto-selection} (the existence of at least two incompatible Pareto equilibria), \textit{non-stationarity}, \textit{stochasticity} (in state transitions or reward calculations), \textit{alter-exploration} (asynchronous exploration-exploitation by agents), and \textit{shadowed equilibria} (reward mechanics can lead to suboptimal Nash equilibria and mask action combinations that yield optimal outcomes). 

IL does not address the non-stationarity or the multi-agent credit assignment problems.

\paragraph{Joint Action Learning Game Theory (JAL-GT)} 
In some settings, agents can observe the actions of other agents. In this situation, they can evaluate their actions based on the joint action, rather than considering only their own action.

If an agent uses a $Q$-value function, it can extend it to depend on the joint action $\mathbf{a*}$, leading to a value function of the form $Q(s, \mathbf{a*})$. However, using such a function requires access to the actions of the other agents. In practice, this means that the agent must predict the actions of other agents.

Instead of using a prediction model, agents can use solutions based on game theory \citep{albrecht2024multi}. Such game-theoretic solution concepts can be for instance minimax \citep{littman1994markov} or Nash equilibrium \citep{hu2003nash}. 

The advantage of using game-theoretic solutions is that agents do not need to learn a model of other agents. However, they need to observe others' actions in order to update their policies.

Taking into account the actions of others and using more accurate action evaluations helps mitigate both the non-stationarity and the credit assignment problems.

\paragraph{Joint Action Learning Agent Modeling (JAL-AM)} 
Relying on game theory solution without predicting the actions of other agents prevents agents from exploiting weaknesses of other agents \citep{albrecht2024multi}.

An alternative approach, still based on agents observing joint actions, consists of learning models of other agents in order to predict their future actions, rather than relying on game-theoretic solution concepts. This approach is named JAL-AM by \cite{albrecht2024multi}.

An agent using JAL-AM can learn a best response to the predicted actions of other agents. This approach allows agents to exploit weaknesses of other agents, or to coordinate better with other agents. However, learning a model of other agents can be complex. The problem of learning a MOA is discussed in Section~\ref{sec:learningmoa}.

This approach presents advantages similar to those of JAL-GT, as it also helps mitigate the non-stationarity and credit assignment problems.

\paragraph{Group Modeling}
Modeling each agent independently assumes that their behavior can be predicted separately. However, sometimes there are strong correlations between agents' actions. The main limitation of modeling all agents independently with separate models, it may fail to capture correlations or coordination patterns among agents. 

Group Modeling addresses this limitation by modeling multiple agents jointly rather than independently. More precisely, the group is represented through a single model whose action space is the joint action space of the agents. It enables the prediction of joint actions while capturing dependencies between agents' actions \citep{albrecht2018autonomous}.

A major drawback of this approach is its sensitivity to the scalability issue, since the model of the group depends on the joint action space, whose grows exponentially with the number of agents. One possible mitigation strategy of this issue is to divide the full group into subgroups \citep{albrecht2018autonomous}

\paragraph{Mean Field}
In the Mean Field (MF) modeling approach, each agent considers the actions of the other agents collectively rather than individually. From an agent's perspective, the system reduces to two entities: the agent itself and the others, treated as a single entity: a mean agent. MF MARL was first introduced by \cite{yang2018mean}.

This approach may appear similar to Group Modeling, since in both cases the other agents are represented collectively. However, there is an important difference: unlike group modeling, MF does not aim to preserve the individual contribution of each other agent. Instead, it aggregates them and considers mean action. As a result, the identities and individual actions of the other agents are not explicitly represented in the model.

Accounting for the behavior of other agents can be computationally too expensive (see Section \ref{sec:marl_scalability}). This is why many algorithms work well with two agents but become impractical when the number of agents grows well beyond two. The Mean Field approach is motivated by the goal of recasting the problem into an effective two-entity setting \citep{yang2018mean, subramanian2021partially}. 

We describe the method from the viewpoint of agent $i$. Agent $i$ models the others as a mean agent. This mean agent does \textit{mean action} $\bar{a}^i$ defined by the Equation \ref{eq:mean_action}. In the latter, $\mathcal{N}^i$ denotes the set of agents in its neighborhood, excluding agent $i$ itself, $a^k$ is the one-hot encoding\footnote{One-hot encoding represents an action as a vector that is zero everywhere except at the index corresponding to the action.} of the action of agent $k$ and $\bar{a}^i$ is the average action of the agents in the neighborhood of agent $i$.

\begin{equation}
    \bar{a}^i = \frac{1}{|\mathcal{N}^i|} \sum_{k \in \mathcal{N}^i}a^k
    \label{eq:mean_action}
\end{equation}

To select its action, agent $i$ uses the mean action and act accordingly. Originally, this mean action was not predicted but observed \citep{yang2018mean}, but then, it was proposed to predict the mean action instead \citep{subramanian2021partially}.

In some situations, it is not desirable to summarize the actions of all other agents with a single mean field, especially when the agents are heterogeneous. For example, in a team-versus-team system, it may be important to maintain different treatments for allies and enemies. Thus, it is possible to consider multiple types of agents. \cite{subramanian2020multi} present the Multi-Type MF MARL approach.

        \subsubsection{Learning MOA} \label{sec:learningmoa}
        
To construct the MOA, agents usually use both their local observations and the actions performed by other agents \citep{jaques2019social, papoudakis2021agent,albrecht2024multi}.

The central questions in policy reconstruction are: (1) what elements in the history of actions should be used, (2) how to use these elements to make predictions \citep{albrecht2018autonomous}. This section presents different approaches for learning the MOA. The following partition of the approaches draws inspiration from \cite{albrecht2018autonomous}.

\paragraph{Action Frequencies}
A simple approach to learn a MOA is to base predictions on past action frequencies. In multi-state settings, these frequencies are typically conditioned on the current observation or context. Fictitious Play (FP) is the best-known example of this approach. It was introduced by \cite{brown1951}. In FP, agents assume that others follow a stationary mixed strategy \citep{berger2005}, that is, they assume that the strategies of the other agents do not change over time. To predict the actions of others, each agent records the past actions of the others and uses this information to estimate their mixed strategies. For example, in a Rock-Paper-Scissors game, if Agent A has played Rock three times and Scissors once, then Agent B will assign a 75\% probability to Rock on the next move and 25\% to Scissors. 
Since it's introduction by \cite{brown1951}, FP has been used in MARL. The first adaptation of FP to MARL was proposed by \cite{claus98}, where it was used in conjunction with JAL-AM. More recently, \cite{heinrich2016deep} incorporated FP into a deep learning framework and was used in a poker setting to converge to an approximate Nash equilibrium. 
However, assuming stationary strategies is restrictive: the faster other agents' policies are updated, the less valid this assumption becomes. Some approaches still rely on action frequencies, but compute these frequencies using selected portions of the history elements. The main difficulty with this type of MOA training lies in selecting appropriate elements from the history. Using inappropriate or not enough elements may lead to unreliable predictions, whereas using too many elements may be too slow and not efficient \citep{albrecht2018autonomous}.

\paragraph{Case-Based Reasoning}
Policy reconstruction methods based on action frequencies have one major drawback: in order to predict the next action in a given state, they require records of past actions taken by others in that same state. Action frequencies based predictions lack the ability to extrapolate from past observations to unseen situations \citep{albrecht2018autonomous}. 

Case-based reasoning (CBR) addresses this limitation by relying on similarity functions. A similarity function defines a distance between observations. Using these functions, the approach retrieves the most similar past observations in order to predict the future actions of others. 

CBR was used by \cite{borck2015case} to predict the behavior of hostile and allied agents in the context of air combat. In that study, CBR outperformed the baseline strategy used for comparison.

\paragraph{Compact Model Representations}
Having to consider a multitude of other agents actions and observations for every action prediction may be too computationally expensive. An alternative to action-frequency-based and CBR predictions is to use a compact machine model from machine learning method for each agent. For instance, one can use neural networks (NN) to model others. The neural networks are trained to make good predictions based on observations of the state and actions of others. \cite{davidson2000improved} present agents that used NNs to predict the next action of their opponent in a game of poker, achieving high prediction accuracy. 

Machine learning methods offer greater flexibility and do not require fully observable situations \citep{albrecht2018autonomous}.


\paragraph{Type-Based Reasoning}
Learning a model of an opponent from scratch is a long and difficult process. During one simulation, an agent may not have enough information to correctly model others (and adapt to it). An alternative of online MOA is to train the agent against a variety of different agents and to record models of these agents. Then, during the simulation, an agent avoids reconstructing a full model for each agent and instead determines which learned model most closely approximates that agent's behavior. Type-based reasoning methods consist in identifying and learning models for a set of agent types, and using them to model previously unseen agents from limited observations.

\cite{baker2007bayesian} use this method for their agents. This article defines four different play styles and the agent uses its observations to determine to which style their opponent resembles most and bases its strategy accordingly. In this example, the different types of agents are explicitly defined by the authors. However, it is also possible to learn them. \cite{hou2022behavior} generate 1000 types of agent, initialized with diverse architectures and trained on historical data from online interactions. Selecting the appropriate type for an opponent among so many possibilities is a difficult task, \cite{hou2022behavior} propose a solution based on two techniques: (1) the extraction of a small and representative subset of candidates, and (2) the identification of the most appropriate model using multi-armed bandits.

    \subsection{Agent} \label{sec:core_agent}
The agent component plays a central role in a MARL system. Its importance comes from its role as an interface between several core modules, such as the learning algorithm, the policy, and, when present, the communication and MOA modules. 
As illustrated in Figure~\ref{fig:marl}, agents call their policy to map received observations to actions and provide the resulting experience to the learning algorithm, which updates the policy.


Most MARL approaches do not require agents to perform operations beyond orchestrating their appropriate modules in the appropriate order. However, some approaches require more specialized agent implementations, in which the agent itself carries additional responsibilities. The following subsections first present the default agent implementation, then illustrate how agents can be extended with additional responsibilities through one representative alternative.

        \subsubsection{Default Agent} \label{sec:core_agent_default}
By default, agents receive observations and reward and redirect this information to other specialized modules, namely the policy and the learning algorithm. Agents can sometimes communicate or model other agents, but this responsibility rely on the dedicated core modules. In this perspective, the agent work more like an orchestrator, aggregating different modules and call them appropriately.

This agent perspective is consistent with the definition give in Section~\ref{sec:rl_agent}.

        \subsubsection{Multi-Policy Agent}
\paragraph{Definition}
The multi-policy approach consist of agents having different policies they use depending on the context. The said policies must be learn separately during different training phase and can be use during the execution phase.

Then the multi-policy agents need to define how it select which policy to use depending on the context. Different possibilities are presented in the examples below.

\paragraph{Motivation}
During training, agents learn alongside other agents. Since the best action for an agent depends on the joint action, the optimal policy may depend on the policies of the other agents. This dependency gives rise to the non-stationarity challenge discussed in Section~\ref{sec:marl_nonstationarity}. Consequently, a policy trained with a specific set of agents may not generalize well when paired with different agents \citep{barrett2015cooperating, zheng2018deep}.

Using a library of policies allow for training with different type of agents and have an appropriate policy against one type of agents. After the training phases against different type of agents, when the agent face a new agent, it will select the more appropriate policy \citep{barrett2015cooperating}. Therefore, the agent do not need to learn a new policy from scratch in order to adapt to the other agents, which is usually a very slow process.

\paragraph{Examples}
\cite{barrett2015cooperating} use a multi-policy approach in a cooperative setting, namely RoboCup, where agents within the same team cooperate while competing against an opposing team. In their situation, they use a multi-policy agent equipped with several learned policies because one agent is trained to cooperate with teams exhibiting different play styles and must adapt to previously unseen teammates. To select which policy the agent should use, they consider two mechanisms for policy selection. The first one consists of modeling policy selection as a multi-armed bandit problem~\citep{sutton2018}. However, this approach is considered too slow in their use case. Instead, they propose PLASTIC-Policy, which selects among the learned policies by updating beliefs over teammate types. PLASTIC-Policy maintains probabilities over the new team being similar to previously observed teams, using the polynomial weights algorithm introduced by \cite{barrett2013teamwork}.

\cite{zheng2018deep} also use a multi-policy approach but in settings that are not necessarily cooperative. In their situation, other agents are non-stationary and may switch between different policies over time. The motivation for maintaining a library of policies comes from the need for fast adaptation to the others' current policy. To select the appropriate policy, the agent uses an extension of Bayesian Policy Reuse (BPR)~\citep{hernandez2016identifying}, namely BPR+, designed to handle non-stationary agents; and combines it with deep neural networks, resulting in deep BPR+. Deep BPR+ bases its policy selection on both received cumulative reward and the observed behavior of other agents. In this version of multi-policy approach, the agent does not only rely on a set of policies learned in a training phase and fixed afterward. Here, the agent can detect previously unseen behaviors of other agents, and start a new policy learning process, and add the resulting policy to its library.

\paragraph{Challenges}
\cite{zheng2018deep} motivate their approach as a solution against a non-stationary problem occurring because agents do not have a fixed policy. Using library of different policies allow for faster adaptation to other agents behavior, reducing the non-stationary issue.

    \subsection{Evaluation Criteria} \label{sec:evaluation_criteria}
    
Evaluating a MARL system is more complex than evaluating a single-agent RL system (see Section~\ref{sec:from_rewards_to_system_goal}). In a single-agent RL setting, a solution is better than another if it achieves a higher cumulative reward. Therefore, evaluating a solution is straightforward. In MARL, except in fully cooperative settings (see Section~\ref{sec:fully_cooperative}), the objective is not to maximize a single value.

The evaluation of a MARL system can rely on a wide range of experiment-specific measures, depending on the objectives of the experiment and the dynamics of the environment. Such measures may include, for example, the number of prey captured (for a prey/hunter experiment), the time required to complete a collective task, or the level of competition between agents for limited resources. Beyond these experiment-specific measures, general solution concepts provide criteria that can be applied across different MARL settings. This section focuses on such concepts, which formalize how the outcomes of multi-agent interactions can be evaluated. The presentation draws inspiration from \cite{albrecht2024multi}.


\cite{busoniu2008comprehensive} identify two main types of learning objectives in MARL: stability and adaptability. Stability refers to a system's ability to converge to a stable state. Adaptability refers to an agent's ability to adapt its strategy as the other agents learn. The concepts presented below can be linked to these two objectives.

        \subsubsection{Stability}
        
\paragraph{Convergence}
Convergence is a criterion related to stability. A MARL algorithm is said to converge if the agents' policies converge to a stationary policy \citep{bowling2002multiagent}.

Analyzing convergent system is easier by looking at the final policies of the agents. This criterion is often sought after in MARL applications \citep{bowling2002multiagent, conitzer2007awesome}.

\paragraph{Opponent Independent Solution}
An another stability criteria for the agent is to have a learning process that is independent of the others' actions. The motivation of such criteria is to have a learning process which stability does not depend on the others algorithms \citep{busoniu2008comprehensive, bowling2000analysis}. 

For instance, \cite{littman1994markov} uses a Minimax-$Q$ algorithm to learn for its agents. The agents learn a policy that is optimal in the worst case scenario. No matter what the others do, agents using Minimax-$Q$ will learn a policy that converges.

        \subsubsection{Adaptability}

\paragraph{Best Response}
A best response is a policy that maximizes an agent's expected reward, given the policies of the other agents. An agent's best response can be defined mathematically using the equation \ref{eq:best_response}.

\begin{equation}
    BR_i(\pi_{-i}) = \arg\max_{\pi_i} \mathbb{E}[G_i | \pi_i, \pi_{-i}]
    \label{eq:best_response}
\end{equation}
Where $G_i$ is the return (cumulative reward) of agent $i$, $\pi_i$ is the policy of agent $i$ and $\pi_{-i}$ is the joint policy of all agents except agent $i$.

A best response does not have to be unique. An agent can have multiple best responses to the same joint policy of the other agents.

Here's an example that illustrates the concept of best response. Consider a simple two-player game: Rock-Paper-Scissors. Assume that player A always plays Rock. Given player A's policy, the best response for player B would be to always play Paper. However, if Player A's policy changes to playing randomly, then playing Rock or playing randomly would both be best responses.

\paragraph{No-regret}
Regret is a measure and is defined as the difference between the rewards an agent could have received by playing the optimal policy and the rewards it actually receives during an episode. It can be mathematically defined with the equation \ref{eq:regret}.

\begin{equation}
    Regret_i(T) = \max_{\pi_i \in \Pi_i} \; \sum_{t=1}^T \Big( \mathbb{E}[r_i^t \mid \pi_i, \pi_{-i}^t] - \mathbb{E}[r_i^t \mid \pi_i^t, \pi_{-i}^t] \Big)
    \label{eq:regret}
\end{equation}
Where $T$ is the number of learning steps, $t$ is a step of learning, $r_i^t$ is the reward of agent $i$ at time $t$, $\pi_i^t$ is the policy of agent $i$ at time $t$, $\pi_{-i}^t$ is the joint policy of all agents except agent $i$ at time $t$, and $\Pi_i$ is the set of all policies of agent $i$.

An algorithm is said to be no-regret if for all agents $i$:
\begin{equation}
    \lim_{T \to \infty} \frac{1}{T} Regret_i(T) = 0
\end{equation}

The no-regret approach has a significant limitation because it evaluates an agent's performance based on the observed policies of other agents. However, since the other agents are also learning, their policies depend on the agent's own policy. Therefore, regret cannot predict what could have happened under different circumstances. Additionally, Albrecht et al. also argue that no-regret does not necessarily equate to maximized rewards. The prisoner's dilemma is a simple example of this: if both agents defect, the outcome is a no-regret situation, yet both agents could have achieved a higher reward by cooperating \citep{albrecht2024multi}.

        \subsubsection{Equilibrium}

Equilibrium is a concept that combines both stability and adaptability. The objective is to converge to a stable and rational solution. The motivation is the combined the benefits of both stability and adaptability. 

There are different types of equilibrium, the most common ones are Nash equilibrium and Pareto optimality, which are presented below.
        
\paragraph{Nash Equilibrium} \label{sec:nasheq}
In MARL, the dynamics of learning may converge to an equilibrium. Convergence is important because it provides a basis for evaluating the outcomes that emerge from their interactions. However, not all equilibria are equally desirable. A widely used concept to evaluate the outcomes of multi-agent interactions is the Nash equilibrium (NE) \citep{nash51}.

A Nash equilibrium occurs when each agent simultaneously adopts the best possible policy, given the choices made by the others. This term can be defined mathematically: $\bm{\pi}^*$ is a Nash equilibrium if and only if:
\begin{equation}
  V^i_{\pi^{*}_i, \pi^{*}_{-i}}(s) \geq V^i_{\pi_{i}, \pi^*_{-i}}(s),  \forall \pi_i \in \Pi_i, \forall s \in \mathcal{S}, \forall i
  \label{eq:nash_equilibrium}
\end{equation}
where $\pi^i$ is the policy of agent $i$, $\pi^{-i}$ the policy of agents other than $i$, $\bm{\pi}^* = \{\pi^{*,i}, \pi^{*,-i}\}$ is the joint policies of all agents and $s$ the state of the environment.

There are two types of Nash equilibrium, depending on whether we consider only deterministic policies (no probabilistic choice, $\pi(s,a) \in \{0,1\}$) or also stochastic policies ($\pi(s,a) \in [0,1]$). In the case of a prisoner's dilemma, there is a deterministic policy that constitutes a Nash equilibrium: both players defect. In other cases, such as Rock-Paper-Scissors, no deterministic policy satisfies the Nash equilibrium condition. However, in this case, there is a stochastic policy that forms a Nash equilibrium: both players randomly select each option with equal probability.

More generally, the Nash existence theorem \citep{nash51} ensures the existence of at least one Nash equilibrium under the following conditions:
\begin{itemize}
  \item Stochastic policies are allowed;
  \item The number of players is finite;
  \item The number of actions of each player is finite.
\end{itemize}

In practice, it can be difficult to reach a Nash equilibrium. In these difficult cases, an approximation of the Nash equilibrium, the $\epsilon$-Nash equilibrium ($\epsilon > 0$), can be used. It is defined almost the same way as the Nash equilibrium, except that each agent's policy can be slightly suboptimal. The greater $\epsilon$ is, the more suboptimal the policies can be. $\bm{\pi}^*$ is an $\epsilon$-Nash equilibrium if and only if:
\begin{equation}
    V^i_{\pi^{*}_i, \pi^{*}_{-i}}(s) \geq V^i_{\pi_{i}, \pi^*_{-i}}(s) - \epsilon,  \forall \pi_i \in \Pi_i, \forall s \in \mathcal{S}, \forall i
    \label{eq:epsilonnash_equilibrium}
  \end{equation}

\paragraph{Pareto Optimality} \label{sec:pareto_optimality}
As mentioned previously in the Nash equilibrium Section \ref{sec:nasheq}, equilibria are not equal and can have different properties. Another one interesting property is Pareto optimality.

A joint policy $\bm{\pi'}$ is Pareto dominant over $\bm{\pi}$ if and only if:
\begin{equation}
    \forall i: U_i(\bm{\pi'}) \geq U_i(\bm{\pi}),  \text{ and } \exists i: U_i(\bm{\pi'}) > U_i(\bm{\pi})
\end{equation}
where $U_i(\pi)$ is the expected cumulative reward of agent $i$ when all agents follow the joint policy $\pi$.

A joint policy $\bm{\pi^*}$ is Pareto optimal if and only if no other joint policy Pareto dominates it \citep{albrecht2024multi}. In other words, an equilibrium is Pareto optimal if and only if it is impossible to increase one agent's reward without decreasing at least one other agent's reward.

In fully competitive settings, the concept of Pareto optimality is meaningless because all joint policies are Pareto optimal. Effectively, because the sum of all agents' rewards is constant, increasing one agent's reward implies decreasing another's. Therefore all equilibria are Pareto optimal.

\paragraph{Limitations of Equilibrium Solutions}
The equilibrium solution concepts presented above have two important limitations. First, convergence to these equilibria does not guarantee maximized rewards. They can be suboptimal for all agents. For instance, in the prisoner's dilemma, both players defecting is the only Nash equilibrium; however, both players would have received a greater reward if they had cooperated. This is a limitation shared with the no-regret concept presented above. In the Prisoner's Dilemma, there are three Pareto equilibria: both players cooperate, or one cooperates while the other defects. The first one is optimal for both players, while the other two are suboptimal for one of them.

This introduces the second limitation: a system can have multiple equilibria of the same type. If an algorithm is designed to guide agents toward strategies that lead to an equilibrium, then the agents should be able to select which equilibrium to converge to. This problem is known as equilibrium selection \citep{albrecht2024multi}.

\subsubsection{Fairness}

One important aspect that is not captured by the previous solution concepts is fairness. Fairness is a measure of how equitably the rewards are distributed among agents. \cite{albrecht2024multi} define the fairness of a joint policy $\bm{\pi}$ as follows:
\begin{equation}
    Fairness(\bm{\pi}) = \prod_{i=1}^{N} U_i(\bm{\pi})
    \label{eq:fairness}
\end{equation}
Where $U_i(\bm{\pi})$ is the expected cumulative reward of agent $i$, $N$ is the number of agents and $\bm{\pi}$ is the joint policy of all agents.

If one agent receives a low reward, it will significantly reduce the overall fairness. This measure encourages fairer distribution of rewards among agents. A joint policy $\bm{\pi}$ that satisfies $\bm{\pi} = \arg\max_{\bm{\pi'}} Fairness(\bm{\pi'})$ is said to be fairness optimal \citep{albrecht2024multi}.

This fairness function is related to performance. For example, consider a two-player game and two joint policies $\bm{\pi}$ and $\bm{\pi'}$ where $\bm{\pi}$ leads to $U_1(\bm{\pi})=10, U_2(\bm{\pi})=1$ and $\bm{\pi'}$ leads to $U_1(\bm{\pi'})=3, U_2(\bm{\pi'})=3$. In this example, $Fairness(\bm{\pi}) = 10 > Fairness(\bm{\pi'}) = 9$. This fairness function encourages more equitable distribution of rewards while considering the system's overall performance. This definition of fairness is one possibility, but other definitions could be used that give more importance to equity.

The concept of fairness is irrelevant in fully cooperative settings, since all agents share the same reward function. Under this definition of the fairness function \ref{eq:fairness}, maximizing fairness is equivalent to maximizing the global reward.

    \subsection{Modular Perspective on Other Papers' Approaches} 

To illustrate the usefulness of this categorization into core modules, we analyze three recent MARL articles and describe their proposed approaches from this modular perspective.

        \subsubsection{MADDPG} \label{sec:maddpg_paper}
        
\cite{lowe2017multi} introduce MADDPG, a MARL actor-critic approach designed for mixed cooperative-competitive environments. The main motivation is to address two difficulties of applying standard RL algorithms to MARL: the non-stationarity induced by simultaneously learning agents, and the high variance of policy-gradient methods when rewards depend on the actions of several agents.

Several experiments are conducted in different environments. Since the system configuration differs across experiments, this analysis focuses on the cooperative navigation experiment using MADDPG.

MADDPG follows an actor-critic approach based on deep deterministic policy gradient (DDPG) and combines it with a CTDE paradigm, where each agent uses a centralized critic during training and a decentralized actor during execution.

Agents have to cover all landmarks and are collectively rewarded based on the proximity of any agent to each landmark. They are also penalized when they collide with each other. Thus, the reward model encourages agents to distribute themselves across landmarks while avoiding physical conflicts.
The reward model is cooperative. However, the article leaves a slight ambiguity as to whether the reward is fully shared. The landmark-coverage component is explicitly described as collective, but it is unclear whether the collision penalty applies to all agents or only to those involved in a collision. The released code shows that the penalty is first assigned to the colliding agents and then collectivized, making the reward model fully cooperative.

This experiment does not rely on an explicit communication mechanism. Agents coordinate through their physical actions and through the information available in their observations, which include the relative positions of other agents and landmarks.

MADDPG requires access to the other agents' policies, or learned approximations of them, to obtain their actions during training. In the default setting, these models are not learned: each agent has direct access to the policies of the other agents. The paper presents this requirement as closely tied to the use of a centralized critic, although it is more accurately understood as a consequence of MADDPG's specific critic formulation. Other applications of centralized critics, such as COMA, do not impose the same requirement \citep{foerster2018counterfactual}. While the paper considers policy inference as an alternative when the true policies are unavailable, this mechanism is presented as an additional extension rather than as the main configuration of the cooperative navigation experiment. We believe that the distinction between centralized critics and explicit modeling of other agents' policies should therefore have been made more prominent.

The agent module follows a standard orchestration role: it calls the appropriate internal modules when needed, such as the policy for action selection and the learning algorithm for policy updates.

The system is evaluated by considering how well agents cover the landmarks and avoid collisions. \cite{lowe2017multi} report in particular the average distance to each landmark and the average number of collisions per episode when comparing MADDPG with DDPG.

        \subsubsection{WoLF-PHC}

\cite{yang2023reinforcement} study how Virtual Power Plants (VPPs) can define bidding strategies in electricity markets. The main motivation is to obtain optimal bidding strategies and market equilibrium results when VPPs participate alongside traditional power plants.

The environment corresponds to a power market game. Agents are power producers, either traditional power plants or VPPs. 

The reward model is mixed. Each agent has its own objective and reward, corresponding to its own profit. Rewards are not shared among agents.

The learning module is based on WoLF-PHC, which combines Policy Hill Climbing with the Win or Learn Fast mechanism. In order to avoid local optima and improve convergence speed, \cite{yang2023reinforcement} use WoLF, which adjusts the learning rate based on performance. Each agent maintains its own Q-table, updated based on its own observations, actions, and rewards.

The training and execution mode corresponds to DTE: agents learn independently and no centralized training or execution mechanism is introduced. The paper does not describe any explicit communication mechanism between agents, nor any MOA strategy. Other power producers are therefore treated as part of the environment. The agent module does not introduce any specific agent-level mechanism: each agent simply selects a bidding strategy according to its current policy and updates this policy through WoLF-PHC.

The system evaluation is mainly based on two market outcomes: power output and revenue. \cite{yang2023reinforcement} compare these quantities in two scenarios: one with only traditional power plants, and one where one traditional power plant is replaced by a VPP.

        \subsubsection{NC-HDQN}

During epidemic episodes, traffic flow patterns can change significantly, with heavier traffic near hospitals and lighter traffic in controlled epidemic communities. To optimize traffic signal control in this context, \cite{zhang2022neighborhood} propose NC-HDQN, a neighborhood cooperative MARL approach for adaptive traffic signal control.
The term NC-HDQN encompasses various pieces of information regarding specific, distinct modules. We will go into detail to explain it.

The environment corresponds to a traffic network modeled as a networked Markov game. Each intersection is modeled as an agent, and actions correspond to traffic signal phases. Agents use local observations and information from physically connected neighboring intersections.

The reward model is cooperative, but not fully cooperative. The objective is to optimize regional traffic conditions, but each agent receives a reward based on its local traffic situation and the rewards of its neighbors, weighted by correlation coefficients. The reward model encourages cooperation while preserving neighborhood-level differences between agents.

Communication is present through neighborhood information sharing. Agents receive observation information from neighboring agents. The network structure is fixed by the road network and the received information is weighted using a correlation mechanism.

The learning module is based on Hysteretic DQN (HDQN). DQN provides the basic value-based architecture, while HDQN introduces different update rates for positive and negative TD errors. \cite{zhang2022neighborhood} justify this choice by the limitations of independent learners in cooperative MARL: because other agents are simultaneously learning and exploring, an agent may observe poor outcomes even when its own action is beneficial. Hysteretic updates reduce the impact of such negative experiences and give more importance to positive ones, which helps agents avoid suboptimal cooperative strategies. To use the information provided by the communication module, \cite{zhang2022neighborhood} propose Neighborhood Cooperative HDQN (NC-HDQN), which weights neighboring observations and rewards according to correlation coefficients.

The training and execution mode corresponds to DTE. Although CTDE is common in MARL, \cite{zhang2022neighborhood} argue that the large number of intersections in ATSC makes centralized training impractical. Agents are therefore trained and executed independently.
Agents do not model others. The agent module follows the default orchestration role described in Section~\ref{sec:core_agent}.

The system is evaluated during execution using three measures: average travel time, average queue length, and the number of vehicles arriving at their destination.

        \subsubsection{Summary of the Modular Analysis}
We summarize our analysis with a core module table (see Table~\ref{table:modular_perspective}). This table illustrates the process of applying the modular perspective to existing MARL approaches by making explicit the choices associated with each core module.
        
\begin{table*}[ht!]
  \centering
  \renewcommand{\arraystretch}{1.2}
  \begin{adjustbox}{width=\textwidth, center}
  \begin{tabular}{|c|c c c|}
    \hline
    Core Module & \cite{lowe2017multi} & \cite{yang2023reinforcement} & \cite{zhang2022neighborhood}
    \\ \hline\hline

    Environment 
    & Cooperative navigation 
    & Power market game 
    & Networked traffic signal control
    \\ \hline

    Agent 
    & Default 
    & Default 
    & Default
    \\ \hline

    Learning 
    & DDPG 
    & WoLF-PHC 
    & HDQN
    \\ \hline

    Training and Execution Modes 
    & CTDE (Centralized Critic)
    & DTE 
    & DTE
    \\ \hline

    Reward Modeling 
    & Fully cooperative 
    & Mixed 
    & Neighborhood Cooperative (NC)
    \\ \hline

    Communication 
    & None 
    & None 
    & Networked. Observation sharing
    \\ \hline

    MOA 
    & Other Policies Directly Available
    & None 
    & None
    \\ \hline

    Evaluation Criteria 
    & Landmark distance + collisions 
    & Power output + revenue 
    & Travel time + queue length + throughput
    \\ \hline
  \end{tabular}
  \end{adjustbox}
  \caption{Modular perspective on selected MARL approaches}
  \label{table:modular_perspective}
\end{table*}

This analysis shows that substantially different approaches can be described using the same decomposition into core modules. By separating the core modules and making their respective implementations explicit, this perspective provides a clearer view of each proposed solution. It also helps identify the specific contribution of a new approach by making explicit which components have changed compared with existing approaches. In addition to describing the proposed solution, the decomposition specifies the context in which it was tested: the task being addressed, the reward model used, and the evaluation criteria used for the solution. 

For instance, \cite{lowe2017multi} extend DDPG by using centralized critics during training. This approach is evaluated in different experiments, including a fully cooperative navigation task where distance to landmarks and collisions were the main evaluation criteria.
The strategy proposed by \cite{yang2023reinforcement} was to modify how a standard PHC learning algorithm updates its policy by using WoLF, an adaptive learning-rate mechanism.

Overall, such a decomposition provides a clearer view of a proposed solution, and helps identify which components could be modified to explore alternative configurations. It offers a common structure for presenting MARL approaches, making it easier to identify the key differences between solutions and between the experimental configurations in which they are evaluated.

\section{Analysis of MARL Tools}

Having established the theoretical relevance of the modular perspective in the previous section, we now examine existing MARL tools through this perspective. We first describe their general organization and main functionalities, before analyzing how they represent the core modules identified in this paper and to what extent they support varying their implementations independently.

Several software tools have been proposed to support MARL experimentation, but they do not all address the same level of abstraction. In the following, we focus on a representative set of widely used tools: PettingZoo \citep{terry2021pettingzoo}, OpenSpiel \citep{lanctot2019openspiel}, RLlib \citep{liang2018rllib}, MARLlib \citep{hu2023marllib}, and EPyMARL \citep{papoudakis2021benchmarking}. This selection is not intended to be exhaustive. Rather, it covers complementary roles in the MARL ecosystem, especially environment standardization and MARL experimentation with different algorithms. This diversity makes these tools useful for assessing how current MARL software supports, or fails to support, the explicit modular exploration of MARL implementations.

\subsection{Tools Rationales}
Among MARL tools, some are especially useful for defining and accessing multi-agent environments. PettingZoo provides a multi-agent alternative to Gymnasium environments, with standardized interfaces that make environments easier to reuse across different learning frameworks. 

OpenSpiel also provides a large collection of multi-agent environments, but its scope is broader: it is designed around game-theoretic modeling and includes both environments and algorithms for studying different classes of games, as well as tools for analyzing learning dynamics and other evaluation metrics.

RLlib provides a general scalable reinforcement learning framework with multi-agent support. RLlib can run multiple copies of the environment in parallel to collect training data faster. RLlib also supports flexible mappings between agents and policies, which makes it possible to implement algorithm self-play, policy self-play, or mixed-play configurations (see Section~\ref{sec:marl_selfplay_mixedplay}). In addition, for components that rely on neural networks, such as policies and critics, RLlib allows the underlying neural architecture to be configured or replaced without modifying the surrounding execution pipeline, provided that the new architecture satisfies the interface and outputs expected by the learning algorithm.

MARLlib builds on top of RLlib\footnote{MARLlib is build on an older version of RLlib.} and provides a higher-level interface dedicated to MARL experimentation. Its primary rationale is to retain RLlib's scalable execution infrastructure while providing a more structured and accessible framework for multi-agent tasks. MARLlib organizes supported algorithms into three groups: independent learning, centralized critic, and value decomposition. This allows algorithms within the same group to base their learning on common data structures. The data consumed by MARLlib is harmonized rather than being directly based on the heterogeneous structures exposed by external environments. Because some algorithms are not suited to certain task settings, MARLlib checks the suitability of the selected algorithm for the task. Its flexible policy-mapping strategy allows policies and their parameters to be shared by all agents, shared within groups, or kept separate for each agent.

EPyMARL also focuses on MARL experimentation, with an emphasis on comparing algorithms across environments coming from different sources, such as PettingZoo, Gymnasium, and SMAC. It defines a common multi-agent environment interface and relies on wrappers to use environments from these different sources. Building on PyMARL\footnote{PyMARL is a framework originally developed by \citet{samvelyan2019smac} alongside SMAC to evaluate different algorithms on this specific benchmark. It is not discussed further here, as it is no longer actively maintained and its inclusion would largely overlap with the analysis of EPyMARL.}, EPyMARL extends the available set of algorithms and provides configurable implementation choices, such as parameter sharing, reward standardization, and network architecture\footnote{EPyMARL notably allows users to choose between fully connected and recurrent networks \citep{papoudakis2021benchmarking}.}. As noted by \cite{papoudakis2021benchmarking}, implementation details can significantly affect algorithm performance. Exposing these choices facilitates their controlled and consistent evaluation.

\subsection{Analysis of MARL Tools Through the Core Module Perspective}
In this section, we analyze the MARL tools from the perspective of our core modules. We also examine the extent to which their architectures follow the single-responsibility principle.

\subsubsection{Environment}

These tools all provide interchangeable and extensible environment abstractions and therefore have modular implementation of the environment. 

PettingZoo defines different types of environment, and is designed to facilitate the integration of different multi-agent environments with external training frameworks.

RLlib defines a common multi-agent environment interface through which the rest of its multi-agent pipeline interacts with environments. Although RLlib provides some of its own environments, its design primarily supports custom environments and the adaptation of environments from external tools, such as Gymnasium, PettingZoo, and OpenSpiel, through wrappers.

MARLlib works in a similar way to RLlib and relies on the same multi-agent environment interface. However, unlike RLlib, which uses one generic environment wrapper for each external tool (PettingZoo, OpenSpiel\etc), MARLlib generally wraps each environment suite separately.

EPyMARL also defines a common multi-agent environment interface and provides wrappers that standardize access to environments from external sources, such as SMAC\footnote{The StarCraft Multi-Agent Challenge (SMAC) is a MARL benchmark introduced by \citet{samvelyan2019smac} for evaluating algorithms on cooperative micromanagement tasks based on StarCraft II} and PettingZoo. Unlike RLlib, EPyMARL does not implement environment dynamics and relies on wrapped external environments.

OpenSpiel directly implements a large collection of games and exposes them through a common reinforcement learning interface. A general component orchestrates the interaction with the selected game and exposes it through a reinforcement learning interface. The game describes its general properties, while its state implementation defines how actions modify the current state and provides the resulting rewards.

Although these tools provide modular and interchangeable environment implementations, they do not separate the responsibilities identified in our modular perspective. In particular, the same environment implementation generally defines both how the system reacts to the agents' actions and how rewards are computed. It therefore combines the Reaction and Reward modules within the same component and does not respect the single-responsibility principle. This coupling leads to the issues discussed in the following section.

\subsubsection{Reward}

In these tools, rewards are computed by the environment\footnote{In OpenSpiel, the concrete \class{State} implementation is responsible for both applying the environment dynamics and providing the rewards; therefore, the same criticism applies.}. Reward modeling is not exposed as an independent, reusable module that can transform, share, or redistribute rewards across different environments. 

Consequently, testing the same environment dynamics under a different reward structure, for example changing from individual rewards to a fully cooperative reward, usually requires modifying or wrapping the environment separately for each experiment. An independent reward-modeling module would instead allow researchers to reuse reward structures across environments and compare different reward models while keeping the environment dynamics unchanged.

MARLlib illustrates this lack of modularity particularly well. To support fully cooperative rewards across different environments, it defines separate cooperative versions of these environments, each overriding the original reward logic in the \method{step} method. Supporting another reward model would similarly require implementing environment-specific variants rather than adding a single reusable reward component.

\subsubsection{Training and Execution Modes}

During training, execution, or both, a centralized process may influence how agents learn or behave. A common centralized technique combines an actor critic learning algorithm with a centralized critic. Other common techniques include value decomposition and parameter sharing managed by a centralized component.

OpenSpiel does not provide dedicated support for any centralized technique. Using them requires implementing their logic as part of a specific algorithm.

PettingZoo is an environment tool and does not implement training or execution modes. This core module is therefore outside its scope.

RLlib directly supports parameter sharing as a centralized training approach: multiple agents can be mapped to the same policy module and therefore trained using a shared set of parameters. Agents can alternatively be mapped to separate policy instances with distinct parameters. Other centralized training techniques, such as centralized critics and value decomposition, are not supported through generic, reusable components. RLlib provides specific implementations of these techniques, but their centralized training logic is embedded in these implementations and must be adapted or reimplemented for use with other algorithms.

EPyMARL directly supports parameter sharing as a centralized approach. It also provides modular support for centralized critics and value decomposition within compatible algorithm families. For instance, the same Q-learning learner can use different value decomposition mixers. 

MARLlib distinguishes three families of algorithms: (1) independent learning, (2) centralized critic, and (3) value decomposition. Its modular organization allows compatible learning algorithms to reuse centralized critic or value decomposition mechanisms instead of implementing the complete centralized training pipeline independently for each algorithm. MARLlib also directly supports different parameter sharing strategies. Policy parameters can be shared among all agents, shared within groups of agents, kept separate for each agent, or assigned through a custom policy-mapping strategy \citep{hu2023marllib}.

\subsubsection{Learning}

The learning module is implemented as an independent component across these tools, except PettingZoo which is not concerned. It is possible to test different learning algorithm and policy on different environments. However, the two sub components: learning algorithm and policy are not always separated. These two components cannot be completely independent, however, if they are separated enough, it allows for learning algorithms to be used with different policies without reimplementing everything. For example, a Q-table based policy could be updated using learning algorithms such as SARSA or Q-learning.

OpenSpiel does not systematically separate these two components. It does not provide dedicated abstractions for policies and learning algorithms that would allow compatible implementations to be combined independently. Moreover, in some algorithms, the same component is responsible for both selecting actions and updating the action selection process.

RLlib has two specific components for the learning algorithm and the policy. It allows for using learners with different policies that are compatibles.

MARLlib separates policy networks from optimization code, but it does not expose policies and learning algorithms as two fully independent components. This is partly due to its reliance on an older version of RLlib. In contrast, the redesign of RLlib explicitly aimed at improving the separation of concerns through a more modular architecture. 

EPyMARL separates these two components. However, the policy itself is distributed across two components: a centralized controller that calls the agents' neural network and converts its outputs into actions, and the neural network that computes the outputs used for action selection.

\subsubsection{Agent}

Having an agent component that manages policies, learning, and potentially other functions, such as communication, enables the implementation of approaches in which the agent itself performs additional operations before selecting an action or initiating learning (see Section~\ref{sec:core_agent}). It also prevents other components from being overloaded with additional responsibilities. However, most of the analyzed tools provide limited support for the agent as an independent core module. 

In PettingZoo, agents are represented only by identifiers within the environment.

RLlib and MARLlib directly associate agent identifiers with policies, without introducing an intermediate agent component responsible for coordinating the decision process.

EPyMARL instead uses a centralized multi-agent controller that manages action selection for all agents. Consequently, it does not provide each agent with an individual orchestration component that can be extended independently.

OpenSpiel provides a more explicit agent abstraction: the execution loop delegates action selection to the Agent. This design makes it possible to extend the Agent with additional behaviors before returning the final action, for example by selecting among several policies or applying a specialized decision mechanism.

\subsubsection{Communication}
Communication requires messages to be generated, transmitted, and received by agents. Even when communication is available in the analyzed tools, it is not handled by an independent module that can be activated or deactivated as needed.

Some environments provided by PettingZoo include communication capabilities. Agents generate messages during action selection, either as their complete action or as one component of it, while the environment transmits these messages to the receiving agents through their observations. 

OpenSpiel also provides games involving communication. The communication setting is defined by the game itself, and communicating generally constitutes an action of the agent rather than a secondary protocol executed in parallel with its action.

RLlib, MARLlib, and EPyMARL do not provide any native implementation or example of an agent communication system. This absence does not mean that communication cannot be implemented using these tools. For example, an external extension of RLlib\footnote{\url{https://github.com/proroklab/rllib_differentiable_comms}} has been developed to implement a differentiable communication channel between agents, demonstrating that such mechanisms can be added despite not being natively supported \citep{blumenkamp2021adversarial}.

\subsubsection{MOA}
MOA is poorly represented in the MARL tools considered above. Some specialized MARL tools exist, such as LeMOL, but MOA has not been generalized within the main MARL frameworks. 

OpenSpiel provides limited support for MOA through methods such as best response and minimax. However, these mechanisms are tied to specific algorithms and require the agent to have access to a model of the environment dynamics\footnote{Approaches in which the agent uses a model of the environment dynamics to support decision-making are referred to as model-based approaches.}.

\subsubsection{System Evaluator}
The analyzed tools provide different levels of support for system evaluation. 

PettingZoo does not include a dedicated evaluation system, which is consistent with its role as an environment API rather than a complete MARL framework. 

OpenSpiel provides several evaluation functions, particularly for game-theoretic criteria such as Nash equilibrium (see Section~\ref{sec:evaluation_criteria}). 

RLlib and MARLlib provide more flexible support for experiment-specific evaluation. Users can define metrics beyond rewards and returns by processing information exposed by the environment or generated during training. This flexibility makes system evaluation a well-supported core module in these frameworks. 

EPyMARL follows a similar approach and can automatically record statistics reported by the environment. However, computing more complex metrics across multiple steps may require modifying the environment or the execution scheduler, notably the \class{ParallelRunner}.

\subsection{BenchMARL}

The preceding analysis examined existing MARL tools from the perspective of the core modules identified in this paper. However, other studies have highlighted complementary limitations of current MARL tools and experimental practices.
In particular, recent work has emphasized the need for more standardized and reproducible evaluation protocols in MARL \citep{gorsane2022towards}. BenchMARL was introduced as a framework specifically designed to address this problem \citep{bettini2024benchmarl}. It enables users to systematically compare different combinations of tasks, algorithms, and models within a common experimental pipeline. This represents an important step towards better experimental practices in MARL experiments.

BenchMARL itself adopts a modular perspective by distinguishing several components of a MARL experiment, including tasks, algorithms, models, experiment configurations, and benchmark.
A task is an environment scenario that defines the MARL problem to be solved. An algorithm defines all the components that determine the training strategy. A model defines the neural network architecture used by components such as the policy for action selection or the critic for value estimation. An experiment is a training run in which the task, algorithm and model are fixed.  Finally, a benchmark component corresponds to a collection of experiments.

However, this modular decomposition does not explicitly expose several of the core modules identified in this paper, such as communication, MOA, reward modeling, or training and execution modes, despite their importance in the design of MARL systems. These mechanisms consequently remain integrated into the components defined by BenchMARL rather than being independently configurable.

A finer decomposition of such approaches would make it easier to study the contribution of these specific modules in isolation. Such a decomposition would make it possible to test an algorithm while independently varying the modules presented in Section~\ref{sec:marl_core_modules}.

\section{MARLKIT} \label{sec:marlkit}
Having examined the extent to which existing MARL tools support the core module perspective, we now examine the practical implications of the proposed decomposition through its implementation in our MARL framework, MARLKIT.

\begin{figure}
    \centering
    \begin{tikzpicture}[>=latex]

\node (env) [rectangle, draw, minimum width=2.5cm, minimum height=1.5cm, align=center] {Environment};

\node (eval) [rectangle, draw, minimum width=2.5cm, minimum height=2cm, align=center]
    at ($(env.west) + (-2cm,0)$) {System\\Evaluation};

\node (agent) [rectangle, draw, minimum width=2.5cm, minimum height=3cm, align=center]
    at ($(env.east) + (3cm,0)$) {Agent};

\node (te) [rectangle, draw, minimum width=6cm, minimum height=1cm, align=center]
    at ($($(env.north)!0.5!(agent.north)$) + (0,1.7cm)$) {Training and Execution};

\node (reward) [rectangle, draw, minimum width=2.5cm, minimum height=1cm, align=center]
    at ($(env.south) + (0,-1.2cm)$) {Reward};

\node (learning) [rectangle, draw, minimum width=2.4cm, minimum height=0.8cm, align=center]
    at ($(agent.east) + (2.5cm,1cm)$) {Learning};

\node (communication) [rectangle, draw, dashed, minimum width=2.4cm, minimum height=0.8cm, align=center]
    at ($(agent.east) + (2.5cm,0)$) {Communication};

\node (moa) [rectangle, draw, dashed, minimum width=2.4cm, minimum height=0.8cm, align=center]
    at ($(agent.east) + (2.5cm,-1cm)$) {MOA};

\draw[-, thick] ($(te.south) + (-2.13cm,0)$) -- (env.north);
\draw[-, thick] ($(te.south) + (2.13cm,0)$) -- (agent.north);

\draw[-, thick] (agent.west) -- (env.east);
\draw[-, thick] (eval.east) -- (env.west);

\draw[-, thick] (env.south) -- (reward.north);

\draw[-, thick] ($(agent.east) + (0,1cm)$) -- (learning.west);
\draw[-, thick] (agent.east) -- (communication.west);
\draw[-, thick] ($(agent.east) + (0,-1cm)$) -- (moa.west);

\end{tikzpicture}
    \caption{MARLKIT core modules.}
    \label{fig:module_diag}
\end{figure}
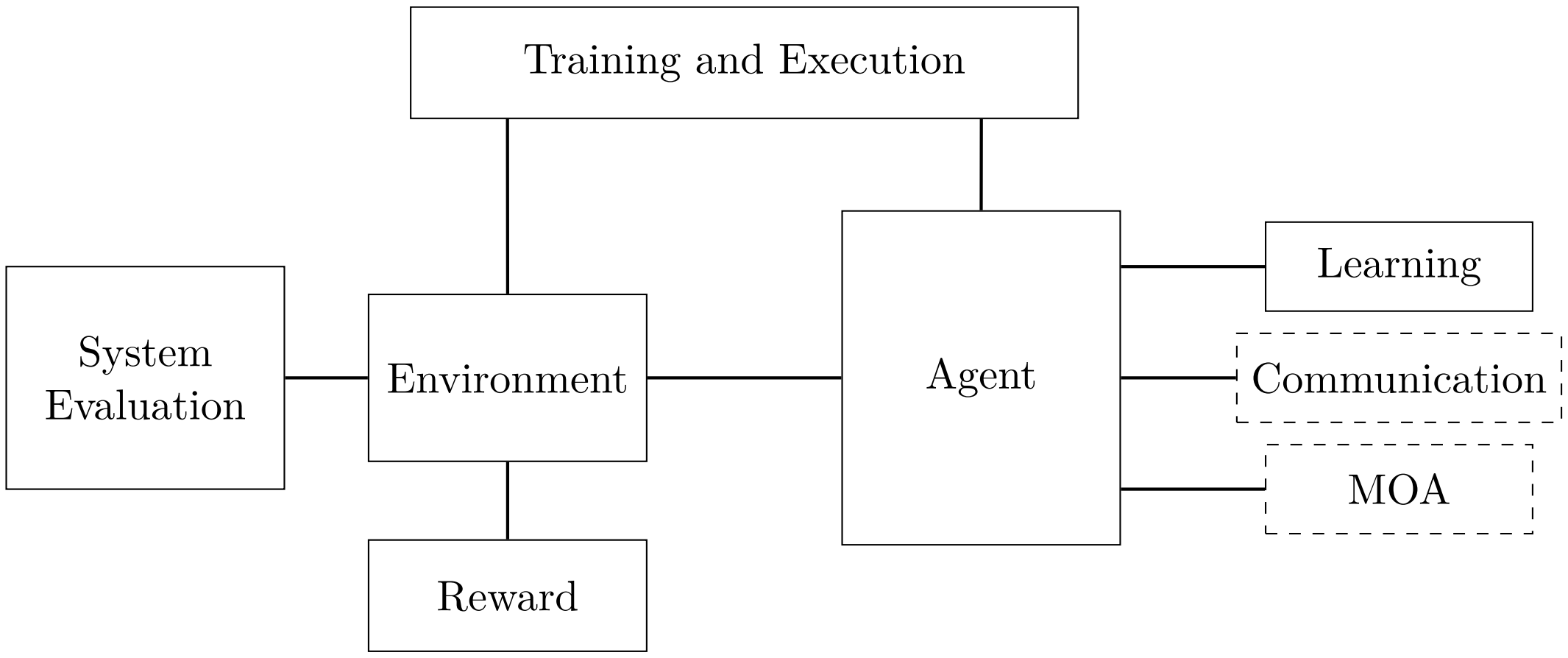

MARLKIT is a Java-based MARL framework that supports the exploration and comparison of MARL methods through experiments involving different implementations of the core modules. The modules presented in Section~\ref{sec:marl_core_modules} are explicitly represented in MARLKIT, making it possible to add new implementations of a module and use them with the rest of the framework. Figure~\ref{fig:module_diag} illustrates the core modules and how they interact in MARLKIT.

MARLKIT was designed according to two main objectives. First, its architecture should be organized around the core modules, exposing them as first-class components. Second, new module implementations should be easy to create, interchangeable with existing implementations, and reusable. Particular attention was also paid to the single responsibility principle, so that each class has a clearly identified responsibility.

MARLKIT serves as a proof of concept illustrating the practical benefits of structuring MARL tools around the proposed core module decomposition. 
Examining how easily module implementations can be added, exchanged, and compared across experiments provides a practical indication of the relevance of the proposed decomposition. In particular, it enables controlled comparisons in which the implementation of a specific module is varied while the remaining experimental configuration is kept unchanged.

Experiments varying representative choices across the core modules provide an initial assessment of how the framework accommodates distinct MARL configurations.

In MARLKIT, each component has a clearly separated responsibility. The framework relies on MaDKit for the simulation and execution of the multi-agent system.

MARLKIT was implemented in Java, despite most MARL tools being python-based. Java offers a stricter class-based architecture with more constraints that help ensure certain design rules are followed. The main purpose of our methodology and decomposition is to provide a clearer and more precisely defined view of concepts such as core modules. In our Java implementation, these concepts are represented by dedicated packages and classes, making their respective responsibilities, operations, and components more explicit. Python bindings could be developed to make the project accessible from Python for the MARL community. It should also be noted that MARLKIT is primarily a proof of concept. The proposed modular approach is not language-specific and could therefore be implemented as a Python-based framework.

The scheduler (\class{MLKScheduler}) orchestrates the overall MARL interaction loop by coordinating calls to the environment, the agents and their associated update methods in a predefined sequence. By default, it enforces a fixed observation–influence-reaction–reward–learning cycle. It can be redefined partly or completely.

In the following sections, we describe how the core modules are implemented in MARLKIT and illustrate their use in experimental settings. The purpose of these experiments is not to achieve state-of-the-art performance, but to examine the decomposition in practice. 

However, exploring alternative solutions in realistic use cases often requires considering multiple implementations across several core modules. Section~\ref{sec:marlkit_exploring} describes how MARLKIT supports batches of simulations involving different combinations of core module implementations, and how the resulting performances can be compared.

    \subsection{Agent}
    
An agent performs actions and influences the environment. \class{MLKAgent} defines the methods required for an agent to participate in the MARL learning loop.

These methods allow the agent to orchestrate its internal core modules. In particular, the agent calls its learning components, such as the policy and the learning algorithm, to select actions and update its behavior from experience. Depending on the considered implementation, the agent may also orchestrate other core modules, such as communication and MOA.

\class{AgentStandard} is a default implementation of \class{MLKAgent}. It implements the standard methods of the MARL learning loop by delegating action selection and learning updates to the learning module, without introducing additional agent-level mechanisms.

    \subsection{Environment}

In MARLKIT, the environment manages the states and dynamics of the system. The implementation of the transition function respect the Influence/Reaction paradigm defined in Section~\ref{sec:env_core}. It receives the agents' influences (\ie joint action) and updates its state according to the reaction.

An environment is represented by the \class{MLKEnvironment} interface, which defines the core methods required by the \class{MLKScheduler}. MARLKIT provides the abstract class \class{EnvironmentStandard}, which implements common methods while leaving the experiment-specific components, namely the states and dynamics, to the user.

        \subsubsection{State}
        
The \class{State} represents the current configuration of the environment. It provides access to different types of data, such as numerical values, and object positions.

The implementation of \class{State} is expected to vary across experiments. However, a common type of state representation is a 2D grid. MARLKIT provides two ready-to-use implementation of \class{State}: \class{State2DGridInt}, which stores integer values in a 2D grid and records agent positions.

        \subsubsection{Reaction Event} \label{sec:reaction_event}
        
The \class{ReactionEvent}s are produced during the reaction phase (2) (see Figure~\ref{fig:influence_reaction}).

\class{ReactionEvent}s represent all meaningful events that occur in the environment as a result of the agents' actions. 

After agents take actions and the environment transitions to a new state, the latter generates a set of \class{ReactionEvent}s for each agent. These \class{ReactionEvent}s represent all events that may influence the reward of the agents or the evaluation of the system. For example, in a foraging environment, a \class{ReactionEvent} can be \textit{"You collected a resource"} or \textit{"You moved"}.

These \class{ReactionEvent}s are not computed independently for each agent, it depends on the joint action and the state transition. For instance, if in the foraging environment, two agents are needed for collecting a resource, then they will receive the \class{ReactionEvent} \textit{"You collected a resource"} only if the other agent also take the action needed for collecting the resource.

\class{ReactionEvent}s are used by the reward model to compute the rewards returned to each agent (see Section~\ref{sec:marlkit_reward_modeling}) and by the system evaluation module to compute different evaluation measures (see Section~\ref{sec:marlkit_system_eval}).

    \subsection{Reward Modeling} \label{sec:marlkit_reward_modeling}
Reward modeling is an independent core module, used by the environment to compute rewards during the reward phase (3) (Figure~\ref{fig:influence_reaction}).

The reward model has reaction events as input and a final reward vector $(r_1, \ldots, r_N)$ as output, where $r_i$ is the reward received by an agent $i$. 

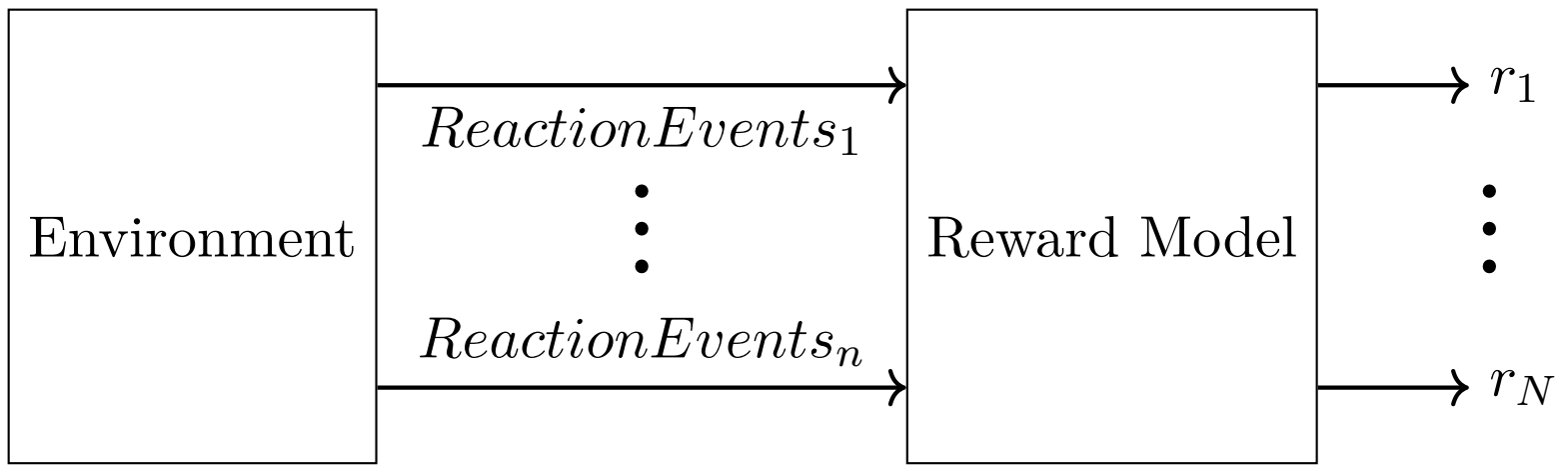
\begin{figure}
    \centering
    \begin{tikzpicture}


\node (env) [rectangle, draw, minimum width=2cm, minimum height=3cm, align=center] {Environment};

\node (rewardmodel) [rectangle, draw, minimum width=2cm, minimum height=3cm, align=center, right=3.5cm of env] {Reward Model};

\draw[->, thick] ($(env.east) + (0, 1cm)$) -- ($(rewardmodel.west) + (0, 1cm)$) node[midway, below] {$ReactionEvents_1$};
\draw[line width=2.5pt, line cap=round, dash pattern=on 0pt off 7pt] ($($(env.east)!0.5!(rewardmodel.west)$) + (0, -0.2cm)$) -- ($($(env.east)!0.5!(rewardmodel.west)$) + (0, 0.3cm)$);
\draw[->, thick] ($(env.east) + (0, -1cm)$) -- ($(rewardmodel.west) + (0, -1cm)$) node[midway, above] {$Reaction Events_n$};

\node (r1) [right=1cm of rewardmodel, yshift=1cm] {$r_1$};
\node (dots) [right=1cm of rewardmodel] {};
\node (rn) [right=1cm of rewardmodel, yshift=-1cm] {$r_N$};

\draw[line width=2.5pt, line cap=round, dash pattern=on 0pt off 7pt] ($(dots) + (0, -0.2cm)$) -- ($(dots) + (0, 0.3cm)$);

\draw[->, thick] (rewardmodel.east) +(0, 1cm) -- (r1.west);
\draw[->, thick] (rewardmodel.east) +(0, -1cm) -- (rn.west);


\end{tikzpicture}
    \caption{Production of the rewards}
    \label{fig:reward_functions}
\end{figure}

        \subsubsection{Reward Notation}

In standard RL, the reward is typically represented as a single scalar value. To support the modular architecture design, we introduce three reward-related quantities. This design allows generic reward models to operate on reward contributions without depending on environment-specific details. 

Each reaction event $e$ carries a \textbf{default event reward} $\rho(e)$. $\rho(e)$ represents an event-level reward contribution and serves as an abstract intermediary between the environment and reward models. 

For each agent $i$, the agent-level \textbf{raw reward} $\bar r_i$ is obtained by summing the default event rewards associated with that agent. Reward models based on reward redistribution can manipulate directly this concept in order to produce the final reward.

The \textbf{final reward} $r_i$ is the reward actually returned to agent $i$ after applying the \class{RewardModel}.

The three reward model presented in Section~\ref{sec:marl_reward_modeling} are good examples of reward models that do not depend on environment-specific details. Their implementations are discussed in the following section.

        \subsubsection{Reward Model}

The \class{RewardModel} is a module that is responsible for computing the reward of each agent. It maps the generated \class{ReactionEvent}s to the final reward received by each agent (see Figure~\ref{fig:reward_functions}). It may aggregate the default event rewards associated with each agent, redistribute rewards across agents, or transform them using additional information from the transition. 

Depending on the \class{RewardModel} used, the reward can be computed in different ways. The main advantage of this design is that it allows to try different reward models without changing the environment dynamics nor having to redesign the reward associated to each \class{ReactionEvent}. 
The three main possible reward models presented in Section \ref{sec:marl_reward_modeling} are implemented in MARLKIT: (1) Mixed Reward Model, (2) Fully Cooperative Reward Model, (3) Fully Competitive Reward Model.

\class{MixedRewardModel} is the default reward model class. With this reward, one agent final reward is the equal to its raw reward ($\forall i, r_i = \bar r_i$).

When the system uses \class{FullyCooperativeReward}, it operates in a fully cooperative setting\footnote{Reaction events and default event rewards do not need to be manually updated; the rewards associated with each \class{ReactionEvent} are averaged and returned to all agents.}.

\class{ZeroSumReward} implements a fully competitive reward model. With this reward model, every times an agent receive a reward value, the same value is taken from the total of the others. This assure the experience is a zero-sum game between the agents.

        \subsubsection{Example} \label{sec:marlkit_rewardmodel_example}

To illustrate how reward models can affect the dynamics of the system on their own, we conduct a collecting resources experiment. In this experiment, agents collect resources from production units with limited stock. We run this experiment with four different \class{RewardModel}s while keeping the rest of the core modules identical. In the \class{collectingresource} package, four different launchers instantiate the same core modules except for the \class{RewardModel}.

The selected evaluation criteria are: total reward, which indicates whether the system reaches a high collective value; lowest individual reward, which is used as a fairness indicator; and the total number of collected resources, which is negatively correlated with the competitiveness of the system.

\begin{figure}[t]
    \centering
    \includegraphics[width=\linewidth]{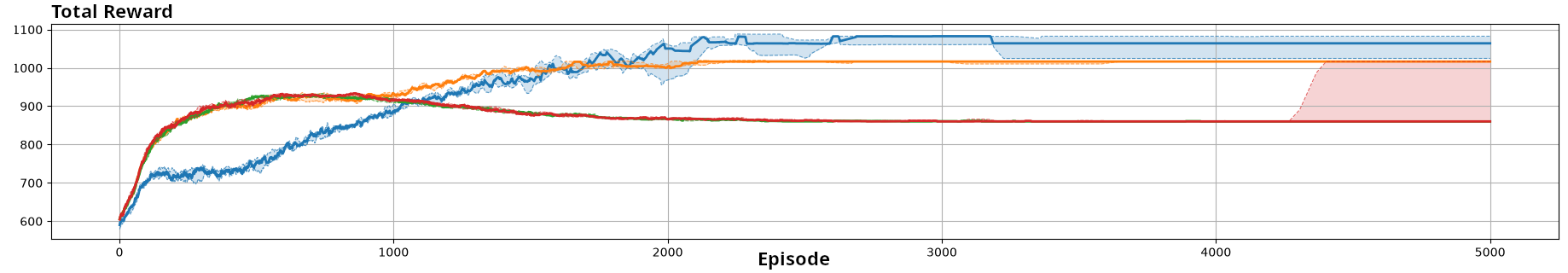}
    \caption{Total of the raw rewards of all agents added together. Green is independent reward, Blue is fully cooperative reward, red is fair mixed reward and orange is logistical reward.}
    \label{fig:collecting_ressource_marlkit}
\end{figure}

Figure~\ref{fig:collecting_ressource_marlkit} illustrates how the resulting system dynamics can strongly differ depending only on the selected reward model. For conciseness, only one evaluation criterion is displayed here, as the purpose of this experiment is to illustrate the modular comparison enabled by MARLKIT rather than to provide a comprehensive evaluation of reward models. The complete results are provided in Appendix~\ref{app:rewardmodel}.

The figure shows that changing only the reward model is sufficient to produce substantially different system-level outcomes.

Additional details about the experimental setup and results are provided in the appendix~\ref{app:rewardmodel} and also in the \texttt{README.md} file of the \class{collectingresource} package. The latter also explains how to run this experiment. A second experiment, detailed in the \texttt{README.md} of the \class{foraging} package, applies two of the reward models evaluated in this article without modifying their implementation, in order to verify their reusability.

    \subsection{Learning} 
The difference between MARLKIT and other MARL tools regarding learning algorithms is that MARLKIT tends to assign less responsibility to the learning algorithm. Following the definition of a learning algorithm introduced in Section~\ref{sec:rl_learning_module}, the \class{Algorithm} is not responsible for generating data, but instead uses available data to update the policy. Therefore, in MARLKIT, the learning algorithm and the policy are two separate classes. \class{Policy} is responsible for action selection as it maps observations to actions, and \class{Algorithm} manages the update of the policy based on the agent's experience. 

Policies and learning algorithms remain interdependent. Some learning algorithms may require specific policy architectures, and conversely, some policies may require specific learning algorithms in order to be properly updated. This separation is designed to be flexible and avoid the need to implement the same policy multiple times. For example, SARSA (\class{Sarsa}) and $Q$-learning (\class{QLearning}) can be implemented as two distinct learning algorithms that both update the same $Q$-table based policy (\class{QValueBasedPolicy}).

    \subsection{Training and Execution Modes}
    
The training and execution mode determines whether agents operate in a centralized or decentralized manner during each phase. In MARLKIT this is managed by a dedicated module: \class{TrainingExecutionStrategy}. The scheduler invokes agent operations through this module, which determines whether they are performed in a centralized or decentralized manner.

CTE implies that agents are trained and executed using centralized information. It is the counterpart of DTE, where agents operate in a decentralized manner during both training and execution. CTDE consists of training agents with centralized information while executing them without access to such information.
    
        \subsubsection{DTE}
DTE is the default mode in MARLKIT. The default \class{TrainingExecutionStrategy} implements decentralized agent operations during both training and execution. DTE therefore serves as the baseline configuration, while other strategies can introduce centralized operations without modifying the scheduler.

        \subsubsection{CTE and CTDE}
CTDE and CTE require the introduction of centralization mechanisms, either during training only (CTDE) or both training and execution (CTE). The mechanisms used to introduce centralization may be identical in both CTDE and CTE\footnote{As explained in Section \ref{sec:cte}, one particular form of CTE is when a single centralized entity, a "super-agent", pilots all the agents. In such a setting, the most natural implementation is to define a single super agent that directly outputs a \class{JointAction}.}. They differ in the phase during which centralized information is accessible: training only for CTDE, and both training and execution for CTE.

In MARLKIT, these mechanisms are implemented through dedicated \class{TrainingExecutionStrategy} implementations, which are used by the scheduler without requiring modifications to it.

For instance, \class{CentralizedCriticTrainingExecutionStrategy} implements centralized training by providing critics with centralized information from multiple agents.
Other agent operations can also be centralized, such as observation generation or policy updates.

        \subsubsection{Example}\label{sec:marlkit_ctde_example}

To illustrate how training and execution modes can vary, we conduct a prey hunter experiment. Two hunters must jointly capture a non-learning prey by both moving within close range of it. We run the experiment using two centralization configurations while keeping all other core modules identical. In the \class{preyhunter} package, separate launchers instantiate the same core modules, differing only in their centralization mechanism.

The selected evaluation criteria are the capture rate, defined as the proportion of episodes in which the hunters successfully capture the prey; the capture time, measuring the time required to achieve capture; and the total reward, reflecting the overall performance of the hunter team.

\begin{figure}[t]
    \centering
    \includegraphics[width=\linewidth]{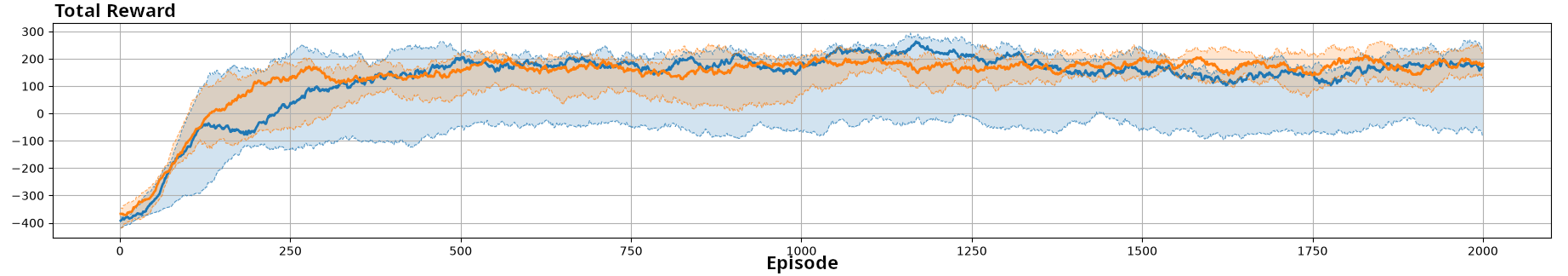}
    \caption{Total of the raw rewards of all agents added together. Blue is DDPG and orange is MADDPG.}
    \label{fig:preyhunter_ddpg_marlkit}
\end{figure}

Figure~\ref{fig:preyhunter_ddpg_marlkit} illustrates how the selected training mode affects the resulting system dynamics. A more comprehensive evaluation is presented in Appendix~\ref{app:training_execution}. The appendix also provides details about the experimental setup, while the \texttt{README.md} file of the \class{preyhunter} package provides additional implementation details and instructions for running the experiment.

    \subsection{Communication}
Communication provides agents with an additional mechanism for interaction by allowing them to exchange information.
Communication in MARLKIT is designed as an optional module. Rather than embedding a fixed communication mechanism directly into the agents, the framework separates the ability to communicate from the communication strategy itself. 
This separation allows different communication protocols to be integrated, replaced, or compared without modifying the core agent implementation nor the simulation scheduler.

        \subsubsection{Communication Model}

\class{CommunicationModel} implements a specific communication strategy. It determines which information is transmitted, which agents are targeted, and how incoming messages are interpreted.

This design makes communication strategies interchangeable: the same agent can be evaluated with different communication mechanisms, while the scheduler and the learning components remain unchanged.

For example, \class{BroadcastObservation} implements a simple observation-sharing mechanism. During the communication step, it broadcasts the agent's local observation to all other agents. During the reception step, it enhances the agent's observation using the observations received from other agents. Such a mechanism can help mitigate partial observability issues by allowing agents to share their local observations.

        \subsubsection{MLKAgentCommunicating}
Similarly to \class{MLKAgent}, which defines the methods used to connect an agent with its learning module, \class{MLKAgentCommunicating} defines the methods which are useful for using the \class{CommunicationModel}.

Agents that implement the \class{MLKAgentCommunicating} interface expose the minimal capabilities required to participate in communication. Such agents rely on \class{CommunicationModel} associated with it.

The communicating agent itself does not define a specific protocol. Instead, it delegates two operations to its communication model. The first operation corresponds to the emission phase, where the agent sends messages to other agents. The second operation corresponds to the reception phase, where the agent handles the messages received from other agents and updates its internal state accordingly.

        \subsubsection{Example}\label{sec:marlkit_communication_example}

To illustrate how communication mechanisms can affect the dynamics of the system on their own, we conduct a prey hunter experiment. This experiment and the evaluation criteria are the same as the ones described in Section~\ref{sec:marlkit_ctde_example}.

The four tested communication configurations are: no communication, observation sharing, policy parameter sharing, and both observation and policy parameter sharing.

\begin{figure}[t]
    \centering
    \includegraphics[width=\textwidth]{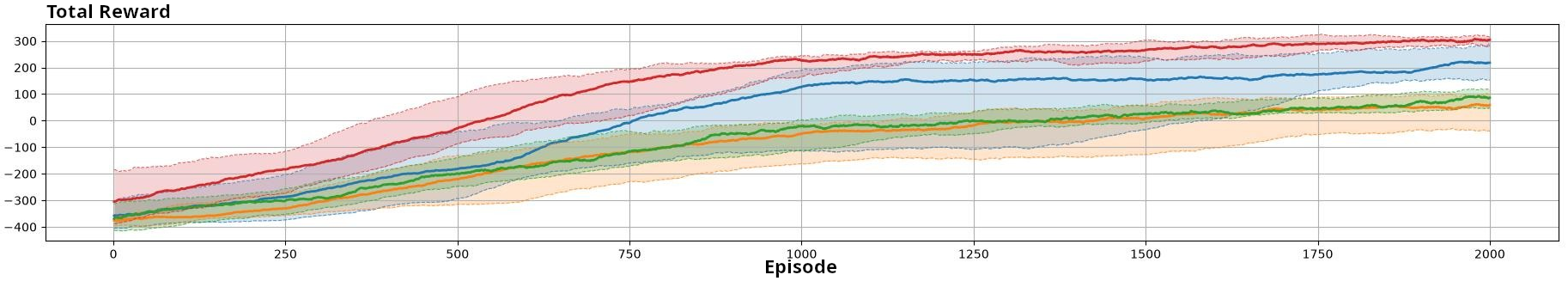}
    \caption{Total of the raw rewards of all agents added together. Orange is no communication, green is policy parameter sharing, blue is observation sharing and red is both observation and policy parameter sharing.}
    \label{fig:prey_hunter_com_marlkit}
\end{figure}

Figure~\ref{fig:prey_hunter_com_marlkit} illustrates how the resulting system dynamics differ depending only on the selected communication mechanism. 

Additional details about the experimental setup and the tested configurations are provided in the appendix \ref{app:communication} and in the \texttt{README.md} file of the \class{preyhunter} package.

    \subsection{MOA}
Modeling other agents can help an agent predict others' behavior and better account for their influence on the environment dynamics (see Section~\ref{sec:moa}).
    
MARLKIT also provides support for agents that maintain explicit models of other agents. This functionality is treated as an optional component of the agent architecture: agents that need such models can include them, while simpler agents can ignore this mechanism entirely. This separation allows different modeling approaches to be integrated, replaced, or compared without modifying the core agent implementation. However, policies that exploit these predictions may need to be adapted to incorporate them into the action-selection process (see Figure~\ref{fig:marl-policy-extended-input}).

        \subsubsection{ModelsManager}
\class{ModelsManager} is responsible for managing the models. It provides an interface for accessing and updating the models used to reason about other agents.

The predictions exposed by the \class{ModelsManager} can then be used by different components of the framework. In particular, compatible policies may use them during action selection. For example, joint-action policies can evaluate candidate actions together with predicted actions of the other agents.

\class{PredictionModelsManager} is a specific class that implements \class{ModelsManager}. It is designed to manage predictive models of others' actions. 

This structure supports different ways of predicting other agents' actions. For instance, it is possible to maintain one model per agent like in fictitious play. Alternatively, a second approach consists in modeling the other agents at the group level, by directly predicting a joint action. Minimax-based approaches follow this second approach by predicting the worst-case joint action of the opponents. 

        \subsubsection{MLKAgentModelingOthers}
Similarly to both \class{MLKAgent} and \class{MLKAgentCommunicating}, that define methods that manipulate, respectively, the learning module and the communication model, \class{MLKAgentModelingOthers} defines the methods which are useful for using the a \class{ModelsManager}. The responsibility of managing the models is delegate to the \class{ModelsManager}. 

A model can be used for different purposes, one of them is to predict the behavior or actions of other agents. \class{MLKAgentModelingOthersAction} is an interface for agents that maintain predictive models of other agents' actions. This interface relies on \class{PredictionModelsManager}, a specific type of \class{ModelsManager} that manages predictive models. 

\subsubsection{Example} \label{sec:marlkit_example_moa}

To illustrate how MOA implementations can affect the dynamics of the system on their own, we conduct a \textit{Go or Yield} experiment. In this experiment, two agents (cars) repeatedly choose between \textit{Go} and \textit{Yield}. We test one agent with different MOA implementations while keeping the implementations of all other modules unchanged.

The tested MOA configurations are:
\begin{itemize}
    \item Independent: the agent does not model the opponent;
    \item JAL-AM with Action Frequencies: the agent evaluates joint actions and predicts the opponent's action based on the opponent's recent actions;
    \item JAL-GT with Minimax: the agent evaluates joint actions and assumes the worst-case action from the opponent.
\end{itemize}

The selected evaluation criteria measure the proportion of episodes associated with each possible joint action: both agents selecting \textit{Go}, resulting in a collision; both selecting \textit{Yield}, resulting in both cars stopping; or one selecting \textit{Go} while the other selects \textit{Yield}, allowing one car to pass.

\begin{figure*}[t]
    \centering
    \includegraphics{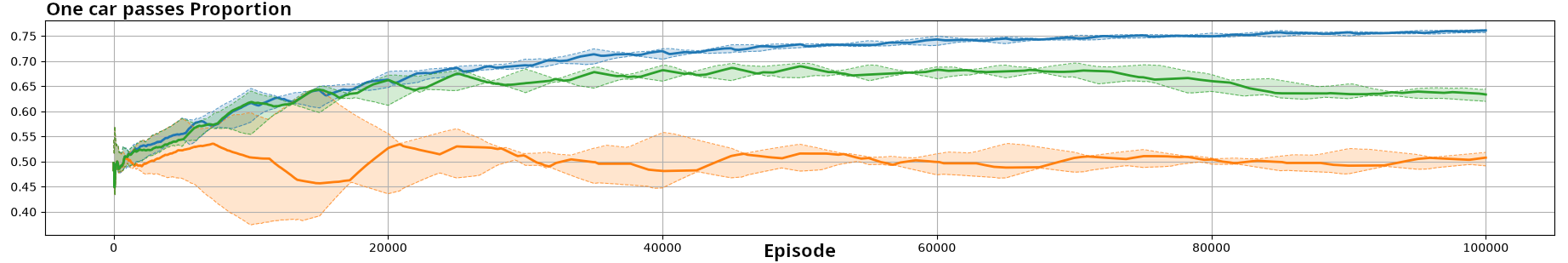}
    \caption{Proportion of episodes in which one car passes, with one agent selecting \textit{Go} and the other \textit{Yield}.}
    \label{fig:gooryield_goyield_marlkit}
\end{figure*}

Figure~\ref{fig:gooryield_goyield_marlkit} illustrates how changing only the MOA implementation leads to different learning dynamics. A more comprehensive analysis of these dynamics, along with further details about the experimental setup, is provided in Appendix~\ref{app:moa}. The \texttt{README.md} file of the \class{gooryield} package also provides details about the setup and explains how to run the experiment.

    \subsection{System Evaluation} \label{sec:marlkit_system_eval}

As explained in Section~\ref{sec:evaluation_criteria}, MARL experiments should be evaluated at the system level by considering the overall performance of the agents as a collective. Because MARLKIT aims to support the exploration of different solutions, comparisons between solutions should be based on explicit metrics. MARLKIT therefore provides a dedicated mechanism for facilitating such evaluation.

\begin{figure}
    \centering
    \begin{tikzpicture}

\node (env) [rectangle, draw, minimum width=2cm, minimum height=3cm, align=center] {Environment};

\node (systemevaluator) [rectangle, draw, minimum width=2cm, minimum height=3cm, align=center, right=3.5cm of env] {System Evaluator};

\draw[->, thick] ($(env.east) + (0, 1cm)$) -- ($(systemevaluator.west) + (0, 1cm)$) node[midway, below] {$ReactionEvents_1$};
\draw[line width=2.5pt, line cap=round, dash pattern=on 0pt off 7pt] ($($(env.east)!0.5!(systemevaluator.west)$) + (0, -0.2cm)$) -- ($($(env.east)!0.5!(systemevaluator.west)$) + (0, 0.3cm)$);
\draw[->, thick] ($(env.east) + (0, -1cm)$) -- ($(systemevaluator.west) + (0, -1cm)$) node[midway, above] {$ReactionEvents_n$};

\node (m1) [right=1cm of systemevaluator, yshift=1cm] {$Measure_1$};
\node (dots) [right=1cm of systemevaluator] {};
\node (mn) [right=1cm of systemevaluator, yshift=-1cm] {$Measure_j$};

\draw[line width=2.5pt, line cap=round, dash pattern=on 0pt off 7pt] ($(dots) + (0, -0.2cm)$) -- ($(dots) + (0, 0.3cm)$);

\draw[->, thick] (systemevaluator.east) +(0, 1cm) -- (m1.west);
\draw[->, thick] (systemevaluator.east) +(0, -1cm) -- (mn.west);

\end{tikzpicture}
    \caption{System evaluation process.}
    \label{fig:system_evaluator}
\end{figure}
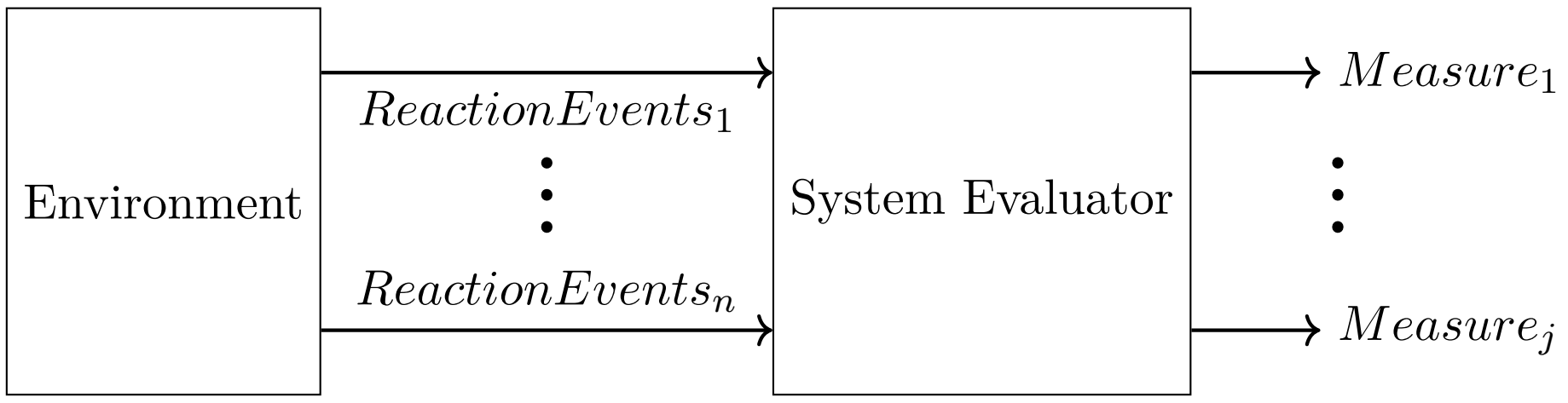

Similarly to reward computation, the default system evaluation mechanism is based on \class{ReactionEvent}s (see Section~\ref{sec:reaction_event}) because \class{ReactionEvent}s describe relevant events that occur in the environment (see Figure~\ref{fig:system_evaluator}).

The \class{Measure} used for system evaluation depend on the environment and on the objectives of the experiment. As discussed in Section~\ref{sec:evaluation_criteria}, different MARL settings may require different evaluation criteria. For instance, the sum of all agents' rewards can be used to evaluate global performance, fairness measures\footnote{Such as the one defined in Equation~\ref{eq:fairness}.} can assess how equitable the distribution of outcomes is among agents, and episode duration can capture the efficiency with which the collective task is completed.

At the end of each episode, the \class{Measure}s are computed and automatically added to a log file. The default \class{SystemEvaluator} updates these measures from the \class{ReactionEvent}s produced by the environment, which allows the evaluation procedure to remain independent of the other modules. This makes it possible to compare different implementations of reward models, learning algorithms, communication mechanisms, or models of other agents while keeping the same evaluation procedure. However, this design is not restricted to event-based measures: the \class{SystemEvaluator} can be extended to update measures from other sources, such as the number of communication messages exchanged during an episode.

    \subsection{Exploring Solutions} \label{sec:marlkit_exploring}
The previous examples illustrate how different implementations of a core module can be explored and compared while keeping the remaining modules unchanged. However, each experimental configuration was defined and executed manually. One important benefit of BenchMARL is that it facilitates the exploration of different systems by allowing multiple experiments with different configurations to be launched together. Such functionality significantly improves the exploration of possible configurations. MARL frameworks designed for this purpose should therefore make it easy to define, execute and compare multiple experimental configurations. MARLKIT provides this capability.

For each core module identified in this paper, MARLKIT provides a corresponding configuration class, such as \class{LearningModule} or \class{CommunicationModule}. Together, these configurable modules define the complete setup of an experiment through an \class{ExperimentConfiguration}. Multiple experiment configurations can then be executed by a dedicated runner, which collects the measures produced by the selected system evaluator for each experiment.

The package \class{multipleexperiences} provides examples of such configurations, along with a \texttt{README.md} file explaining how to define and run them.

\section{Conclusion}
As prior research has highlighted, there are difficulties in comparing MARL systems and experiments in a clear and structured way. Considering this issue, in this paper we argue that a key obstacle lies in the tendency to describe MARL approaches solely through their proposed \textit{algorithms}. This narrow framing obscures the underlying architectural aspects that shape the actual behavior of MARL systems. Consequently, this monolithic perspective hinders our ability to trace how specific modeling choices affect the dynamics of the system.

To address this issue, this paper advocated the adoption of a modular perspective in MARL system modeling. In particular, beyond standard software engineering benefits, our primary contribution is the formulation of a modular perspective that (1) makes all underlying modeling choices explicit when designing MARL systems and (2) facilitates the structured comparison and exploration of diverse modeling choices.

To this end, in this paper we have introduced a modular perspective for modeling MARL systems through a set of core modules. We argue that MARL experimentation would benefit from such a modular design.
This decomposition provides a common language to discuss MARL systems and facilitates the comparison of MARL approaches. The core modules identified in this paper are: (1) agent, (2) environment, (3) learning, (4) reward modeling, (5) training and execution modes, (6) communication, (7) modeling of other agents, and (8) evaluation criteria.

Many MARL experiment exhibit high variance in their implementations and inconsistent baseline, making approaches difficult to compare. Existing tools have made important progress toward more standardized experimentation, but they do not always expose MARL-specific design choices as independent and reusable components. To address this limitation, we introduced MARLKIT, a proof of concept tool for MARL experimentation that implements the core modules identified in this paper as independent first-class components.

MARLKIT was designed to be modular and extensible. It allows users to easily design new module implementations and compare them with existing ones. By facilitating the isolation, reuse, and comparison of core modules, this tool supports a more systematic exploration of design choices in multi-agent systems. These capabilities provide practical evidence for the relevance of organizing MARL modeling choices around the proposed core modules.

At this stage, MARLKIT already provides several implementations for each core module, but it is not intended to be complete. In particular, many MARL algorithms and many module implementations are not yet available. MARLKIT should therefore be understood as a proof of concept demonstrating the relevance of the proposed modular decomposition, rather than as an exhaustive MARL library equipped with state-of-the-art algorithms.

Although MARLKIT currently serves primarily as a proof of concept, we believe it provides a promising foundation for a more mature and practically useful framework. Future work should therefore extend its capabilities and robustness, while using these developments to further substantiate the relevance of a modular perspective for MARL tools.
Future work should first focus on using MARLKIT to conduct detailed experimental studies for each core module, in order to assess the advantages, limitations, and interactions of different implementations. Such studies would help clarify how specific design choices affect MARL performance, reproducibility, and robustness. 

Another important direction is to enrich MARLKIT with additional module implementations, especially learning algorithms, so as to progressively turn the current proof of concept into a more complete experimental framework.

This paper establishes initial links between the proposed core modules and the main MARL challenges, but these relationships should be further investigated in future work. Non-stationarity, partial knowledge, credit assignment, and scalability do not affect all modules in the same way. Making these relationships more explicit would help researchers identify which parts of a MARL system should be modified, extended, or compared depending on the difficulties encountered in a given use case.

\section{Thanks}
This work is part of the Slam-B project within the exploratory PEPR FairCarboN and has benefited from funding provided by the French government, managed by the National Research Agency under the France 2030 initiative, with reference ANR-22-PEXF-0003.

This work received support from the French government, managed by the National Research Agency, under the France 2030 program, reference ANR-16-CONV-0004.

\bibliography{ref}

\appendix
\section{Experimentation Details}
All the experiments presented in Section~\ref{sec:marlkit} was run with ten different seeds. The following give more details about each experiments.

In all these experiment, the agents' behavior follows the default behavior presented in Section~\ref{sec:core_agent_default}. Each agent has one policy and one learning algorithm, which are called when required.

    \subsection{Reward Model Experimentation} \label{app:rewardmodel}
The example presented in Section~\ref{sec:marlkit_rewardmodel_example} consists of four variants that differ only in the implementation of the reward model module. The implementations of the remaining modules are detailed below.

The environment dynamics follow those of the collecting resource experiment. Entities produce resources, and agents can collect one unit of a resource at each timestep from their selected production unit. In MARLKIT, this is implemented in the \class{EnvCollectingResource} class.

For the learning module, the learning algorithm is Q-learning, a value-based algorithm based on a Q-table \citep{jang2019q}. For each state-action pair, the table stores a value corresponding to the expected return. The update follows the equation: $Q_{t+1} = Q_t + \alpha \cdot (r_t + \gamma \cdot Q_\max - Q_t)$, where $Q$ is the table value, $\alpha$ is the learning rate, and $\gamma$ is the discount factor. The value of $\alpha$ is set to $0.2$, and $\gamma$ is set to $0.95$. The policy is based on the Q-table and selects the action associated with the highest value. The policy is epsilon-greedy, and the $\epsilon$ parameter follows the equation:
\begin{equation}
    \epsilon = (1-d)^{t}
    \label{eq:epsilongreedy_rewardmodel_appendix}
\end{equation}
where $d$ is the decay rate, set to $3\cdot10^{-3}$, and $t$ is the number of episodes. Thus, the $\epsilon$ parameter decreases exponentially.
    
Training and execution are completely decentralized (DTE). There is no communication between agents, and agents do not build or use models of other agents.

To evaluate the system, three metrics were selected. The evaluation criteria are the total reward, which indicates whether the system reaches a high collective value; the lowest individual reward, which is used as a fairness indicator; and the total number of collected resources, which is negatively correlated with the competitiveness of the system.

The four tested \class{RewardModel}s are: 
\begin{itemize}
    \item Independent reward (\class{MixedReward}): the final reward received by each agent is equal to its raw reward, see Section~\ref{sec:reward_modeling_mixed}.
    \item Fully cooperative reward (\class{FullyCooperativeReward}): rewards are averaged and returned to all agents, see Section~\ref{sec:fully_cooperative}.
    \item Fair mixed reward (\class{FairMixedReward}): extends the independent reward model with a fairness term. It first computes the raw reward of each agent and identifies the lowest raw reward $\bar{r}_{\min}$. Then, the final reward for an agent $i$ is computed as follows: $r_i = (1-\delta)\times\bar{r}_i + \delta\times\bar{r}_{\min}$, with $\delta \in [0,1]$ the fairness parameter.
    \item Logistical reward (\class{LogisticalRewardModel}): an environment-specific reward model that applies a penalty when agents request too many resources.
\end{itemize}

Reward models manipulate the default event reward $\rho$ to generate the final reward $r$ for each agent (see Section~\ref{sec:marlkit_reward_modeling}). The following list presents the possible events in this experiment and their associated default rewards:
\begin{itemize}
    \item \textit{Collect Resource} : This event is associated with an agent when it collects a resource. The corresponding reward is $\rho(e_{\text{collect\_resource}}) \in \{0, 3, 5, 30\}$, depending on the resource acquired. The distribution of the collected resources' values is highly unequal. Because high-value resources are produced in limited quantities, this creates incentives for greater competition.
    \item \textit{Distance Penalty} : This event is associated with an agent when it travels to collect a resource. The value of $\rho(e_{\text{distance}})$ is negative and depends on the distance traveled. In this scenario, this penalty cannot be lower than $-1.5$.
\end{itemize}

\begin{figure}[htbp]
    \centering
    \begin{subfigure}[t]{\textwidth}
    \includegraphics{figure/collectingressource_total.png}
    \caption{Total of the raw rewards of all agents added together.}
    \label{fig:collecting_total}
    \end{subfigure}
    
    \begin{subfigure}[t]{\textwidth}
    \includegraphics{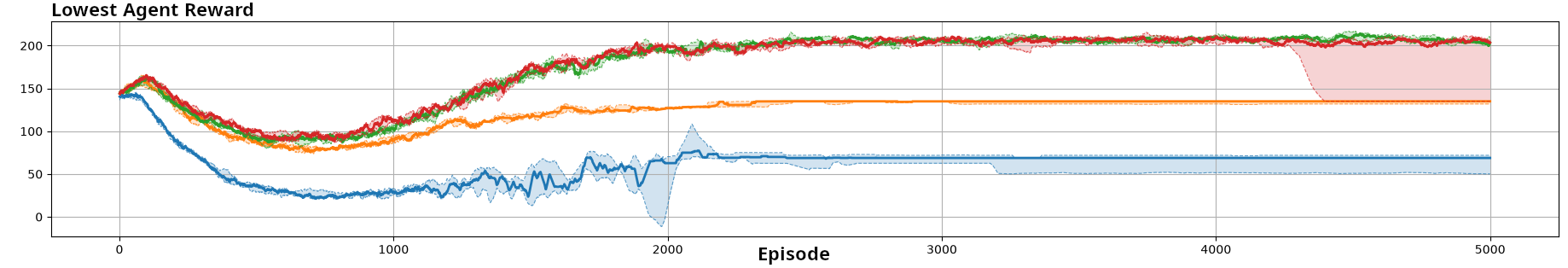}
    \caption{Lowest individual raw reward received by an agent.}
    \label{fig:collecting_lowest}
    \end{subfigure}
    
    \begin{subfigure}[t]{\textwidth}
    \includegraphics{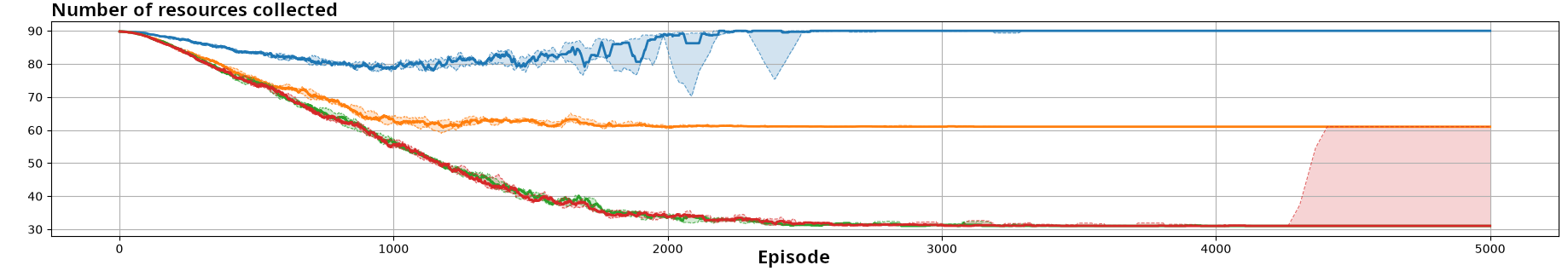}
    \caption{Number of collected resources.}
    \label{fig:collecting_nb}
    \end{subfigure}
    
    \caption{All collecting resources results. Green is independent reward, blue is fully cooperative reward, red is fair mixed reward and orange is logistical reward.}
    \label{fig:collecting_resources}
\end{figure}

Figure~\ref{fig:collecting_resources} presents the results obtained with the different reward models according to the selected evaluation criteria. The curves represent the median across all runs, while the colored areas indicate the range between the first and third quartiles. The data are smoothed using a moving average with a window size of $100$.

The fully cooperative setting results in the highest total reward, which was expected because this reward model makes this metric the common objective of all agents. Therefore, it also maximizes the total number of collected resources because global reward maximization eliminates competition between agents for resource collection. However, this reward model also increases inequality in this experiment: one agent ends up always collecting the resource associated with the smallest positive reward. 

In this experiment, the independent and fair reward models produce similar results: the lowest total reward and the smallest number of collected resources, but a better balance between agents' rewards. Completely egoistic behaviors lead agents to converge toward a balanced distribution of rewards, which explains why the independent and fair reward models produce similar results.

The logistical reward model penalizes competition. It leads to intermediate results between the fully cooperative and independent reward models. Two agents ended up competing for the highest value resource, while the remaining agent collects the second most valuable resource.

    \subsection{Training and Execution Mode Experimentation} \label{app:training_execution}
Section~\ref{sec:marlkit_ctde_example} presents an example in which decentralized and centralized training and execution are compared. The two presented algorithms are DDPG and MADDPG. 

The environment dynamics follow a prey hunter experiment. One prey moves randomly, and two hunters must both be close enough to catch it. Each hunter's action is a continuous two-dimensional movement vector generated by its policy. The hunters have a limited viewing range. The episode ends when either the prey has been caught or $100$ timesteps have passed without a capture.

The selected reward model is the mixed reward model described in Section~\ref{sec:reward_modeling_mixed}. This reward model aggregates the default rewards $\rho$ associated with the events experienced by an agent to determine its final reward. The possible events and their corresponding default rewards are defined as follows:
\begin{itemize}
    \item \textit{Prey Caught}: This event is associated with an agent when it catches the prey. The corresponding event default reward is $\rho(e_{\text{prey\_caught}}) = 300$.
    \item \textit{Distance}: This event is associated with an agent and depends on its distance from the prey. When the agent is close to the prey, the associated reward satisfies $0\leq\rho(e_{\text{distance\_}})\leq2$. When it is too far from the prey, it receives a negative reward given by $\rho(e_{\text{distance\_}}) = -0.5 \times d$ where $d$ denotes the distance between the agent and the prey.
\end{itemize}

Agents do not communicate information through messages, the communication module is set to no communication.

To evaluate the system, three metrics are used:
\begin{itemize}
    \item Total reward: The sum of the raw rewards received by all agents.
    \item Capture success rate: The proportion of episodes in which the prey is caught.
    \item Capture time: The number of timesteps required to catch the prey. This metric is set to $100$ when the prey is not caught.
\end{itemize}

The learning algorithm is Deep Deterministic Policy Gradient (DDPG), an actor-critic learning algorithm (see Section~\ref{sec:actor-critic}) designed for continuous action spaces. \class{DDPG} learns a deterministic policy and trains its actor and critic from experiences sampled from a replay buffer. \class{MADDPG} is a variation of DDPG based on centralized critics. To train the critic, agents also need models of the other agents' policies to predict their actions. More details about MADDPG are provided in Section~\ref{sec:maddpg_paper}. In MARLKIT, both algorithms use the \class{MLPDeterministicPolicy} class as their policy.

DDPG and MADDPG use the same hyperparameters. The actor has two hidden layers of $64$ neurons, while the critic has one hidden layer of $64$ neurons. The actor and critic learning rates are set to $10^{-4}$ and $10^{-3}$, respectively. The discount factor $\gamma$ is set to $0.99$, and the target-network update coefficient $\tau$ to $0.005$. Training uses batches of $64$ experiences sampled from a replay buffer with a capacity of $100 000$ experiences. Exploration is performed by adding Gaussian noise to the actor's actions. The standard deviation for the Gaussian noise is set $0.4$.

\begin{figure}[htbp]
    \centering
    \begin{subfigure}[t]{\textwidth}
    \includegraphics{figure/preyhunter_ddpg_totalreward.png}
    \caption{Total of the raw rewards of all agents added together.}
    \label{fig:preyhunterddpg_total}
    \end{subfigure}
    
    \begin{subfigure}[t]{\textwidth}
    \includegraphics{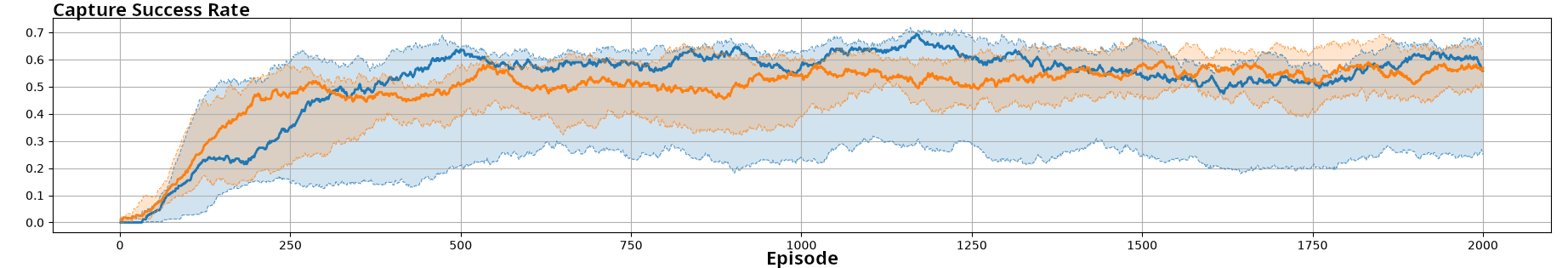}
    \caption{Proportion of episodes in which the prey is caught}
    \label{fig:preyhunterddpg_capturerate}
    \end{subfigure}
    
    \begin{subfigure}[t]{\textwidth}
    \includegraphics{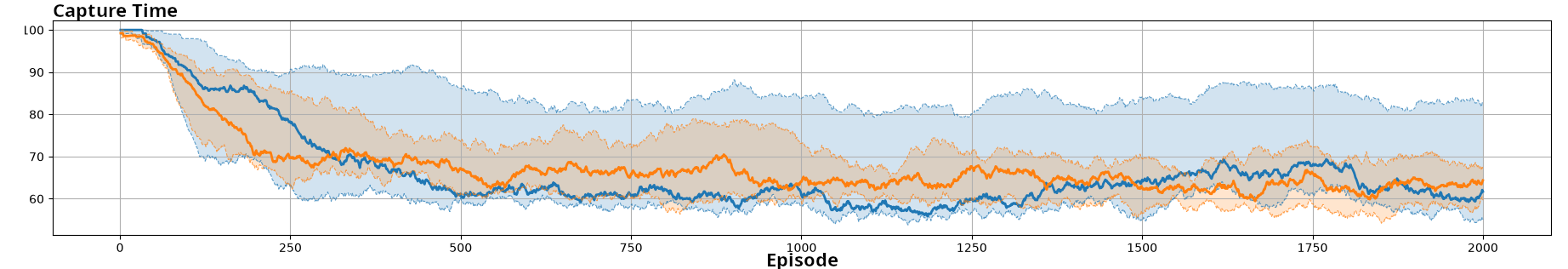}
    \caption{Number of timesteps required to catch the prey.}
    \label{fig:preyhunterddpg_capturetime}
    \end{subfigure}
    
    \caption{Results of the prey-hunter experiment. Blue is DDPG and orange is MADDPG.}
    \label{fig:prey_hunter_ddpg}
\end{figure}

Figure~\ref{fig:prey_hunter_ddpg} presents the results obtained with DDPG and MADDPG, corresponding respectively to decentralized and centralized training. The curves represent the median across all runs, while the colored areas indicate the range between the first and third quartiles. The data are smoothed using a moving average with a window size of $100$.

In this experiment, DDPG and MADDPG present similar median performance across the selected evaluation criteria. However, MADDPG produces less dispersed results, as indicated by the smaller distance between the first and third quartiles.

    \subsection{Communication Experimentation} \label{app:communication}
Section~\ref{sec:marlkit_communication_example} presents an example in which four different communication modes are compared.

The environment dynamics follow the same prey hunter experiment as the one described in Appendix~\ref{app:training_execution}. The reward model and evaluation criteria are also unchanged.

For the learning module, the learning algorithm is \class{PPOCategorical}, an implementation of Proximal Policy Optimization (PPO), a policy-based algorithm \citep{schulman2017proximal}. 
PPO updates the policy by increasing the probability of successful actions and decreasing the probability of unsuccessful ones. It limits how much the policy can change during each update, making learning more stable.
$\epsilon$ is the clipping parameter and is set to $0.2$. The learning rate is set to $0.001$, and the discount factor $\gamma$ is set to $0.95$. The policy is updated over $4$ epochs at each learning step. 
The policy is implemented by the \class{NeuralNetworkCategoricalPolicy} class and is based on a neural network composed of two hidden layers containing $32$ neurons each. The network produces a categorical probability distribution over the possible actions. Actions are selected using a softmax function with a temperature parameter set to $1.0$. The PPO algorithm improves the policy while preventing excessively large updates through the clipped objective.

Training and execution are completely decentralized (DTE), and agents do not build or use models of other agents.

This experiment compares different communication modes. The four communication modes considered are: 
\begin{itemize} 
    \item \textit{No communication}: The agents do not communicate with one another. 
    \item \textit{Observation sharing}: The agents broadcast their observations to the other agents. They then use the received messages to enrich their own observations. 
    \item \textit{Policy parameter sharing}: The agents broadcast their policy parameters. Each agent then averages its own parameters with those received from the other agents and uses the resulting averaged parameters. 
    \item \textit{Observation and policy parameter sharing}: This mode combines observation sharing with policy parameter sharing. 
\end{itemize}

\begin{figure}[htbp]
    \centering
    \begin{subfigure}[t]{\textwidth}
    \includegraphics{figure/preyhuntercom_total_reward.png}
    \caption{Total of the raw rewards of all agents added together.}
    \label{fig:preyhunter_total}
    \end{subfigure}
    
    \begin{subfigure}[t]{\textwidth}
    \includegraphics{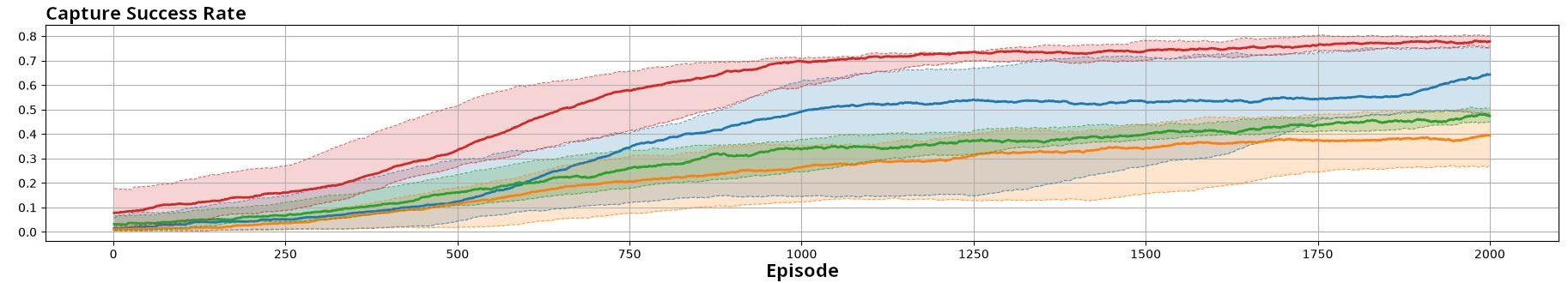}
    \caption{Proportion of episodes in which the prey is caught}
    \label{fig:preyhunter_capturerate}
    \end{subfigure}
    
    \begin{subfigure}[t]{\textwidth}
    \includegraphics{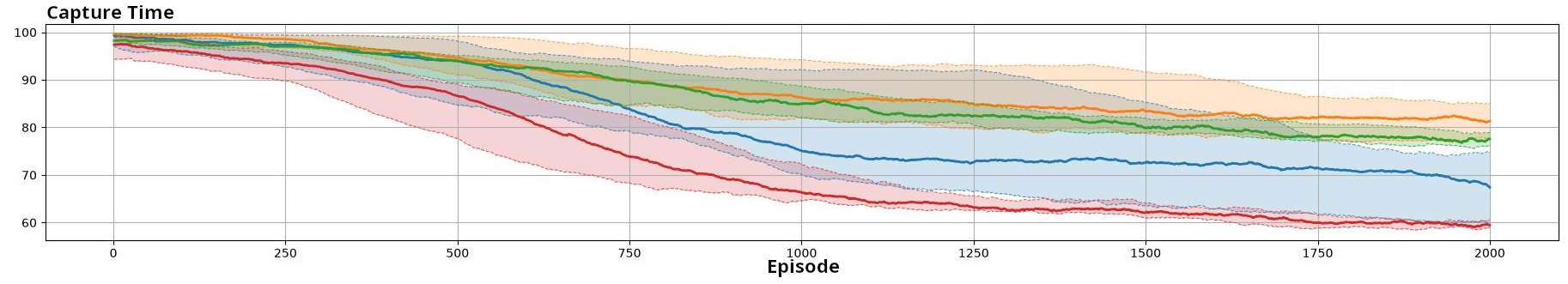}
    \caption{Number of timesteps required to catch the prey.}
    \label{fig:preyhunter_capturetime}
    \end{subfigure}
    
    \caption{Results of the prey-hunter experiment. Orange is no communication, green is policy parameter sharing, blue is observation sharing and red is both observation and policy parameter sharing.}
    \label{fig:prey_hunter_com}
\end{figure}

Figure~\ref{fig:prey_hunter_com} presents the results obtained for the different communication modes. The curves represent the median across all runs, while the colored areas indicate the range between the first and third quartiles. The data are smoothed using a moving average with a window size of $500$.

In this experiment, even with a well trained policy, the agents may fail to catch the prey because they must both have it within their observation range and be close enough to capture it. Observation sharing leads to higher rewards because it increases the likelihood that these conditions are met. With well trained policies, only one agent needs to locate the prey for the agents to coordinate and capture it. Therefore, agents that share their observations generally achieve better performance. 

Sharing policy parameters also facilitates learning. When one agent improves its policy, this improvement propagates to the other agents through parameter sharing. In this experiment, policy parameter sharing not only leads to better median performance but also produces more consistent results across runs.

\subsection{MOA Experimentation} \label{app:moa}

Section~\ref{sec:marlkit_example_moa} presents an example comparing three different modes for modeling other agents. The experimental configuration and results are detailed below.

The environment dynamics follow a \textit{Go or Yield} experiment. Both agents represent cars that can either \textit{Yield} or \textit{Go}. Each episode consists of a single timestep in which both agents select an action, resulting in a collision, one car passing, or both cars stopping.

The selected reward model is the mixed reward model (see Section~\ref{sec:reward_modeling_mixed}). The possible joint actions and their corresponding default rewards $\rho$ are defined by the reward matrix presented in Figure~\ref{fig:gooryield_matrix}.

\begin{figure}
    \centering
    \begin{tikzpicture}[>=latex]

\matrix (m) [
    matrix of nodes,
    nodes={
        rectangle,
        draw,
        minimum width=2.4cm,
        minimum height=1cm,
        align=center,
        anchor=center
    },
    column sep=-\pgflinewidth,
    row sep=-\pgflinewidth
] {
    $(-10,-10)$ & $(5,0)$ \\
    $(0,5)$    & $(-2,-2)$ \\
};

\node[above=0.25cm of m-1-1] {Go};
\node[above=0.25cm of m-1-2] {Yield};

\node[left=0.25cm of m-1-1] {Go};
\node[left=0.25cm of m-2-1] {Yield};

\node[above=0.9cm of m-1-1, xshift=1.2cm] {Agent 2};
\node[left=1.8cm of m-1-1, yshift=0.4cm, rotate=90] {Agent 1};

\end{tikzpicture}
    \caption{Go or Yield reward matrix. In each cell, the reward on the left is received by Agent 1 and the reward on the right by Agent 2.}
    \label{fig:gooryield_matrix}
\end{figure}

The tested MOA configurations are:
\begin{itemize}
    \item Independent: the agent does not model the opponent;
    \item JAL-AM with Action Frequencies: the agent evaluates joint actions and predicts the opponent's action based on the opponent's $1000$ most recent actions;
    \item JAL-GT with Minimax: the agent evaluates joint actions and assumes the worst-case action from the opponent.
\end{itemize}

In this experiment, one agent is evaluated while the other serves as its opponent. The learning agent uses a Q-learning algorithm (\class{QLearning}) and selects actions using a Q-table. The parameters $\alpha$ and $\gamma$ are respectively set to $0.2$ and $0.95$
The policy is epsilon greedy, following the Equation~\ref{eq:epsilongreedy_rewardmodel_appendix}. 
where $d$ is the decay rate, set to $10^{-4}$, and $t$ is the number of episodes.

The only difference between the learning module configurations in this experiment lies in the policy implementation. Agents that consider the opponent's actions use \class{QValueBasedJALPolicy}, which relies on the MOA module and maps joint actions to expected returns. By contrast, the standard \class{QValueBasedPolicy} evaluates only the agent's own actions.

The opponent is preconfigured with five policies, which it uses sequentially. Each policy defines fixed probabilities of selecting \textit{Go} or \textit{Yield}. Every $5000$ episodes, the opponent switches to another policy with different action probabilities. This setup evaluates whether the tested agent can adapt to changes in the opponent's behavior. The simulation lasts for $100000$ episodes.

Training and execution are decentralized, and no communication mechanism is used in this experiment.

The selected evaluation criteria measure the proportion of episodes resulting in each possible outcome: a collision, both cars stopping, or one car passing.

\begin{figure}[htbp]
    \centering
    \begin{subfigure}[t]{\textwidth}
    \includegraphics{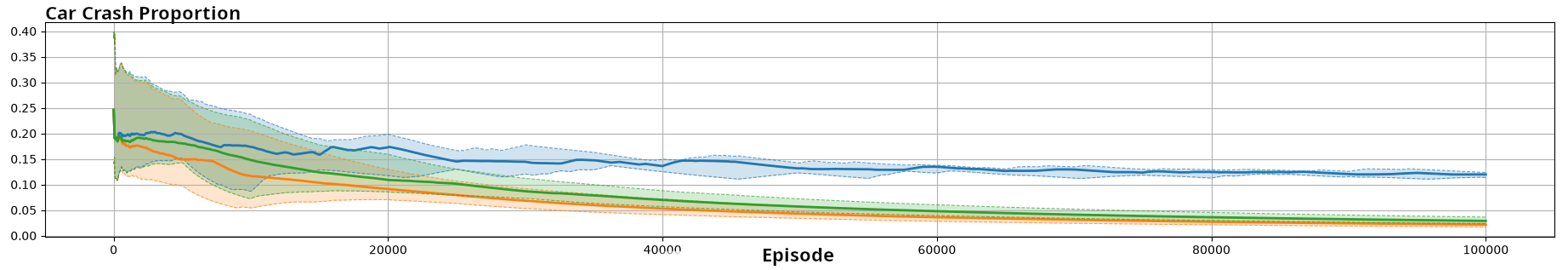}
    \caption{Proportion of episodes resulting in a collision, with both agents selecting \textit{Go}.}
    \label{fig:gooryield_gogo}
    \end{subfigure}
    
    \begin{subfigure}[t]{\textwidth}
    \includegraphics{figure/gooryield_goyield.png}
    \caption{Proportion of episodes in which one car passes, with one agent selecting \textit{Go} and the other \textit{Yield}.}
    \label{fig:gooryield_goyield}
    \end{subfigure}
    
    \begin{subfigure}[t]{\textwidth}
    \includegraphics{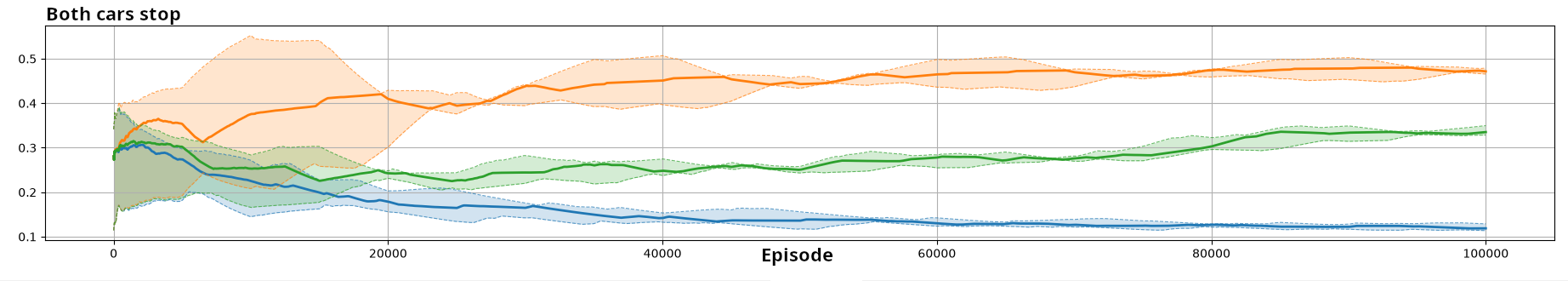}
    \caption{Proportion of episodes in which both cars stop, with both agents selecting \textit{Yield}.}
    \label{fig:gooryield_yieldyield}
    \end{subfigure}
    
    \caption{Results of \textit{Go or Yield} experiment. Green is independent, orange is minimax and blue is action frequencies modeling.}
    \label{fig:gooryield}
\end{figure}

Figure~\ref{fig:gooryield} presents the results obtained with the three different modeling modes. The curves represent the median across all runs, while the colored areas indicate the range between the first and third quartiles. The data are smoothed using a moving average with a window size of $100$.

In this experiment, the three modeling modes lead to significantly different outcomes. Minimax assumes the worst-case action from the opponent and therefore tends to select \textit{Yield} to avoid collisions. Action-frequency modeling maximizes the proportion of outcomes in which one agent selects \textit{Go} and the other selects \textit{Yield}. It adapts to the opponent's recent actions and can therefore exploit policies that frequently select \textit{Yield}. However, abrupt changes in the opponent's behavior temporarily increase the collision rate because the model requires time to adapt. The independent configuration exhibits intermediate behavior between the two MOA based configurations.


\end{document}